%% file: template.tex
\documentclass{article}
\pdfoutput=1
\usepackage[utf8]{inputenc}
\usepackage[T1]{fontenc}
\usepackage{bm}
\usepackage{siunitx}
 \usepackage{subfig}
 \usepackage{float}
\usepackage{caption}
\usepackage{arxiv}
\usepackage{xcolor}
\usepackage{arxiv}
\usepackage{graphicx}
\usepackage{subfig}

\usepackage{lmodern}
\usepackage{siunitx}
\usepackage{booktabs}
\usepackage{etoolbox}
\usepackage{soul}

\usepackage[utf8]{inputenc} 
\usepackage[T1]{fontenc}    
\usepackage{hyperref}       
\usepackage{url}            
\usepackage{booktabs}       
\usepackage{amsfonts}       
\usepackage{nicefrac}       
\usepackage{microtype}      
\usepackage[toc,page,titletoc]{appendix}
\usepackage{lineno}
\usepackage{amsmath,bm}
\usepackage{amsfonts,amssymb}
\usepackage{amsmath}

\usepackage{soul}
\usepackage[utf8]{inputenc} 
\usepackage[T1]{fontenc}    
\usepackage{hyperref}       
\usepackage{url}            
\usepackage{booktabs}       
\usepackage{amsfonts}       
\usepackage{nicefrac}       
\usepackage{microtype}      
\usepackage{lipsum}
\usepackage{graphicx}
\graphicspath{ {./images/} }
\title{COVID-19 Hospitalizations Forecasts Using Internet Search Data}
\author{Tao Wang\thanks{H. Milton Stewart School of Industrial and Systems Engineering, Georgia Institute of Technology, USA} \\ \And Simin Ma \footnotemark[1]\\\And Soobin Baek \footnotemark[1]\\\And Shihao Yang \thanks{Corresponding author.}}

\begin{document}
\maketitle


\begin{abstract}
As the COVID-19 spread over the globe and new variants of COVID-19 keep occurring, reliable real-time forecasts of COVID-19 hospitalizations are critical for public health decision on medical resources allocations such as ICU beds, ventilators, and personnel to prepare for the surge of COVID-19 pandemics. Inspired by the strong association between public search behavior and hospitalization admission, we extended previously-proposed influenza tracking model, ARGO (AutoRegression with GOogle search data), to predict future 2-week national and state-level COVID-19 new hospital admissions. Leveraging the COVID-19 related time series information and Google search data, our method is able to robustly capture new COVID-19 variants' surges, and self-correct at both national and state level. Based on our retrospective out-of-sample evaluation over 12-month comparison period, our method achieves on average 15\% error reduction over the best alternative models collected from COVID-19 forecast hub. Overall, we showed that our method is flexible, self-correcting, robust, accurate, and interpretable, making it a potentially powerful tool to assist health-care officials and decision making for the current and future infectious disease outbreak.

\end{abstract}

\thispagestyle{empty}

\section*{Introduction}
COVID-19 (SARS-CoV-19), an acute respiratory syndrome disease caused by a coronavirus, has spread worldwide causing over 120,695,785 reported cases and 4,987,755 reported deaths \cite{owidcoronavirus}. During the continuous spread of COVID-19, many variants (alpha, delta, omicron, etc.) of COVID-19 emerges, leading to drastic surges in hospital admissions and shortages in health care resources \cite{hospital_shortage}. Therefore, an accurate hospital admissions forecasting model is crucial to assist hospitals and policy makers with the possibilities and the timings of rapid changes, so as to further respond to and prepare for future COVID-19 spread.

The Centers for Disease Control and Prevention (CDC) \cite{Cramer2021-hub-dataset} has been collecting hospitalizations predictions from various research teams and making ensemble and baseline predictions since May 2020.
According to the weekly COVID-19 forecasts submissions compiled by CDC\cite{Cramer2021-hub-dataset}, machine learning \cite{GTDeep,UCSB_attention} and compartmental models \cite{MOBS,Bucky} are the most popular forecasting approaches \cite{GTDeep,UCSB_attention}. For example, Rodríguez, et al. \cite{GTDeep} uses a neural network architecture incorporating COVID-19 time series, and mobility information as inputs, whereas Jin, et al. \cite{UCSB_attention} utilizes attention and transformer models to compare and combine past COVID-19 trends for future forecasts. On the other hand, Vespignani, et al. \cite{MOBS} and Kinsey, el al. \cite{Bucky} adapt SEIR (compartmental model) as the baseline structure, and combine different exogenous variables including spatial-temporal and mobility information to build more sophisticated models to capture COVID-19 disease dynamics and forecast hospitalization. Meanwhile, statistical and data-driven models, taking advantage of COVID-19 public search information for hospitalizations predictions, have not drawn much attention. 

For the last decade, numerous studies have suggested online search data could be a valuable component to monitor and forecast infectious diseases, such as influenza \cite{GFT_2008, ARGO, Santillana_2015, Lu_2019, polgreen2008using, yuan2013monitoring}, HIV/AIDS \cite{young2018using}, dengue \cite{yang2017advances}, etc. For instance, Google Flu Trend (GFT) \cite{GFT_2008}, a digital disease detection system that uses the volume of selected Google search terms to estimate current influenza-like illnesses (ILI) activity, was the first among many studies to demonstrate how big data and public search behavior compliment traditional statistical predictive analysis. Later, many studies proposed different methodologies to improve upon GFT and provide more robust and accurate real-time forecast estimates of influenza activity in the U.S., including machine learning models \cite{Santillana_2015}, statistical models \cite{ARGO}, ensemble models \cite{Lu_2019}. Online search information from other sources, such as Yahoo \cite{polgreen2008using} and Baidu \cite{yuan2013monitoring}, have demonstrated their predictive power for influenza tracking as well.

On the other hand, to best of our knowledge, the only hospitalizations prediction model that is based on the Internet search data is a vector error correction model (VECM) proposed by Philip J.Turk, et al. \cite{turk_tran_rose_mcwilliams_2021}. VECM combines Google search data and healthcare chatbot scores and produces COVID-19 hospitalizations predictions in Greater Charlotte market area, where the Google search queries are selected based on existing literature, and filtered through hand-craft procedures. Yet, their selection approach highly depends on literature considered and healthcare chatbot scores can be limited in-access across different geographical areas. So far, none of the existing
Internet-search-based methods provide robust weeks-ahead hospitalisation forecasts for different geographical areas in the
United States that fully utilize COVID-19 related online search queries and account for their sparsity and noisiness. 

In this paper, we propose the ARGO inspired model (ARGO), which combines online search data and lagged COVID-19 related information in a L1-norm penalized linear regression model to produce real-time 1-2 weeks ahead national and state-level hospitalizations predictions over 12-month out-of-sample evaluation period in the United States. Our results show that ARGO, leveraging predictive information in the selected interpretable Google search queries, is able to largely outperform baseline models and all other public available models from CDC's COVID-19 forecast hub, which implies potential in ARGO to help with urgent public health decisions on health resources allocations.

\section*{Results}
We conducted retrospective evaluation of weekly hospital admissions for the period between January 4, 2021 and December 27, 2021, on both national and state level. To evaluate prediction performance, we calculated the root mean square error (RMSE), the mean absolute error (MAE) and the Pearson correlation coefficient (Cor) of one-week-ahead and two-week-ahead predictions (detailed in the Methods section). All comparisons are based on original scale of the ground truth of new hospital admissions released by U.S. Department of Health and Human Services (HHS)\cite{HHS_Hosp}.

\subsection*{Comparisons of National COVID-19 Hospitalizations Predictions}
 National one-week-ahead and two-week-ahead predictions of new hospitalizations were generated using (i) ARGO inspired model, (ii) persistence (naive) model and (iii) AR7 model. The naive method simply uses past week's hospitalizations as current week's forecasts, without any modeling effort. AR7 is an autoregressive model of lag 7 (implemented in R package \texttt{forecast}\cite{arima}). For fair comparisons, all models were trained on a 56-day rolling window basis. Retrospective out-of-sample predictions of daily national hospitalizations were made every week from January 4, 2021 to December 27, 2021 by the three models and were then aggregated into weekly new hospitalizations. To further demonstrate the prediction accuracy and robustness of ARGO, we collected hospitalizations predictions of the two benchmark methods (COVIDhub-baseline and COVIDhub-ensemble) from COVID-19 forecast hub \cite{Cramer2021-hub-dataset}. The COVIDhub-baseline is a persistent method that 
uses latest daily observation as future daily predictions \cite{Cramer2021-hub-dataset}. The COVIDhub-ensemble uses medians of hospitalizations predictions submitted to COVID-19 forecast hub as its ``ensemble'' forecasts.  We also provide a full comparison of the top 5 teams from CDC's COVID-19 forecast hub in Supplementary Materials (Table \ref{tab:Nat_all}). 

\begin{table}[htbp]
\sisetup{detect-weight,mode=text}
\renewrobustcmd{\bfseries}{\fontseries{b}\selectfont}
\renewrobustcmd{\boldmath}{}
\newrobustcmd{\B}{\bfseries}
\addtolength{\tabcolsep}{-4.1pt}
\footnotesize
\centering
\caption{National Level Comparison Error Metrics}\label{tab:Nat_two_teams}
\begin{tabular}{|l|r|r|r|r|r|r|}
  \hline
\multicolumn{1}{|c}{Methods} & \multicolumn{2}{|c}{RMSE} & \multicolumn{2}{|c}{MAE}& \multicolumn{2}{|c|}{Cor} \\  \cline{2-7}
 & 1 Week Ahead & 2 Weeks Ahead & 1 Week Ahead & 2 Weeks Ahead& 1 Week Ahead & 2 Weeks Ahead\\\hline
ARGO & \B4667.552& \B 14010.648 & \B2896.857 &  \B9435.759 &\B0.988 &\B0.906\\\hline
AR7 & 6408.431 & 15750.873 &4726.405    & 10957.369 &   0.973& 0.834\\\hline
COVIDhub-ensemble\cite{Cramer2021-hub-dataset} &  9578.408  & 16327.146 &  6301.981  & 10856.115 &  0.942  & 0.851  \\\hline
COVIDhub-baseline\cite{Cramer2021-hub-dataset} &  10442.092  & 19210.623 &  7167.538  & 13183.231  &  0.916  & 0.738  \\\hline
Naive & 10528.225 & 19831.180 &7033.212  & 13633.750 &  0.918 & 0.732\\\hline
\end{tabular}

\vspace{1ex}
{\raggedright Error metrics of national one-week-ahead and two-week-ahead new hospitalizations predictions. The best scores are highlighted with boldface. All comparisons are based on the original scale of hospitalizations released by HHS. Methods are sorted by their average RMSE of one-week-ahead and two-week-ahead predictions. On average, the ARGO model outperforms the best alternative method by approximately 18\% in RMSE, 25\% in MAE and 4\% in Cor. Overall, ARGO has better predictions than all the benchmark methods during our comparison period.  \par}
\end{table}
Table \ref{tab:Nat_two_teams} summarizes the national one-week-ahead and two-week-ahead predictions performance from January 4, 2021 to December 27, 2021. During this period, ARGO outperforms all the benchmark models in every error metric for both one and two weeks ahead predictions. Specifically, for the national one-week-ahead predictions, ARGO performs better than the best alternative method by around 27\% in RMSE, 39\% in MAE and 1.5\% in Cor. The two-week-ahead ARGO forecasts have slightly lower error reduction in RMSE and MAE, and higher increase in Cor. The results demonstrate that ARGO is able to produce accurate and robust retrospective out-of-sample national one-week-ahead and two-week-ahead hospitalizations predictions during the evaluation period.

\begin{figure}[htbp]
\centering
\subfloat[One-Week-Ahead National Level Predictions]
{\includegraphics[width=0.49\textwidth]{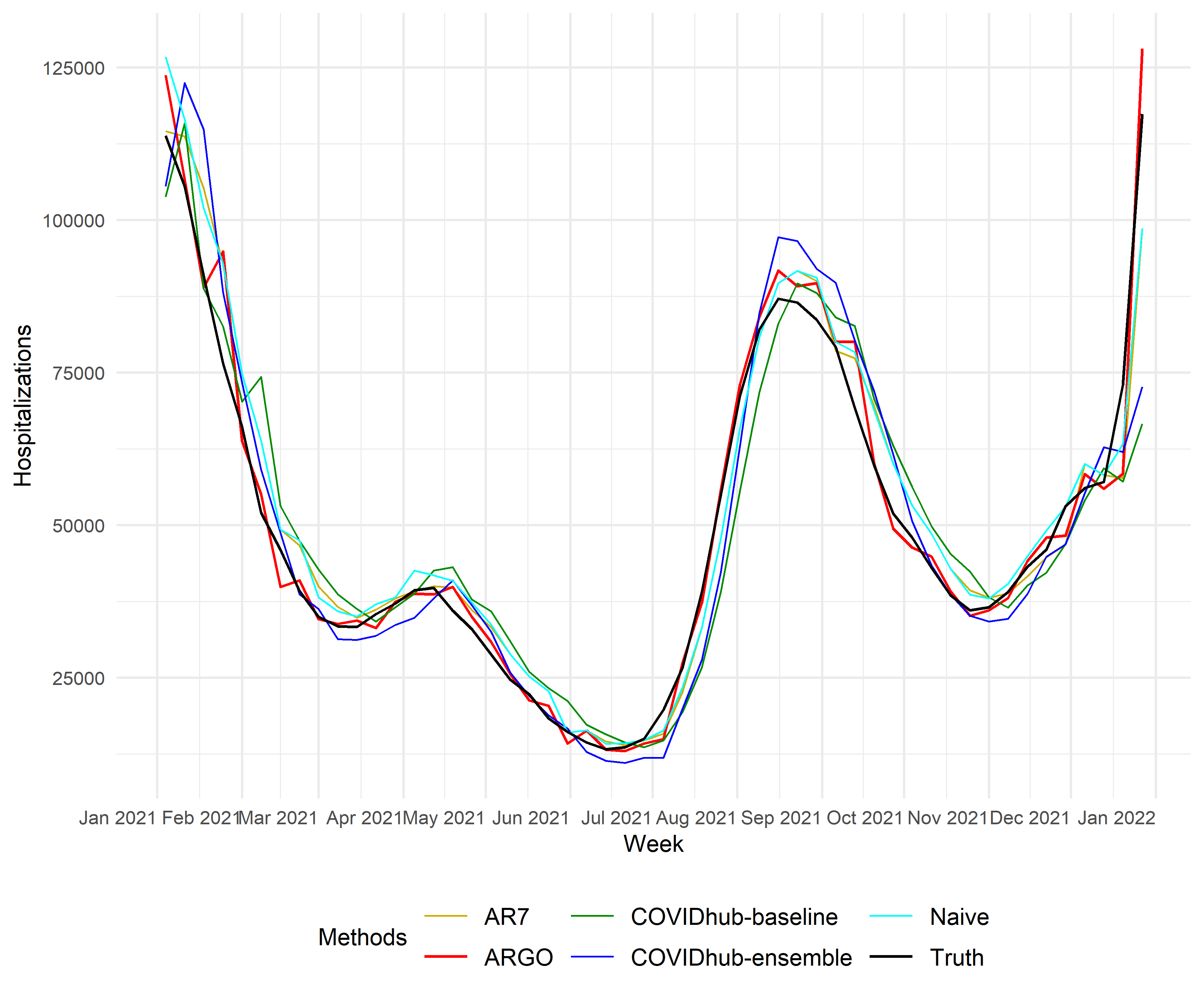}\label{fig:Nat_Compare_Our_1Week}}
\hfill
\subfloat[Two-week-ahead National Level Predictions]
{\includegraphics[width=0.49\textwidth]{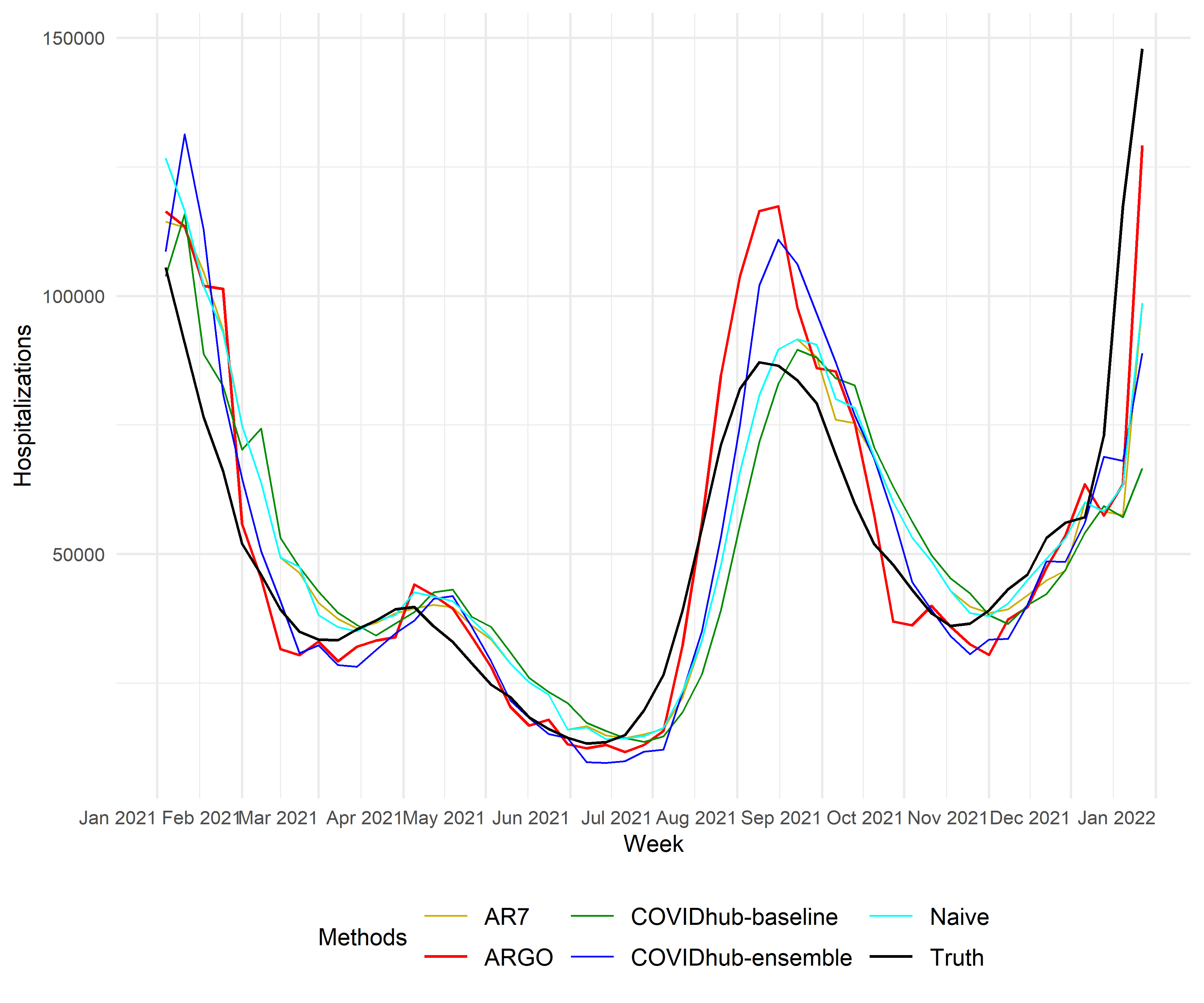}\label{fig:Nat_Compare_Our_2Weeks}}
\caption{One-week-ahead and two-week-ahead hospitalizations predictions of the 5 compared models. 1 and 2 weeks ahead predictions are compared weekly from 2021-01-04 to 2021-12-27. True new hospital admissions released by HHS are marked in black. Predictions of the 5 models, which consist of ARGO, AR7, naive, COVIDhub-baseline, COVIDhub-ensemble, are marked in red, gold, cyan, green, and blue, respectively.}
\label{fig1}
\end{figure}

Figure \ref{tab:Nat_two_teams} displays the 5 compared methods' estimations. During the 12-month comparison period, ARGO can accurately capture the overall trend as well as local fluctuations (the spike between April, 2021 and May, 2021) in new hospitalizations. All of the time series forecasting methods exhibit some delaying behaviors to various degree, due to the input feature of the lagged information. Fortunately, by utilizing Google search information, ARGO is able to overcome such delayed behavior and is the only method that captures the hospital admission peaks around April 2021 and September 2021 as well as the surge around December 2021 possibly caused by omicron for both 1 and 2 weeks ahead predictions. Moreover, by leveraging time series information and the Internet search information, ARGO is able to avoid sudden spikes and drops in the prediction. ARGO is also self-correcting and can quickly recover from the over-shooting behavior (e.g. between August to September 2021 for two weeks ahead prediction).

\subsection*{Comparisons of State-level COVID-19 Hospitalizations}
We also conducted retrospective out-of-sample one and two weeks ahead predictions for the 51 states in the U.S. (including Washington D.C.) during the same period of January 4, 2021 to December 27, 2021. 

\begin{table}[htbp]
\sisetup{detect-weight,mode=text}
\renewrobustcmd{\bfseries}{\fontseries{b}\selectfont}
\renewrobustcmd{\boldmath}{}
\newrobustcmd{\B}{\bfseries}
\addtolength{\tabcolsep}{-4.1pt}
\footnotesize
\centering
\caption{State-Level Comparison Error Metrics}\label{tab:state_two_teams}
\begin{tabular}{|l|r|r|r|r|r|r|}
  \hline
\multicolumn{1}{|c}{Methods} & \multicolumn{2}{|c}{RMSE} & \multicolumn{2}{|c}{MAE}& \multicolumn{2}{|c|}{Cor} \\  \cline{2-7}
 & 1 Week Ahead & 2 Weeks Ahead & 1 Week Ahead & 2 Weeks Ahead& 1 Week Ahead & 2 Weeks Ahead\\\hline
ARGO& \B170.960& \B 374.330 & \B114.98 &  \B243.765 &\B0.960
&\B0.879\\\hline
AR7 & 193.424 & 399.253 &134.454    & 276.255 &0.951& 0.849\\\hline
COVIDhub-ensemble\cite{Cramer2021-hub-dataset} &  259.348& 427.926  &  161.523  & 265.042 &  0.937  & 0.867 \\\hline
Naive & 265.789& 469.206 &179.084  & 322.021  &  0.934 & 0.825\\\hline
COVIDhub-baseline\cite{Cramer2021-hub-dataset}  &  296.963 & 482.160 &  193.150 & 320.048  &  0.862  & 0.751 \\\hline

\hline
\end{tabular}

\vspace{1ex}
{\raggedright Error metrics of state-level one-week-ahead and two-week-ahead new hospitalizations predictions, averaging across the states. The best scores are highlighted with boldface. All comparisons are based on the original scale of hospitalizations released by HHS. Methods are sorted based on their average RMSE of one-week-ahead and two-week-ahead predictions. ARGO outperforms the best alternative method by approximately 8\% in RMSE, 12\% in MAE and 1\% in Cor. Overall, ARGO is the best-performing prediction model compared with other listed models. \par}
\end{table}

\begin{figure}[H]
\centering
\includegraphics[width=0.8\linewidth]{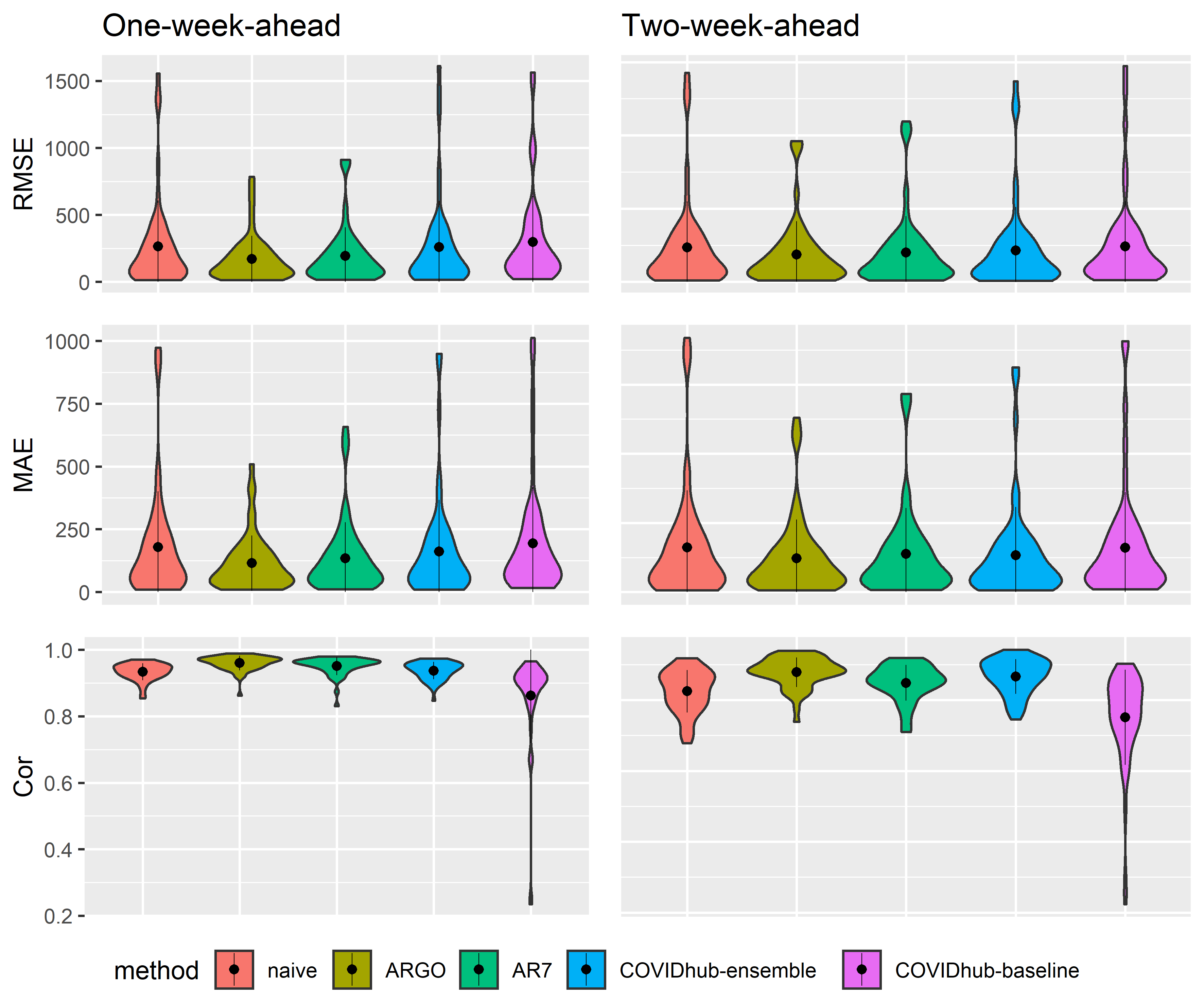}
\caption{ The distributions of error metrics of one-week-ahead and two-week-ahead predictions from 2021-01-04 to 2021-12-27 over the 51 states of the 5 compared models. The embedded black dot indicates mean. The vertical line represents 1 standard deviation range which is trimmed to be inside normal range of RMSE, MAE and correlations coefficients.}
\label{fig:state-distribution}
\end{figure}

Table \ref{tab:state_two_teams} summarizes the average error metrics of all methods' state-level predictions from January 4, 2021 to December 27, 2021. For the one-week-ahead predictions, ARGO is able to achieve uniformly best performance in all error metrics. Compared with the two benchmark models from COVID-19 forecast hub, ARGO yields roughly 35\% error reduction in RMSE, around 30\% error reduction in MAE and around 2\% increase in Pearson correlation coefficient. For the two-week-ahead predictions, ARGO achieves around 12\% error reduction in RMSE, approximate 8\% error reduction in MAE  and around 1\% increase in Pearson correlation coefficient compared with best alternative models from COVID-19 forecast hub (table \ref{tab:state_all}). Overall, ARGO gives the leading performance in state-level compared with the benchmark models, by efficiently utilizes relevant public search information and incorporating cross-state cross-region information as model features. The full comparison of state-level hospitalizations predictions is shown in SI Table \ref{tab:state_all}. Figure \ref{fig:state-distribution} further demonstrates the accuracy and robustness of ARGO in state-level hospitalizations predictions. The violin charts, which present the distributions of each model's predictions errors in all three error metrics, show that the 1 standard deviation ranges of ARGO are the smallest in RMSE and MAE, and are the second best in Cor.  

\section*{Discussion}
The ARGO inspired model combines autoregressive COVID-19 related information and online search data. It is able to produce accurate, reliable real-time hospitalizations predictions at both national and state-level for 1-2 weeks ahead predictions. The state-level real-time hospitalizations predictions made by ARGO could help local public health officials make timely allocation decisions of healthcare resources, such as ventilator, ICU beds, personal protective equipment, personnel, etc, as well as account and promptly prepare for future surges of COVID-19 pandemics caused by new virus variants. Furthermore, our ARGO hospitalization prediction model is a straightforward adaptation from the original ARGO model for influenza \cite{ARGO}, which reduces the chance of overfitting and again demonstrates ARGO's robustness and flexibility.

Although ARGO shows strong performance in hospitalizations forecasts, its accuracy is controlled by the reliability of the inputs. Google search data can be noisy due to the instability of Google Trends’ sampling approach and public fear. Especially for state-level Google search data, the lack of search intensity can make the search data unrepresentative of the real interest of the people. Luckily, the IQR filter\cite{ARGOX-COVID-19} and moving average smoothing applied to Google search data are able to minimize the risk caused by such noisiness, and help ARGO produce robust output. To further account for the instability in the state-level Google search queries, the query terms are identified using the national level data where the search frequencies are more representative with lower noise and higher stability. ARGO selects the most representative search queries according to their Pearson correlation coefficients with hospital admission. In addition, the national level search frequency is directly used as input features for state-level predictions. An optimal delay between the search data and the hospital admissions is also identified for each query term. All together, ARGO is able to overcome the sparsity issues of Google search queries and produce robust future estimations (Figure \ref{fig1}) while avoiding over-fitting.


Another challenge in using online search data to estimate hospitalizations is that the predictive information in Google search data die down as forecasts horizon expands (shown in Table \ref{tab:Nat_two_teams} and \ref{tab:state_two_teams}). In our results of COVID-19 hospitalizations predictions, the optimal lags (delays) of some Google search terms are relative small shown in table \ref{tab:5query_optimallag} which indicates those Google search queries are more effective for short-term prediction of hospitalizations. Nevertheless, by leveraging COVID-19 related time series information, ARGO is able to adjust the focus between the time series and the Internet search information features when forecast horizon extends, thanks to the L1-norm penalty and the dynamic training that selects the most relevant Google search terms. 

By effectively combining the Google search data and COVID-19 related time series information, ARGO has a stable model structure that is able to make accurate and robust national and state level 1-2 weeks ahead hospitalizations predictions. The simple structure of ARGO makes it universal adaptable to other COVID-19 related forecasts. With its simplicity and strengthened accuracy over other benchmark methods, ARGO could help public health decision making for the local monitoring and control of COVID-19 to better prepare for future surges of hospitalizations, patients in ICU, deaths and cases caused by new COVID-19 variants such as alpha, delta, omicron, etc. 

\section*{Data and Methods}
We focused on national hospital admission predictions and state-level predictions of 51 states in the United States, including Washington D.C.. The inputs consist of confirmed incremental cases, percentage of vaccinated population, confirmed new hospital admissions, and Google search query frequencies. Both state-level data and national data were directly obtained from respective data sources outlined in this section. Our prediction method is inspired by ARGO\cite{ARGO}, with details presented in this section as well.

\subsection*{COVID-19 Related Data}
We used reported COVID-19 confirmed incremental cases from JHU CSSE data\cite{JHU_Data}, percentage of vaccinated population from Centers for Disease Control and Prevention (CDC)\cite{CDC_Cases} and COVID-19 confirmed new hospital admissions from HHS \cite{HHS_Hosp}. The data sets were collected from July 15, 2020 to January 15, 2021. 
\subsection*{Google Search Data}
Google Trends provides estimated Google search frequency for the specified query term\cite{GoogleTrends}. We obtained online search data from Google Trends\cite{GoogleTrends} for the period from July 15, 2020 to January 15, 2021. To retrieve the time series search frequencies of a desired query, one needs to specify the query's geographical information and time frame on Google Trends. The returned frequency from Google Trends is obtained by sampling all raw Google search frequencies that contain this query\cite{GoogleTrends}. We collected daily Google Search frequencies of 256 top searched COVID-19-related terms based on the previous work of COVID-19 death forecasts\cite{ARGOX-COVID-19}.  

\subsubsection*{Inter-Quantile Range (IQR) Filter and Optimal Lag for Google Search Data}
The raw Google search frequencies obtained from Google Trends\cite{GoogleTrends} are observed to be unstable and sparse\cite{ARGOX-COVID-19}. Such instability and sparsity can negatively affect prediction performance of linear regression models which are sensitive to outliers. To deal with such outliers in Google search data, we used an IQR filter \cite{ARGOX-COVID-19} to remove and correct outliers on a rolling window basis. The search data that is beyond 3 standard deviation from the past 7-day average are examined and removed\cite{ARGOX-COVID-19}.

\begin{figure}[htbp]
\centering
\includegraphics[width=0.7\textwidth]{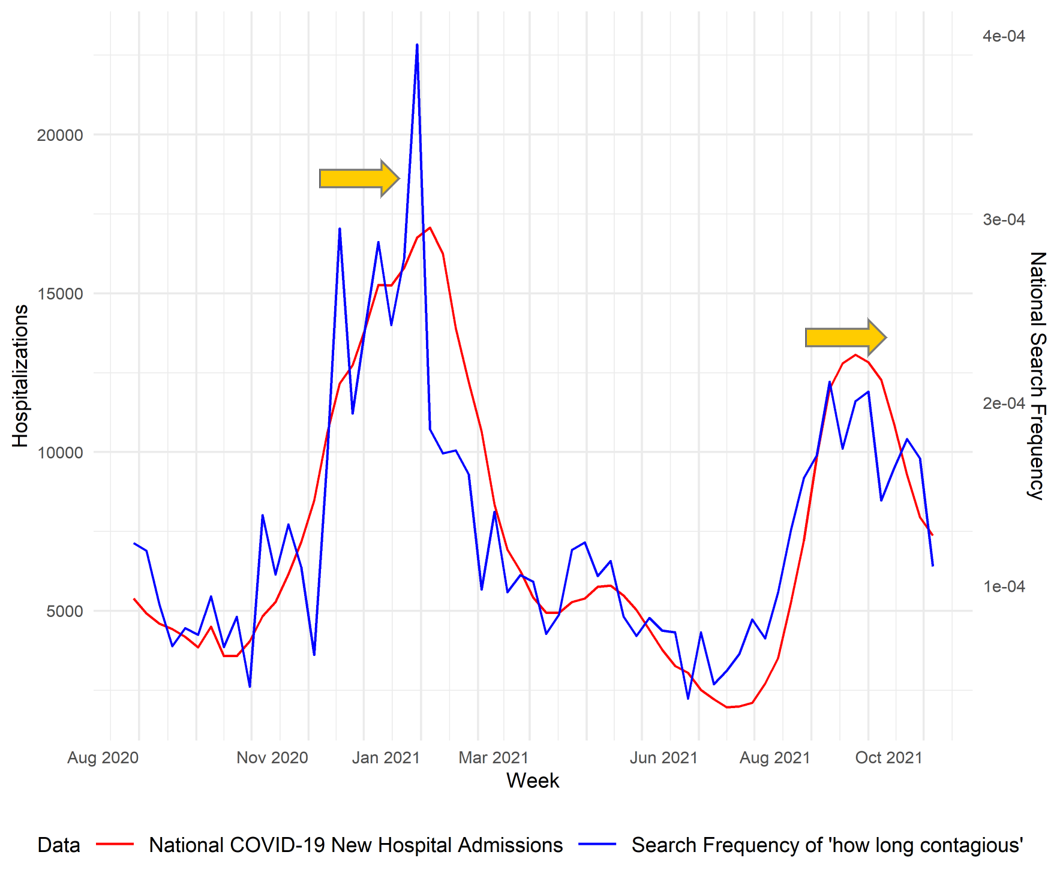}
\caption{{\bf Google search query ``how long contagious'' and COVID-19 weekly new hospitalizations}
Illustration of delay in peak between Google search query search frequencies (how long contagious in blue) and COVID-19 national level weekly new hospitalizations (red). Y-axis are adjusted accordingly.}
\label{Optimal_lag_intuition}
\end{figure}

The trends of Google search frequencies are often a few days ahead of hospitalizations, indicating that the search data might contain predictive information about hospitalizations. Figure \ref{Optimal_lag_intuition} demonstrates the delay behavior between Google search query frequencies and national hospitalizations. To fully utilize the predictive information in national Google search terms, we apply optimal lags\cite{ARGOX-COVID-19} to filtered Google search frequencies to match the trends of national hospitalizations. For each query, a linear regression of COVID-19 new hospitalizations is fitted against lagged Google search frequency, considering a range of lags. The lag results in lowest mean square error is selected as the optimal lag for that query. The data used to find optimal lags are from August 1, 2020 to December 31, 2020.

\subsubsection*{Selection of Google Search Terms}
Among the 256 COVID-19 related terms, we further selected the queries that have highest correlation coefficients larger than 0.6 with national COVID-19 hospitalizations for the period from August 1, 2020 to December 31, 2020. We applied 7-day moving average to further smooth out weekly fluctuations in the selected Google search queries. Table \ref{tab:5query_optimallag} shows the 11 selected important terms as well as their optimal lags. Table \ref{tab:5query_optimallag} supports our hypothesis that when people get infected, they would probably search for general query like "symptoms of the covid-19" first as this query has relative large optimal lag. After the symptoms develop, people might begin to search for specific symptoms such as "loss of smell" which has relatively smaller optimal lag.
\begin{table}[ht]
\footnotesize
\centering
\caption{Optimal Lags of Selected Important Terms}\label{tab:5query_optimallag}
\begin{tabular}{llll}
Selected Google Search Term & Optimal Lag (in days)\\
  \hline
how long contagious & 	4\\
loss of smell &          			4  \\
loss of taste   &       		9 \\
covid-19 vaccine. & 7 \\
Cough & 7 \\
pneumonia & 7\\
how long covid-19    &          4\\
sinus & 21 \\
symptoms of the covid-19       &     			7\\
contagious coronavirus   &     			4\\
coronavirus vaccine & 4\\ 

\hline
\end{tabular}
\end{table}

\subsection*{ARGO Inspired Prediction}
Let $\hat{y}_{t,r}$ be the daily hospital admissions of region $r$ on day $t$; $X_{k,t}$ be the Google search data of term $k$ on day $t$; $c_{t,r}$ be the JHU COVID-19 incremental confirmed cases on day $t$ of region $r$; $v_{t,r}$ be the cumulative percent of people who get vaccinated by day $t$ of region $r$; $\mathbb{I}_{\{t, d\}}$ be the weekday indicator for day $t$ (i.e. $\mathbb{I}_{\{t, 1\}}$ indicates day $t$ being Monday).
Standing on day T, to predict $l$-day-ahead hospital admission of state $r$, $\hat{y}_{T+l,r}$, we used penalized linear estimator as following: 
\begin{equation}\label{eqn:argo_yhat_state}
\begin{aligned}
    \hat{y}_{T+l,r} = \hat{\mu}_{y,r}+\sum^{I}_{i=0}\hat{\alpha}_{i,r}y_{T-i,r} +
    \sum_{j\in {J}}\hat{\beta}_{j,r}c_{T+l-j,r}+
    \sum_{m\in {M}_{r}}\hat{\gamma}_{m,r}y_{T,m}+
    \sum_{q\in {Q}}\hat{\phi}_{q,r}v_{T+l-q,r}+
    \sum^{K}_{k=1}\hat{\delta}_{k,r}X_{k,T+l-\hat{O}_k}
    + \sum^6_{d=1}\hat{\tau}_{d,r}\mathbb{I}_{\{T+l, d\}}
\end{aligned}
\end{equation}

Where $I=6$ considering consecutive one week lagged daily hospital admissions; $J=\max\left(\{7,28\},l\right)$, considering lagged confirmed cases; ${M}_{r}$ is the set of geographical neighboring states of state $r$; $Q=\max\left(7,l\right)$, considering vaccination data lagged by one week; $\hat{O}_k=\max\left(O_k,l\right)$ is the adjusted optimal lag for term $k$;  $K=11$, considering 11 selected Google search terms. The coefficients for $l$-day-ahead predictions of region $r$, $\{\mu_{y,r},\bm{\alpha}=(\alpha_{1,r},\ldots,\alpha_{6,r}), \bm{\beta}=(\beta_{1,r}, \ldots, \beta_{|J|,r}), \bm{\gamma}=(\gamma_{1,r},\ldots,\gamma_{|{{M}_{r}}|,r}), \bm{\phi}=\phi_{max(7,l),r}, \bm{\delta}=(\delta_{1,r},\ldots,\delta_{11,r}), \bm{\tau}=(\tau_{1,r}, \ldots, \tau_{6,r})\}$, were computed by

\begin{equation}\label{eqn:argo_obj_state}
    \begin{aligned}
    \underset{\mu_{y,r},\bm{\alpha},\bm{\beta},\bm{\gamma},\bm{\phi},\bm{\delta},\bm{\tau},\bm{\lambda}}{\mathrm{argmin}} 
    \sum_{t=T-M-l+1}^{T-l} &\omega^{T-l-t+1}\Bigg( y_{t+l,r}-\mu_{y,r} - \sum^{6}_{i=0}{\alpha}_{i,r}y_{t-i,r}-\sum_{j\in {J}}\hat{\beta}_{j,r}c_{t+l-j,r}-\sum_{m\in {M}_{r}}\hat{\gamma}_{m,r}y_{t,m}\\
    \;\;\;&- \sum_{q\in {Q}}\hat{\phi}_{q,r}v_{t+l-q,r}
    -\sum^{5}_{k=1}\hat{\delta}_{k,r}X_{k,t+l-\hat{O}_k}
    - \sum^6_{d=1}\hat{\tau}_{d,r}\mathbb{I}_{\{t+l, d\}}\Bigg)^2\\
    \;\;\;&
    + \lambda_\alpha\|\bm{\alpha}\|_1+\lambda_\beta\|\bm{\beta}\|_1+\lambda_\gamma\|\bm{\gamma}\|_1+ \lambda_\phi \|\bm{\phi}\|_1+\lambda_\delta\|\bm{\delta}\|_1 +\lambda_\tau \|\bm{\tau}\|_1
    \end{aligned}
\end{equation}
M = 56 which is the length of our training period; $\omega = 0.8$ is the exponentially time-decaying weight which assigns higher weight on more recent observation. Region $\bm{r}$ consists of U.S. and its 51 states, including Washington D.C.. For U.S. national level training, the hospitalizations of neighboring states, $y_{t,m}$, and their coefficients, $\bm{\gamma}$, are excluded. To address the sparsity of Google search data, we used penalty of L1-norm. For simplicity, the hyperparameters $\bm{\lambda}=(\lambda_{\alpha},\lambda_{\beta},\lambda_{\gamma},\lambda_{\phi},\lambda_{\delta},\lambda_{\tau})$ for L1-norm penalty were set to be equal and obtained via 10-folds cross-validation.\\

With the formulation above, on each Monday from January 4, 2021 to December 27, 2021, we iteratively trained our model and made national and state-level retrospective out-of-sample hospitalizations predictions up to 14 days into future. We then aggregated daily predictions into one-week-ahead and two-week-ahead predictions. For example, $\hat{y}_{T+1:T+7,r}=\sum^7_{i=1}\hat{y}_{T+i,r}$ and $\hat{y}_{T+8:T+14,r}=\sum^{14}_{i=8}\hat{y}_{T+i,r}$ are the one-week-ahead prediction and two-week-ahead prediction on day $T$ of region $r$, respectively.

\subsection*{Evaluation Metrics}
    \label{s:errormetric}
Root Mean Squared Error (RMSE) between a hospitalization estimate $\hat{y}_t$ and the true value $y_t$ over period $t=1,\ldots, T$ is $\sqrt{\frac{1}{T}\sum_{t=1}^T \left(\hat{y}_t - y_t\right)^2}$. Mean Absolute Error (MAE) between an estimate $\hat{y}_t$ and the true value $y_t$ over period $t=1,\ldots, T$ is $\frac{1}{T}\sum_{t=1}^T \left|\hat{y}_t - y_t\right|$. 
Correlation is the Pearson correlation coefficient between $\hat{\bm{y}}=(\hat{y}_1, \dots, \hat{y}_T)$ and $\bm{y}=(y_1,\dots, y_T)$. All estimates $\hat{y}_t$ and the true value $y_t$ were weekly aggregated before calculating RMSE, MAE and Cor. 

\bibliographystyle{unsrt}
\bibliography{References}

\section*{Author contributions statement}
T.W., S.M., S.B. and S.Y. designed the research; T.W., S.M., S.B. and S.Y. performed the research; T.W. and S.B. analyzed data and conducted the experiment(s); T.W. and S.M. analysed the results. T.W., S.M. and S.Y. wrote the paper. All authors reviewed the manuscript. 

\section*{Competing interests}
The authors declare no competing interests.

\clearpage
\section*{Additional information}
\input{supplementary}

\end{document}

%% file: supplementary.tex
\renewcommand{\thefigure}{S\arabic{figure}}
\renewcommand{\thetable}{S\arabic{table}}

After filtering out teams with incomplete submissions and miss-aligned prediction dates, we collected predictions of the top 5 models from COVID-19 forecast hub \cite{Cramer2021-hub-dataset} for comparisons.

\begin{table}[htbp]
\sisetup{detect-weight,mode=text}
\renewrobustcmd{\bfseries}{\fontseries{b}\selectfont}
\renewrobustcmd{\boldmath}{}
\newrobustcmd{\B}{\bfseries}
\addtolength{\tabcolsep}{-4.1pt}
\footnotesize
\centering
\caption{National Level Comparison Error Metrics}\label{tab:Nat_all}
\begin{tabular}{|l|r|r|r|r|r|r|}
  \hline
\multicolumn{1}{|c}{Methods} & \multicolumn{2}{|c}{RMSE} & \multicolumn{2}{|c}{MAE}& \multicolumn{2}{|c|}{Cor} \\  \cline{2-7}
 & 1 Week Ahead & 2 Weeks Ahead & 1 Week Ahead & 2 Weeks Ahead& 1 Week Ahead & 2 Weeks Ahead\\\hline
 
ARGO & \B4667.551& \B 14182.659 & \B2896.857 &  \B9435.759 &\B0.988 &\B0.906\\\hline
AR7 & 6408.430 & 15750.873 &4726.405    & 10957.369 &   0.973& 0.834\\\hline
MOBS-GLEAM-COVID\cite{MOBS} &  8211.132  & 15909.210  &  6579.988
  & 11714.361 &  0.960  & 0.937 \\\hline
COVIDhub-ensemble\cite{Cramer2021-hub-dataset} &  9578.408  & 16327.146 &  6301.981  & 10856.115 &  0.942  & 0.851  \\\hline
GT-DeepCOVID\cite{GTDeep} &  9179.514  & 17704.803   &  6481.583  & 12721.385 &  0.943 & 0.815 \\\hline
COVIDhub-baseline\cite{Cramer2021-hub-dataset} &  10442.092  & 19210.623 &  7167.538  & 13183.231  &  0.916  & 0.738  \\\hline
Naive & 10528.225 & 19831.180 &7033.212  & 13633.750 &  0.918 & 0.732\\\hline
JHUAPL-Bucky\cite{Bucky} &  12393.690 & 18200.825&  9121.523  & 12944.933 &  0.915 & 0.877\\\hline

\hline
\end{tabular}

\vspace{1ex}
{\raggedright Error metrics of national one-week-ahead and two-weeks-ahead new hospitalizations predictions. The best scores are highlighted with boldface. All comparisons are based on the original scale of hospitalizations released by HHS. Methods are sorted by their average RMSE of one-week-ahead and two-weeks-ahead predictions. CDC forecast hub collects predictions from every team that submits their weekly predictions, which all have their strengths in different time periods. On average, the ARGO model outperforms the benchmark methods (Naive, AR7, COVIDhub-ensemble,COVIDhub-baseline) by approximately 18\% in RMSE, 25\% in MAE and 4\% in Cor. Overall, ARGO has better predictions than all other models during our comparison period.  \par}
\end{table}

\begin{table}[htbp]
\sisetup{detect-weight,mode=text}

\renewrobustcmd{\bfseries}{\fontseries{b}\selectfont}
\renewrobustcmd{\boldmath}{}
\newrobustcmd{\B}{\bfseries}
\addtolength{\tabcolsep}{-4.1pt}
\footnotesize
\centering
\caption{State-Level Comparison Error Metrics}\label{tab:state_all}
\begin{tabular}{|l|r|r|r|r|r|r|}
  \hline
\multicolumn{1}{|c}{Methods} & \multicolumn{2}{|c}{RMSE} & \multicolumn{2}{|c}{MAE}& \multicolumn{2}{|c|}{Cor} \\  \cline{2-7}
 & 1 Week Ahead & 2 Weeks Ahead & 1 Week Ahead & 2 Weeks Ahead& 1 Week Ahead & 2 Weeks Ahead\\\hline
ARGO& \B170.960& \B 374.33 & \B114.98 &  \B243.765 &\B0.960
&\B0.879\\\hline
AR7 & 193.424 & 399.253 &134.454    & 276.255 &0.951& 0.849\\\hline
COVIDhub-ensemble\cite{Cramer2021-hub-dataset} &  259.348& 427.926  &  161.523  & 265.042 &  0.937  & 0.867 \\\hline
Naive & 265.789& 469.206 &179.084  & 322.021  &  0.934 & 0.825\\\hline
MOBS-GLEAM-COVID\cite{MOBS} &  281.464 & 467.701   &  195.785 & 314.702  &  0.910  & 0.809 \\\hline
COVIDhub-baseline\cite{Cramer2021-hub-dataset}  &  296.963 & 482.16 &  193.150 & 320.048  &  0.862  & 0.751 \\\hline
GT-DeepCOVID\cite{GTDeep} &  282.921  & 497.770   &  197.385  & 344.886 &  0.900 & 0.778 \\\hline
JHUAPL-Bucky\cite{Bucky} &  458.423 & 636.001 &  291.534 & 408.232 &  0.874  & 0.811 \\\hline

\hline
\end{tabular}

\vspace{1ex}
{\raggedright Error metrics of state-level one-week-ahead and two-weeks-ahead new hospitalizations predictions, averaging across the states. The best scores are highlighted with boldface. All comparisons are based on the original scale of hospitalizations released by HHS. Methods are sorted based on their average RMSE of one-week-ahead and two-weeks-ahead predictions. On average, the ARGO model outperforms the best alternative method by approximately 18\% in RMSE, 25\% in MAE and 4\% in Cor. Overall, ARGO has better prediction performance than other listed methods during our comparison period. \par}
\end{table}

\clearpage
\newgeometry{left=1.5cm,bottom=2cm}
\subsection*{Detailed comparison for each state}
\input{State_Compare}
\restoregeometry

%% file: State_Compare.tex
\begin{table}[ht]
\footnotesize
\centering
\begin{tabular}{|l|r|r|r|r|r|r|}
   \hline\multicolumn{1}{|c}{Methods} & \multicolumn{2}{|c}{RMSE} & \multicolumn{2}{|c}{MAE}& \multicolumn{2}{|c|}{Cor} \\  \cline{2-7} & 1 Week Ahead & 2 Week Ahead & 1 Week Ahead & 2 Week Ahead & 1 Week Ahead & 2 Week Ahead \\ 
  \hline
ARGO & 13.333 & 18.005 & 9.921 & 12.438 & 0.881 & 0.799 \\ 
   \hline
AR7 & 17.197 & 22.752 & 12.477 & 15.208 & 0.802 & 0.673 \\ 
   \hline
COVIDhub-ensemble & 15.310 & 17.608 & 10.365 & \textbf{12.288} & 0.847 & 0.815 \\ 
   \hline
Naive & \textbf{12.996} & \textbf{16.775} & \textbf{9.635} & \textbf{12.288} & \textbf{0.883} & \textbf{0.826} \\ 
   \hline
MOBS-GLEAM\_COVID & 18.376 & 24.910 & 13.055 & 16.764 & 0.840 & 0.708 \\ 
   \hline
COVIDhub-baseline & 22.075 & 26.171 & 16.558 & 19.596 & 0.688 & 0.597 \\ 
   \hline
GT-DeepCOVID & 14.265 & 18.030 & 11.369 & 13.202 & 0.864 & 0.808 \\ 
   \hline
JHUAPL-Bucky & 19.169 & 24.398 & 14.328 & 17.772 & 0.762 & 0.710 \\ 
   \hline
\end{tabular}
\caption{Comparison of different methods for state-level COVID-19 1 to 2 weeks ahead hospitalizations predictions in Vermont (VT). The MSE, MAE, and correlation are reported and best performed method is highlighted in boldface.} 
\end{table}
\begin{figure}[!h] 
  \centering 
\includegraphics[width=.8\linewidth, page=2]{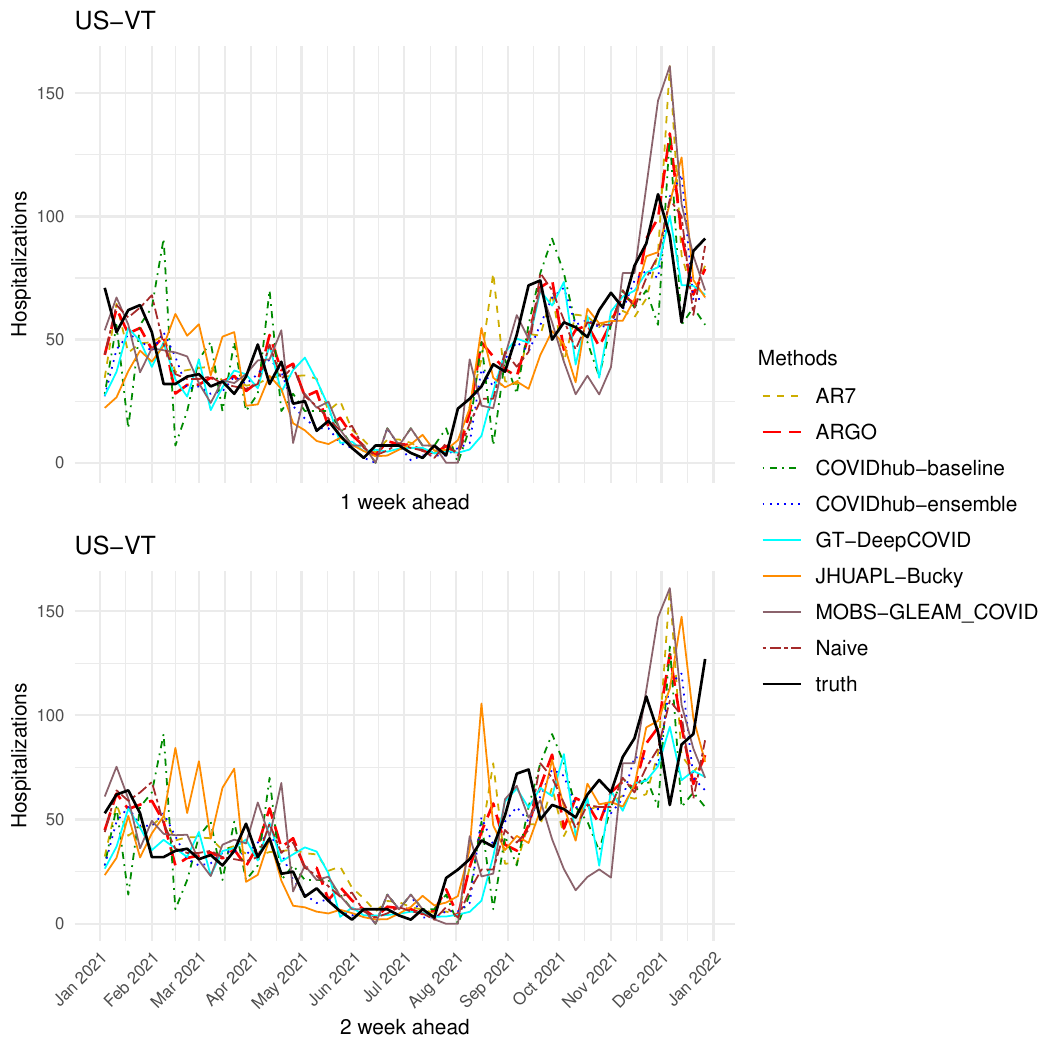} 
\caption{Plots of the COVID-19 hospitalizations 1 week (top), 2 weeks (bottom) ahead estimates of all compared models for Vermont(VT).}
\label{fig:State_Ours_VT}
\end{figure}
\newpage   
\begin{table}[ht]
\footnotesize
\centering
\begin{tabular}{|l|r|r|r|r|r|r|}
   \hline\multicolumn{1}{|c}{Methods} & \multicolumn{2}{|c}{RMSE} & \multicolumn{2}{|c}{MAE}& \multicolumn{2}{|c|}{Cor} \\  \cline{2-7} & 1 Week Ahead & 2 Week Ahead & 1 Week Ahead & 2 Week Ahead & 1 Week Ahead & 2 Week Ahead \\ 
  \hline
ARGO & \textbf{30.019} & \textbf{46.728} & \textbf{21.430} & \textbf{35.221} & \textbf{0.958} & \textbf{0.910} \\ 
   \hline
AR7 & 31.614 & 53.212 & 23.479 & 40.973 & 0.952 & 0.866 \\ 
   \hline
COVIDhub-ensemble & 32.954 & 51.814 & 24.404 & 39.231 & 0.950 & 0.884 \\ 
   \hline
Naive & 30.921 & 48.101 & 23.404 & 37.904 & 0.952 & 0.888 \\ 
   \hline
MOBS-GLEAM\_COVID & 43.928 & 60.280 & 32.330 & 45.463 & 0.925 & 0.854 \\ 
   \hline
COVIDhub-baseline & 38.770 & 49.583 & 31.731 & 41.154 & 0.925 & 0.882 \\ 
   \hline
GT-DeepCOVID & 48.603 & 70.530 & 36.624 & 55.702 & 0.880 & 0.765 \\ 
   \hline
JHUAPL-Bucky & 63.595 & 112.865 & 39.121 & 64.859 & 0.911 & 0.829 \\ 
   \hline
\end{tabular}
\caption{Comparison of different methods for state-level COVID-19 1 to 2 weeks ahead hospitalizations predictions in New Hampshire (NH). The MSE, MAE, and correlation are reported and best performed method is highlighted in boldface.} 
\end{table}
\begin{figure}[!h] 
  \centering 
\includegraphics[width=.8\linewidth, page=3]{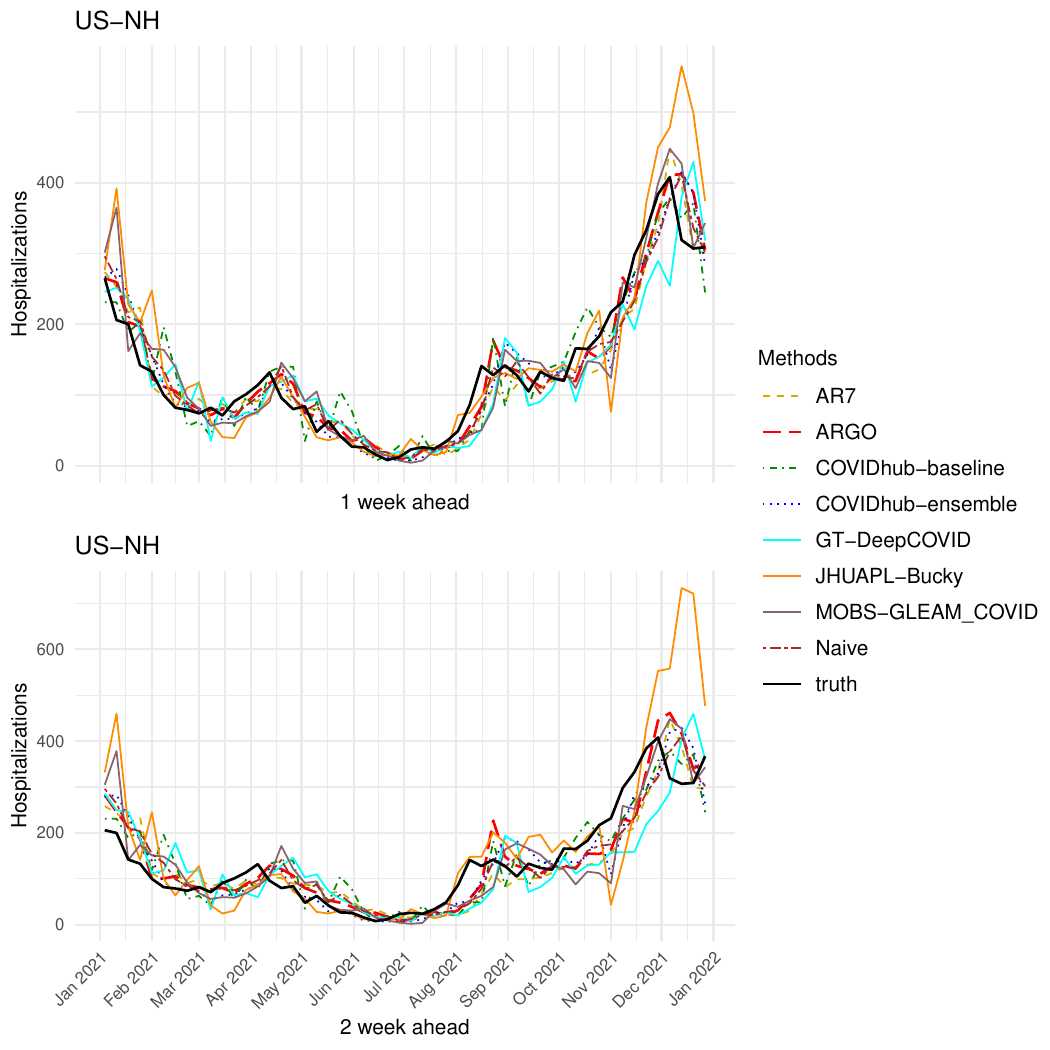} 
\caption{Plots of the COVID-19 hospitalizations 1 week (top), 2 weeks (bottom) ahead estimates of all compared models for New Hampshire (NH).}
\label{fig:State_Ours_NH}
\end{figure}
\newpage   
\begin{table}[ht]
\footnotesize
\centering
\begin{tabular}{|l|r|r|r|r|r|r|}
   \hline\multicolumn{1}{|c}{Methods} & \multicolumn{2}{|c}{RMSE} & \multicolumn{2}{|c}{MAE}& \multicolumn{2}{|c|}{Cor} \\  \cline{2-7} & 1 Week Ahead & 2 Week Ahead & 1 Week Ahead & 2 Week Ahead & 1 Week Ahead & 2 Week Ahead \\ 
  \hline
ARGO & 52.250 & 75.913 & 38.046 & 52.019 & 0.953 & 0.899 \\ 
   \hline
AR7 & 51.600 & 78.175 & 40.400 & 58.499 & 0.949 & 0.879 \\ 
   \hline
COVIDhub-ensemble & 49.051 & \textbf{70.939} & 36.288 & \textbf{50.788} & \textbf{0.959} & \textbf{0.921} \\ 
   \hline
Naive & \textbf{48.420} & 78.998 & \textbf{35.385} & 57.827 & 0.951 & 0.870 \\ 
   \hline
MOBS-GLEAM\_COVID & 59.286 & 85.498 & 42.706 & 63.731 & 0.933 & 0.865 \\ 
   \hline
COVIDhub-baseline & 63.577 & 90.055 & 44.923 & 67.750 & 0.940 & 0.867 \\ 
   \hline
GT-DeepCOVID & 58.170 & 86.678 & 45.178 & 63.402 & 0.942 & 0.884 \\ 
   \hline
JHUAPL-Bucky & 66.075 & 126.709 & 47.948 & 86.908 & 0.954 & 0.920 \\ 
   \hline
\end{tabular}
\caption{Comparison of different methods for state-level COVID-19 1 to 2 weeks ahead hospitalizations predictions in Idaho (ID). The MSE, MAE, and correlation are reported and best performed method is highlighted in boldface.} 
\end{table}
\begin{figure}[!h] 
  \centering 
\includegraphics[width=.8\linewidth, page=4]{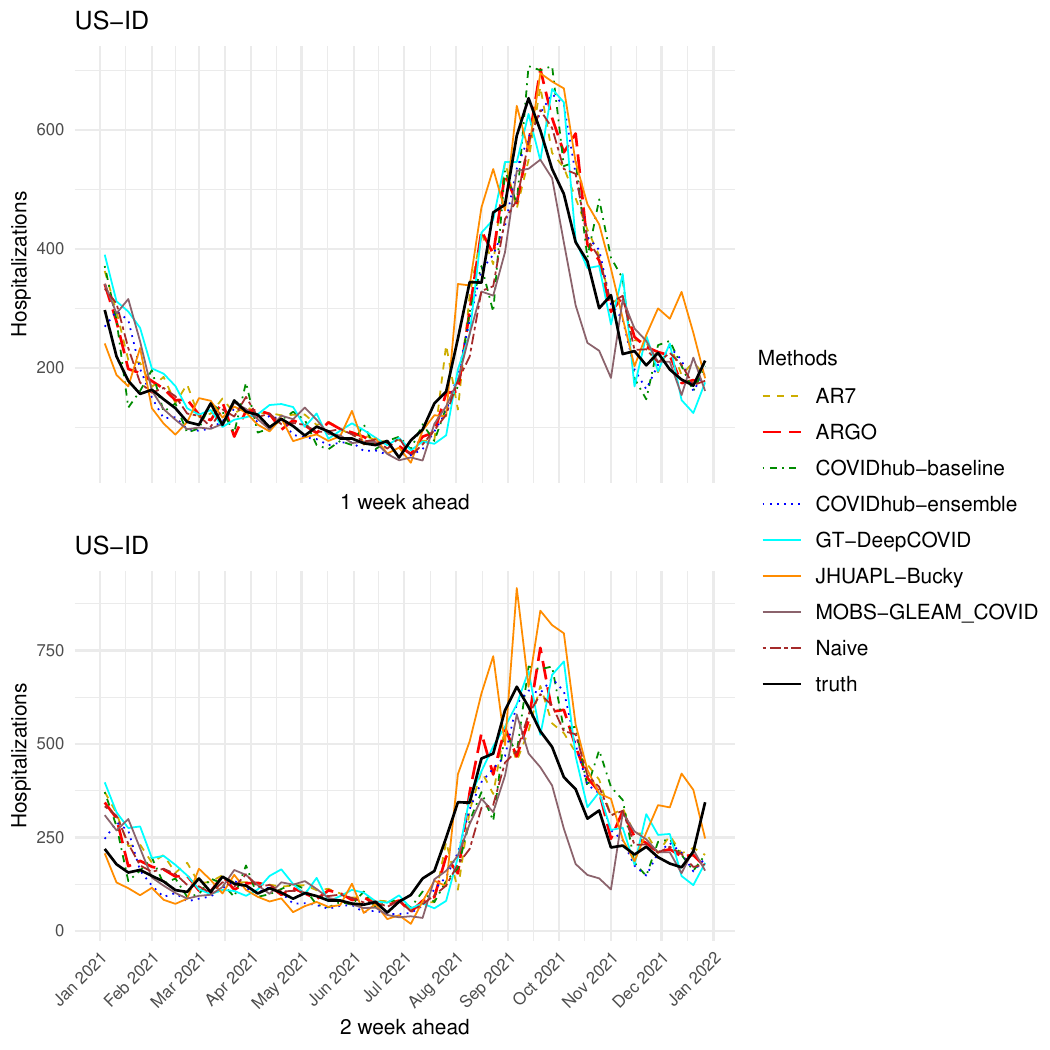} 
\caption{Plots of the COVID-19 hospitalizations 1 week (top), 2 weeks (bottom) ahead estimates of all compared models for New Idaho (ID).}
\end{figure}
\newpage

\begin{table}[ht]
\footnotesize
\centering
\begin{tabular}{|l|r|r|r|r|r|r|}
   \hline\multicolumn{1}{|c}{Methods} & \multicolumn{2}{|c}{RMSE} & \multicolumn{2}{|c}{MAE}& \multicolumn{2}{|c|}{Cor} \\  \cline{2-7} & 1 Week Ahead & 2 Week Ahead & 1 Week Ahead & 2 Week Ahead & 1 Week Ahead & 2 Week Ahead \\ 
  \hline
ARGO & \textbf{92.105} & 228.777 & \textbf{68.979} & \textbf{152.275} & \textbf{0.975} & \textbf{0.846} \\ 
   \hline
AR7 & 98.239 & \textbf{221.348} & 72.824 & 162.710 & \textbf{0.975} & 0.819 \\ 
   \hline
COVIDhub-ensemble & 210.292 & 362.934 & 125.000 & 203.173 & 0.907 & 0.746 \\ 
   \hline
Naive & 143.484 & 272.894 & 96.250 & 180.288 & 0.915 & 0.696 \\ 
   \hline
MOBS-GLEAM\_COVID & 212.609 & 352.741 & 128.139 & 209.235 & 0.927 & 0.819 \\ 
   \hline
COVIDhub-baseline & 166.796 & 286.944 & 107.558 & 181.442 & 0.879 & 0.652 \\ 
   \hline
GT-DeepCOVID & 201.068 & 369.295 & 136.673 & 250.048 & 0.891 & 0.676 \\ 
   \hline
JHUAPL-Bucky & 312.020 & 497.312 & 184.330 & 262.479 & 0.839 & 0.673 \\ 
   \hline
\end{tabular}
\caption{Comparison of different methods for state-level COVID-19 1 to 2 weeks ahead hospitalizations predictions in Mississippi (MS). The MSE, MAE, and correlation are reported and best performed method is highlighted in boldface.} 
\end{table}
\begin{figure}[!h] 
  \centering 
\includegraphics[width=.8\linewidth, page=5]{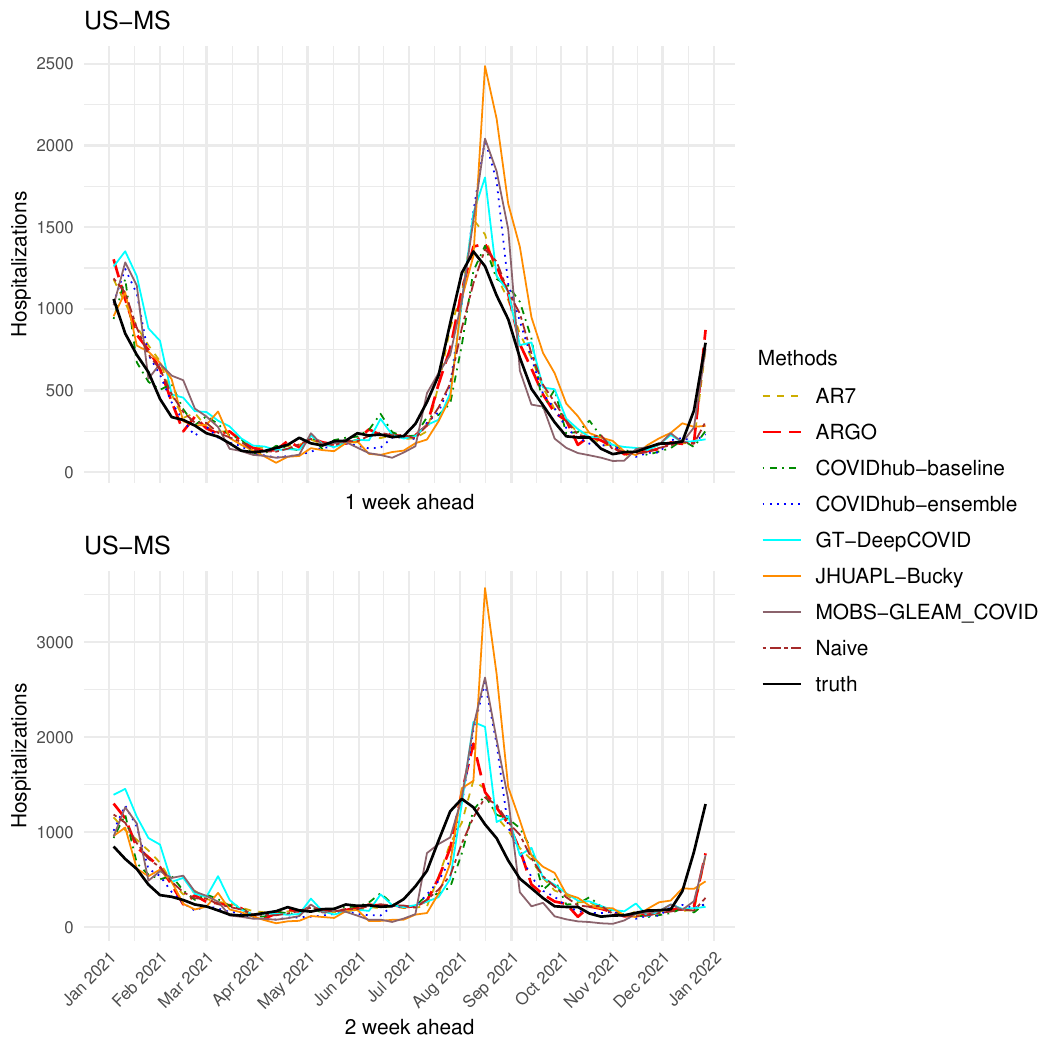} 
\caption{Plots of the COVID-19 hospitalizations 1 week (top), 2 weeks (bottom) ahead estimates of all compared models for New Mississippi (MS).}
\end{figure}
\newpage

\begin{table}[ht]
\footnotesize
\centering
\begin{tabular}{|l|r|r|r|r|r|r|}
   \hline\multicolumn{1}{|c}{Methods} & \multicolumn{2}{|c}{RMSE} & \multicolumn{2}{|c}{MAE}& \multicolumn{2}{|c|}{Cor} \\  \cline{2-7} & 1 Week Ahead & 2 Week Ahead & 1 Week Ahead & 2 Week Ahead & 1 Week Ahead & 2 Week Ahead \\ 
  \hline
ARGO & \textbf{27.871} & \textbf{59.532} & \textbf{19.053} & \textbf{36.003} & \textbf{0.960} & \textbf{0.844} \\ 
   \hline
AR7 & 36.439 & 72.537 & 24.072 & 43.727 & 0.931 & 0.746 \\ 
   \hline
COVIDhub-ensemble & 48.815 & 77.182 & 29.365 & 45.423 & 0.890 & 0.772 \\ 
   \hline
Naive & 45.194 & 78.542 & 28.000 & 46.077 & 0.888 & 0.692 \\ 
   \hline
MOBS-GLEAM\_COVID & 39.931 & 65.546 & 26.723 & 42.401 & 0.915 & 0.796 \\ 
   \hline
COVIDhub-baseline & 87.433 & 109.850 & 46.154 & 61.000 & 0.667 & 0.492 \\ 
   \hline
GT-DeepCOVID & 39.913 & 69.336 & 28.192 & 45.846 & 0.919 & 0.771 \\ 
   \hline
JHUAPL-Bucky & 44.486 & 70.817 & 31.108 & 46.346 & 0.903 & 0.826 \\ 
   \hline
\end{tabular}
\caption{Comparison of different methods for state-level COVID-19 1 to 2 weeks ahead hospitalizations predictions in Hawaii (HI). The MSE, MAE, and correlation are reported and best performed method is highlighted in boldface.} 
\end{table}
\begin{figure}[!h] 
  \centering 
\includegraphics[width=.8\linewidth, page=6]{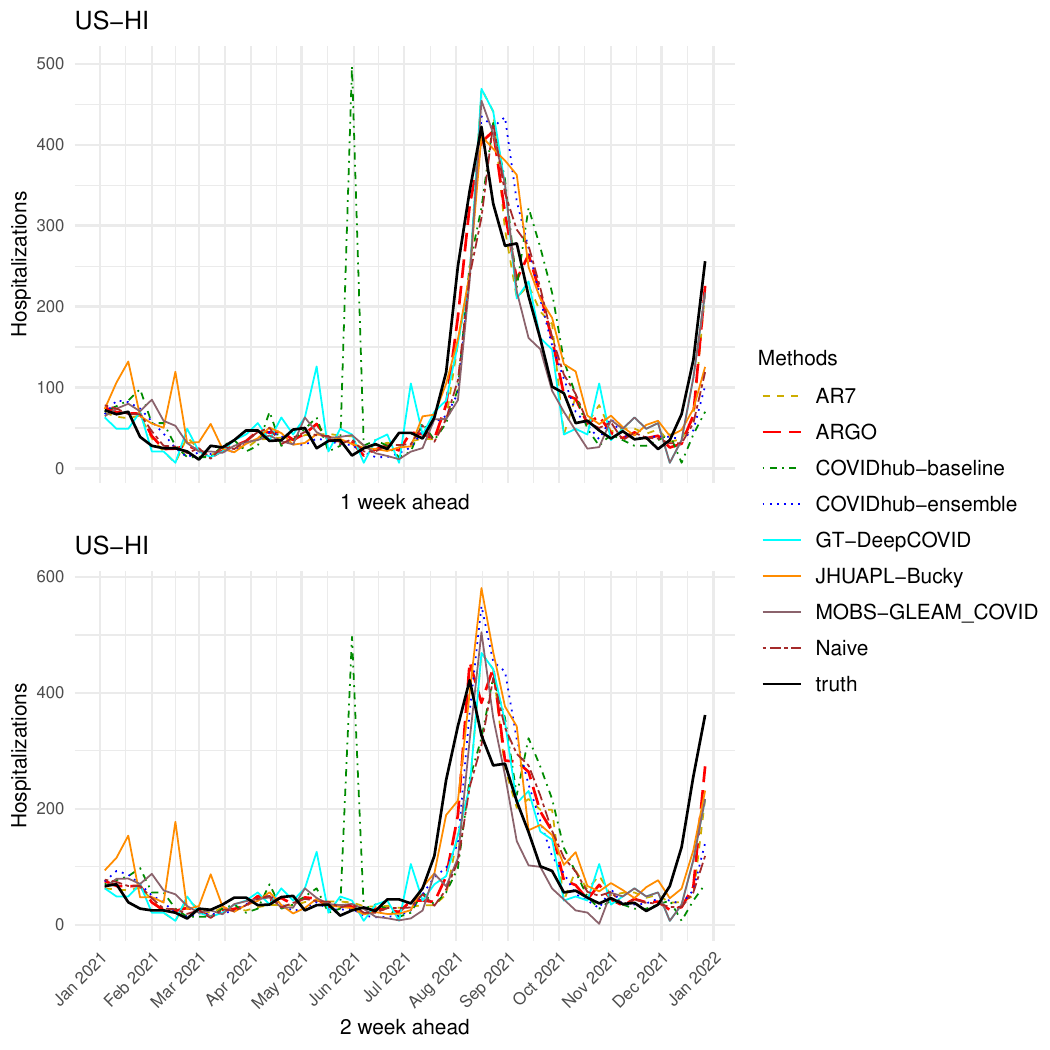} 
\caption{Plots of the COVID-19 hospitalizations 1 week (top), 2 weeks (bottom) ahead estimates of all compared models for New Hawaii (HI).}
\end{figure}
\newpage

\begin{table}[ht]
\footnotesize
\centering
\begin{tabular}{|l|r|r|r|r|r|r|}
   \hline\multicolumn{1}{|c}{Methods} & \multicolumn{2}{|c}{RMSE} & \multicolumn{2}{|c}{MAE}& \multicolumn{2}{|c|}{Cor} \\  \cline{2-7} & 1 Week Ahead & 2 Week Ahead & 1 Week Ahead & 2 Week Ahead & 1 Week Ahead & 2 Week Ahead \\ 
  \hline
ARGO & \textbf{84.596} & 162.298 & \textbf{64.685} & 124.406 & \textbf{0.968} & \textbf{0.904} \\ 
   \hline
AR7 & 90.964 & \textbf{149.828} & 68.958 & 122.481 & 0.961 & 0.898 \\ 
   \hline
COVIDhub-ensemble & 99.377 & 157.419 & 75.885 & \textbf{117.404} & 0.957 & 0.903 \\ 
   \hline
Naive & 93.028 & 155.334 & 75.673 & 129.962 & 0.958 & 0.892 \\ 
   \hline
MOBS-GLEAM\_COVID & 126.640 & 210.765 & 95.340 & 144.185 & 0.936 & 0.836 \\ 
   \hline
COVIDhub-baseline & 132.690 & 176.805 & 101.731 & 144.788 & 0.920 & 0.862 \\ 
   \hline
GT-DeepCOVID & 154.898 & 247.475 & 114.148 & 188.196 & 0.887 & 0.732 \\ 
   \hline
JHUAPL-Bucky & 249.706 & 421.022 & 146.733 & 250.953 & 0.898 & 0.810 \\ 
   \hline
\end{tabular}
\caption{Comparison of different methods for state-level COVID-19 1 to 2 weeks ahead hospitalizations predictions in Minnesota (MN). The MSE, MAE, and correlation are reported and best performed method is highlighted in boldface.} 
\end{table}
\begin{figure}[!h] 
  \centering 
\includegraphics[width=.8\linewidth, page=7]{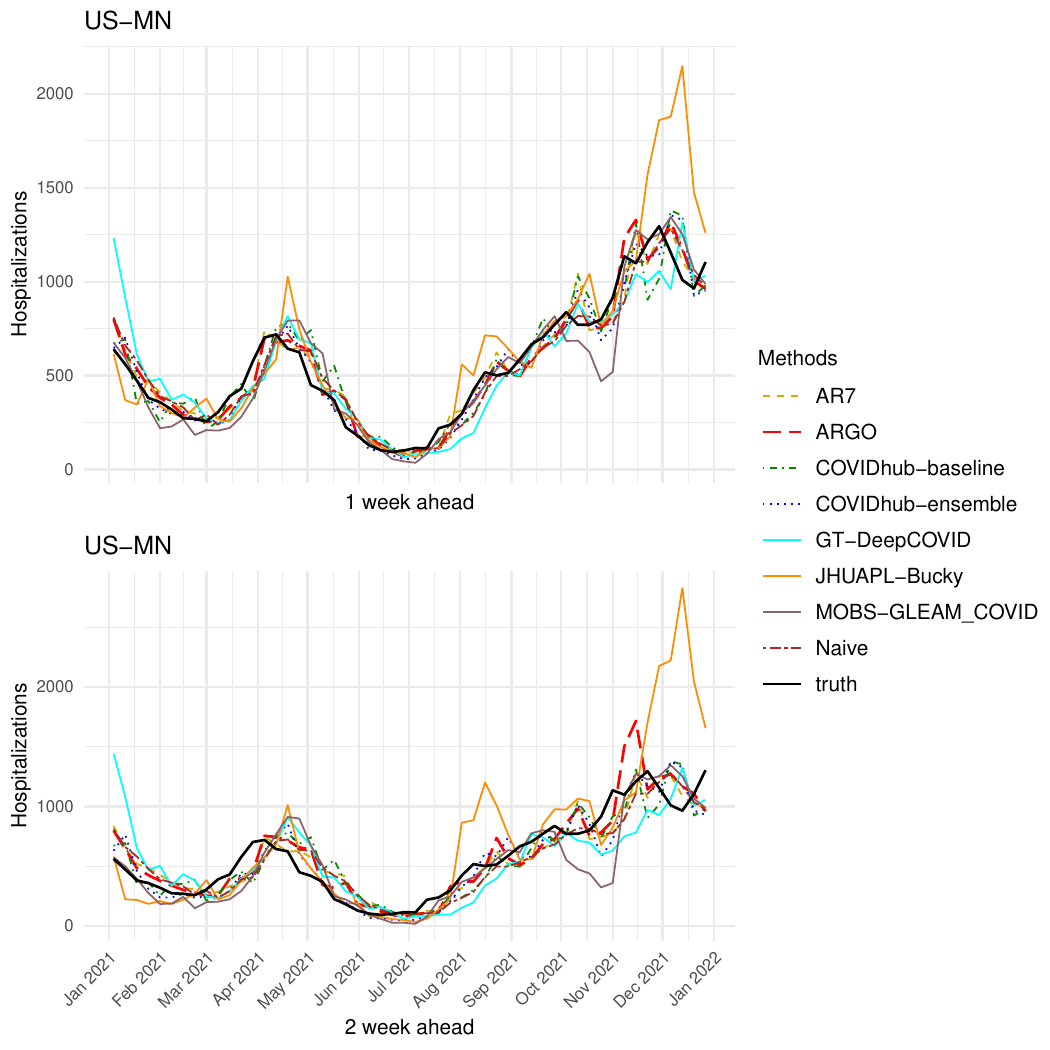} 
\caption{Plots of the COVID-19 hospitalizations 1 week (top), 2 weeks (bottom) ahead estimates of all compared models for New Minnesota (MN).}
\end{figure}
\newpage

\begin{table}[ht]
\footnotesize
\centering
\begin{tabular}{|l|r|r|r|r|r|r|}
   \hline\multicolumn{1}{|c}{Methods} & \multicolumn{2}{|c}{RMSE} & \multicolumn{2}{|c}{MAE}& \multicolumn{2}{|c|}{Cor} \\  \cline{2-7} & 1 Week Ahead & 2 Week Ahead & 1 Week Ahead & 2 Week Ahead & 1 Week Ahead & 2 Week Ahead \\ 
  \hline
ARGO & 27.134 & \textbf{33.196} & \textbf{17.813} & \textbf{25.948} & 0.928 & 0.895 \\ 
   \hline
AR7 & 26.937 & 37.586 & 18.663 & 29.121 & 0.927 & 0.857 \\ 
   \hline
COVIDhub-ensemble & \textbf{26.631} & 36.734 & 20.423 & 26.750 & \textbf{0.943} & \textbf{0.898} \\ 
   \hline
Naive & 27.142 & 38.988 & 19.827 & 28.365 & 0.928 & 0.853 \\ 
   \hline
MOBS-GLEAM\_COVID & 41.629 & 59.711 & 31.201 & 43.233 & 0.872 & 0.767 \\ 
   \hline
COVIDhub-baseline & 37.552 & 50.556 & 26.538 & 35.654 & 0.873 & 0.770 \\ 
   \hline
GT-DeepCOVID & 35.562 & 51.977 & 26.549 & 37.724 & 0.868 & 0.735 \\ 
   \hline
JHUAPL-Bucky & 48.270 & 73.256 & 27.672 & 42.234 & 0.826 & 0.725 \\ 
   \hline
\end{tabular}
\caption{Comparison of different methods for state-level COVID-19 1 to 2 weeks ahead hospitalizations predictions in North Dakota (ND). The MSE, MAE, and correlation are reported and best performed method is highlighted in boldface.} 
\end{table}
\begin{figure}[!h] 
  \centering 
\includegraphics[width=.8\linewidth, page=8]{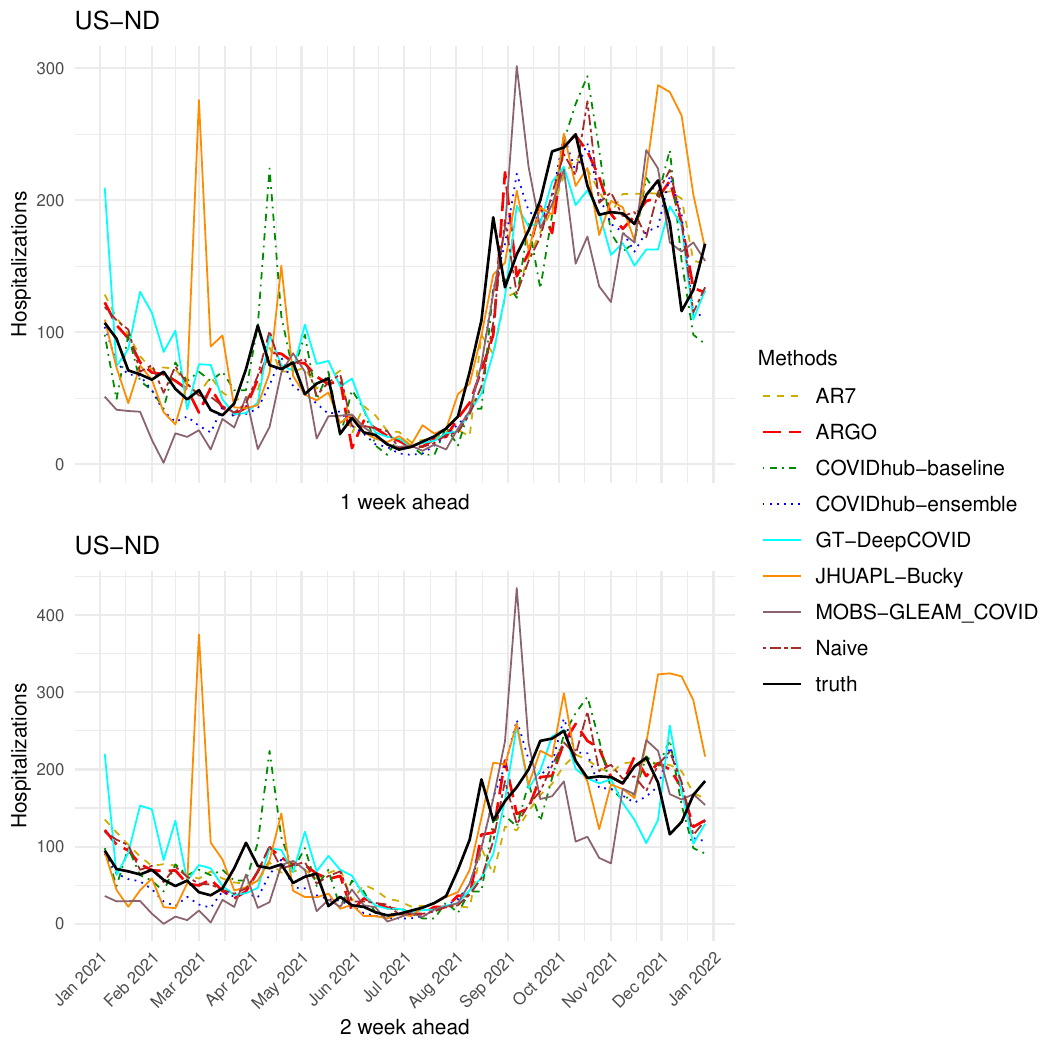} 
\caption{Plots of the COVID-19 hospitalizations 1 week (top), 2 weeks (bottom) ahead estimates of all compared models for New North Dakota (ND).}
\end{figure}
\newpage

\begin{table}[ht]
\footnotesize
\centering
\begin{tabular}{|l|r|r|r|r|r|r|}
   \hline\multicolumn{1}{|c}{Methods} & \multicolumn{2}{|c}{RMSE} & \multicolumn{2}{|c}{MAE}& \multicolumn{2}{|c|}{Cor} \\  \cline{2-7} & 1 Week Ahead & 2 Week Ahead & 1 Week Ahead & 2 Week Ahead & 1 Week Ahead & 2 Week Ahead \\ 
  \hline
ARGO & 52.161 & 90.248 & \textbf{26.781} & 41.176 & \textbf{0.915} & 0.766 \\ 
   \hline
AR7 & \textbf{43.867} & 83.042 & 29.299 & 46.742 & 0.909 & 0.741 \\ 
   \hline
COVIDhub-ensemble & 54.563 & 84.071 & 26.885 & \textbf{41.038} & 0.885 & 0.760 \\ 
   \hline
Naive & 60.218 & 94.871 & 30.596 & 47.558 & 0.834 & 0.655 \\ 
   \hline
MOBS-GLEAM\_COVID & 53.551 & 84.700 & 32.923 & 45.077 & 0.870 & 0.737 \\ 
   \hline
COVIDhub-baseline & 63.271 & 98.794 & 39.096 & 53.250 & 0.802 & 0.608 \\ 
   \hline
GT-DeepCOVID & 78.457 & 107.707 & 38.295 & 57.312 & 0.679 & 0.506 \\ 
   \hline
JHUAPL-Bucky & 53.965 & \textbf{72.660} & 33.434 & 50.459 & 0.863 & \textbf{0.845} \\ 
   \hline
\end{tabular}
\caption{Comparison of different methods for state-level COVID-19 1 to 2 weeks ahead hospitalizations predictions in District of Columbia (DC). The MSE, MAE, and correlation are reported and best performed method is highlighted in boldface.} 
\end{table}
\begin{figure}[!h] 
  \centering 
\includegraphics[width=.8\linewidth, page=9]{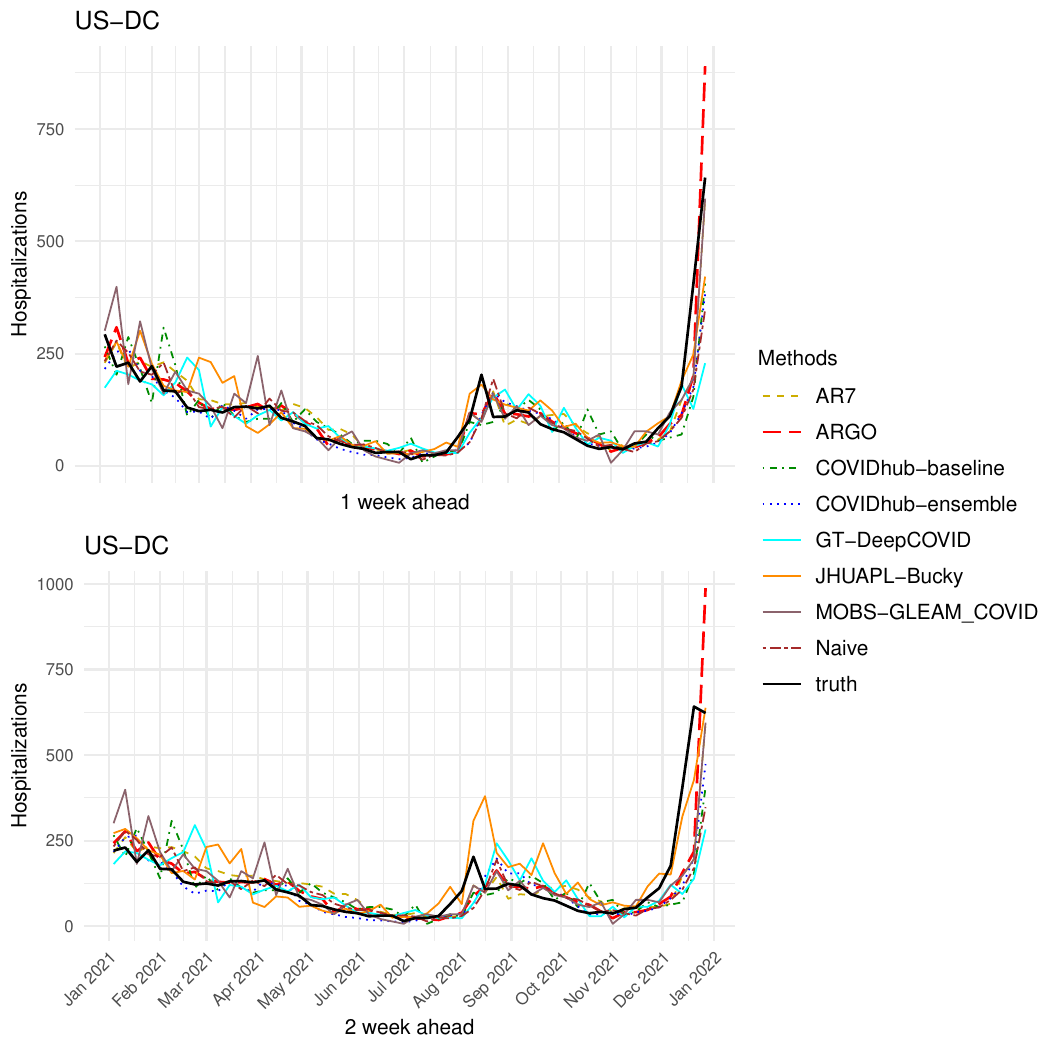} 
\caption{Plots of the COVID-19 hospitalizations 1 week (top), 2 weeks (bottom) ahead estimates of all compared models for New District of Columbia (DC).}
\end{figure}
\newpage

\begin{table}[ht]
\footnotesize
\centering
\begin{tabular}{|l|r|r|r|r|r|r|}
   \hline\multicolumn{1}{|c}{Methods} & \multicolumn{2}{|c}{RMSE} & \multicolumn{2}{|c}{MAE}& \multicolumn{2}{|c|}{Cor} \\  \cline{2-7} & 1 Week Ahead & 2 Week Ahead & 1 Week Ahead & 2 Week Ahead & 1 Week Ahead & 2 Week Ahead \\ 
  \hline
ARGO & \textbf{52.704} & 86.347 & \textbf{39.438} & 61.805 & \textbf{0.965} & 0.898 \\ 
   \hline
AR7 & 59.496 & 101.480 & 42.110 & 67.705 & 0.960 & 0.870 \\ 
   \hline
COVIDhub-ensemble & 55.654 & \textbf{75.835} & 41.750 & \textbf{61.558} & 0.960 & \textbf{0.925} \\ 
   \hline
Naive & 61.445 & 98.872 & 43.788 & 70.981 & 0.951 & 0.867 \\ 
   \hline
MOBS-GLEAM\_COVID & 80.503 & 119.381 & 60.808 & 89.135 & 0.932 & 0.849 \\ 
   \hline
COVIDhub-baseline & 86.486 & 119.014 & 65.481 & 83.942 & 0.904 & 0.811 \\ 
   \hline
GT-DeepCOVID & 91.979 & 121.802 & 66.766 & 87.935 & 0.903 & 0.823 \\ 
   \hline
JHUAPL-Bucky & 130.309 & 190.096 & 82.353 & 115.266 & 0.788 & 0.696 \\ 
   \hline
\end{tabular}
\caption{Comparison of different methods for state-level COVID-19 1 to 2 weeks ahead hospitalizations predictions in New Mexico (NM). The MSE, MAE, and correlation are reported and best performed method is highlighted in boldface.} 
\end{table}
\begin{figure}[!h] 
  \centering 
\includegraphics[width=.8\linewidth, page=10]{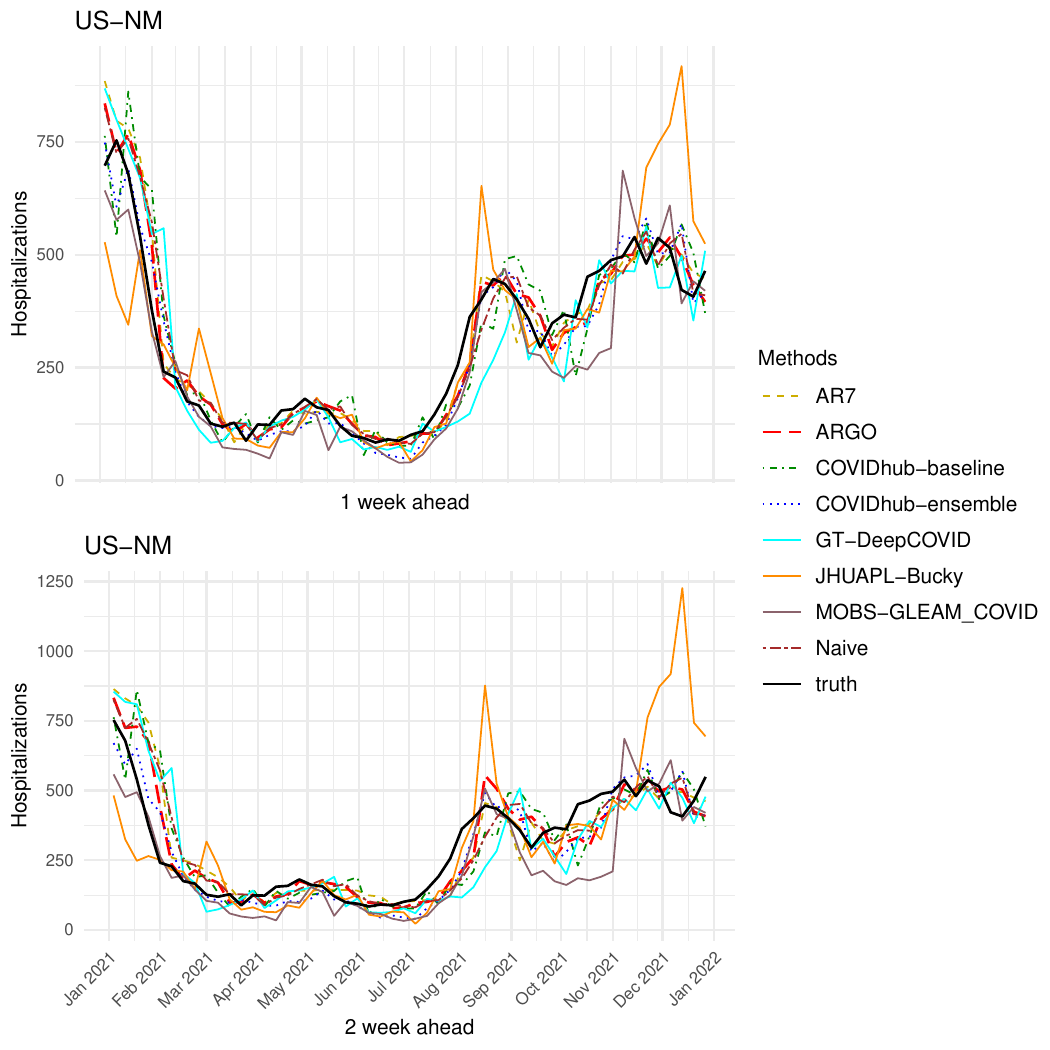} 
\caption{Plots of the COVID-19 hospitalizations 1 week (top), 2 weeks (bottom) ahead estimates of all compared models for New New Mexico (NM).}
\end{figure}
\newpage

\begin{table}[ht]
\footnotesize
\centering
\begin{tabular}{|l|r|r|r|r|r|r|}
   \hline\multicolumn{1}{|c}{Methods} & \multicolumn{2}{|c}{RMSE} & \multicolumn{2}{|c}{MAE}& \multicolumn{2}{|c|}{Cor} \\  \cline{2-7} & 1 Week Ahead & 2 Week Ahead & 1 Week Ahead & 2 Week Ahead & 1 Week Ahead & 2 Week Ahead \\ 
  \hline
ARGO & \textbf{292.893} & 572.129 & \textbf{173.725} & \textbf{345.365} & \textbf{0.958} & 0.845 \\ 
   \hline
AR7 & 320.133 & 623.691 & 205.600 & 439.986 & 0.951 & 0.801 \\ 
   \hline
COVIDhub-ensemble & 384.678 & \textbf{559.940} & 232.000 & 370.038 & 0.938 & \textbf{0.882} \\ 
   \hline
Naive & 397.855 & 671.008 & 253.673 & 453.885 & 0.916 & 0.754 \\ 
   \hline
MOBS-GLEAM\_COVID & 503.154 & 602.235 & 311.804 & 397.714 & 0.903 & 0.867 \\ 
   \hline
COVIDhub-baseline & 480.136 & 696.934 & 293.904 & 466.615 & 0.887 & 0.752 \\ 
   \hline
GT-DeepCOVID & 425.103 & 771.140 & 289.004 & 518.862 & 0.935 & 0.819 \\ 
   \hline
JHUAPL-Bucky & 881.825 & 889.419 & 618.246 & 659.238 & 0.731 & 0.678 \\ 
   \hline
\end{tabular}
\caption{Comparison of different methods for state-level COVID-19 1 to 2 weeks ahead hospitalizations predictions in Kentucky (KY). The MSE, MAE, and correlation are reported and best performed method is highlighted in boldface.} 
\end{table}
\begin{figure}[!h] 
  \centering 
\includegraphics[width=.8\linewidth, page=11]{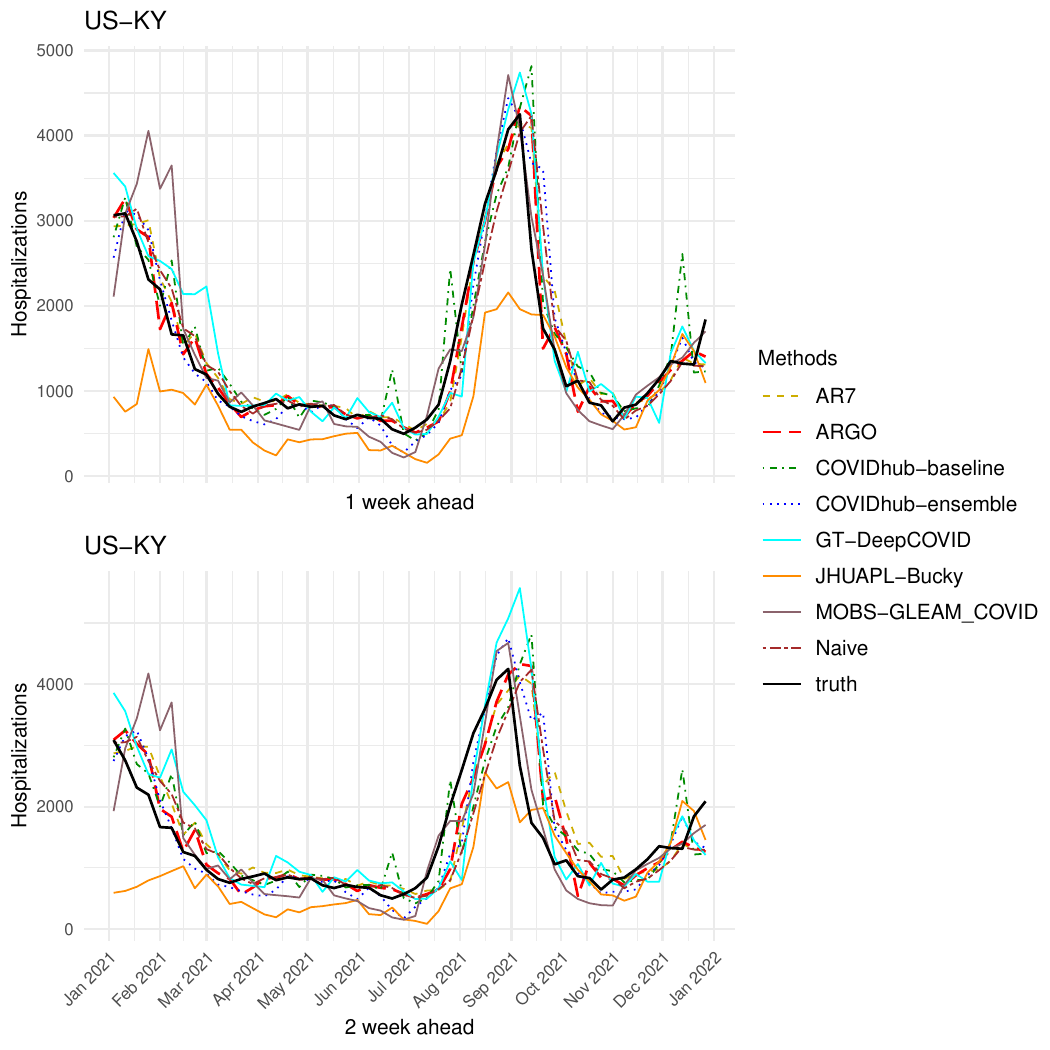} 
\caption{Plots of the COVID-19 hospitalizations 1 week (top), 2 weeks (bottom) ahead estimates of all compared models for New Kentucky (KY). }
\end{figure}
\newpage 
\begin{table}[ht]
\footnotesize
\centering
\begin{tabular}{|l|r|r|r|r|r|r|}
   \hline\multicolumn{1}{|c}{Methods} & \multicolumn{2}{|c}{RMSE} & \multicolumn{2}{|c}{MAE}& \multicolumn{2}{|c|}{Cor} \\  \cline{2-7} & 1 Week Ahead & 2 Week Ahead & 1 Week Ahead & 2 Week Ahead & 1 Week Ahead & 2 Week Ahead \\ 
  \hline
ARGO & \textbf{26.297} & 43.293 & \textbf{16.800} & 26.303 & \textbf{0.923} & 0.823 \\ 
   \hline
AR7 & 33.210 & 52.850 & 20.324 & 34.080 & 0.876 & 0.723 \\ 
   \hline
COVIDhub-ensemble & 30.591 & \textbf{32.607} & 18.962 & \textbf{22.385} & 0.899 & \textbf{0.905} \\ 
   \hline
Naive & 31.985 & 46.681 & 19.673 & 29.731 & 0.884 & 0.789 \\ 
   \hline
MOBS-GLEAM\_COVID & 30.318 & 44.141 & 21.517 & 29.541 & 0.915 & 0.832 \\ 
   \hline
COVIDhub-baseline & 156.728 & 155.376 & 48.212 & 52.962 & 0.258 & 0.289 \\ 
   \hline
GT-DeepCOVID & 43.020 & 56.805 & 29.981 & 40.554 & 0.802 & 0.712 \\ 
   \hline
JHUAPL-Bucky & 57.462 & 89.356 & 43.875 & 59.223 & 0.913 & 0.865 \\ 
   \hline
\end{tabular}
\caption{Comparison of different methods for state-level COVID-19 1 to 2 weeks ahead hospitalizations predictions in Rhode Island (RI). The MSE, MAE, and correlation are reported and best performed method is highlighted in boldface.} 
\end{table}
\begin{figure}[!h] 
  \centering 
\includegraphics[width=.8\linewidth, page=12]{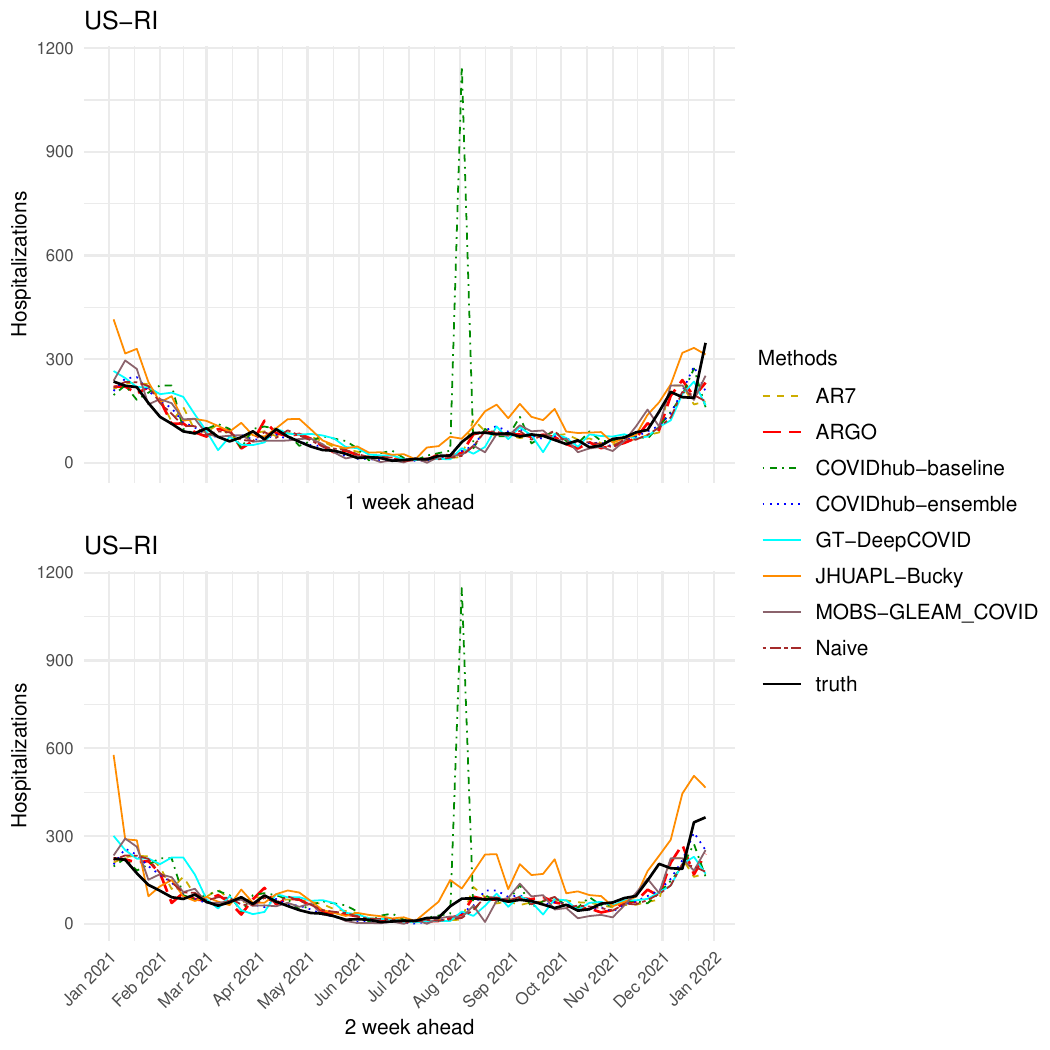} 
\caption{Plots of the COVID-19 hospitalizations 1 week (top), 2 weeks (bottom) ahead estimates of all compared models for New Rhode Island (RI). }
\end{figure}
\newpage 
\begin{table}[ht]
\footnotesize
\centering
\begin{tabular}{|l|r|r|r|r|r|r|}
   \hline\multicolumn{1}{|c}{Methods} & \multicolumn{2}{|c}{RMSE} & \multicolumn{2}{|c}{MAE}& \multicolumn{2}{|c|}{Cor} \\  \cline{2-7} & 1 Week Ahead & 2 Week Ahead & 1 Week Ahead & 2 Week Ahead & 1 Week Ahead & 2 Week Ahead \\ 
  \hline
ARGO & \textbf{55.085} & \textbf{105.777} & \textbf{44.320} & \textbf{77.583} & \textbf{0.976} & \textbf{0.918} \\ 
   \hline
AR7 & 72.486 & 135.293 & 56.607 & 100.931 & 0.959 & 0.859 \\ 
   \hline
COVIDhub-ensemble & 83.651 & 127.510 & 63.231 & 96.231 & 0.954 & 0.898 \\ 
   \hline
Naive & 77.737 & 138.919 & 60.135 & 104.558 & 0.953 & 0.853 \\ 
   \hline
MOBS-GLEAM\_COVID & 136.195 & 178.262 & 92.409 & 133.542 & 0.918 & 0.836 \\ 
   \hline
COVIDhub-baseline & 99.115 & 146.553 & 80.865 & 112.135 & 0.921 & 0.832 \\ 
   \hline
GT-DeepCOVID & 126.739 & 201.046 & 98.453 & 150.733 & 0.926 & 0.804 \\ 
   \hline
JHUAPL-Bucky & 143.570 & 274.272 & 98.404 & 172.571 & 0.875 & 0.777 \\ 
   \hline
\end{tabular}
\caption{Comparison of different methods for state-level COVID-19 1 to 2 weeks ahead hospitalizations predictions in Kansas (KS). The MSE, MAE, and correlation are reported and best performed method is highlighted in boldface.} 
\end{table}
\begin{figure}[!h] 
  \centering 
\includegraphics[width=.8\linewidth, page=13]{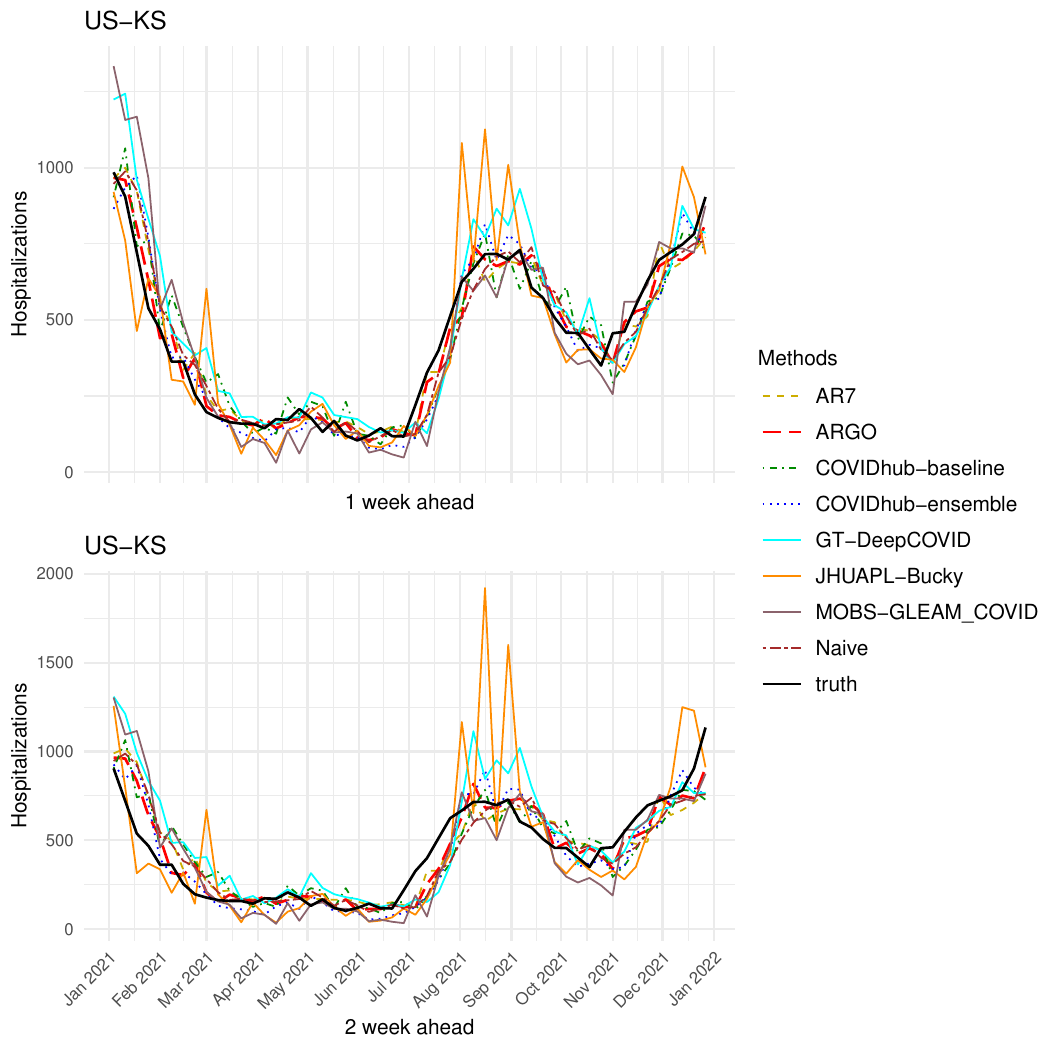} 
\caption{Plots of the COVID-19 hospitalizations 1 week (top), 2 weeks (bottom) ahead estimates of all compared models for New Kansas (KS). }
\end{figure}
\newpage 
\begin{table}[ht]
\footnotesize
\centering
\begin{tabular}{|l|r|r|r|r|r|r|}
   \hline\multicolumn{1}{|c}{Methods} & \multicolumn{2}{|c}{RMSE} & \multicolumn{2}{|c}{MAE}& \multicolumn{2}{|c|}{Cor} \\  \cline{2-7} & 1 Week Ahead & 2 Week Ahead & 1 Week Ahead & 2 Week Ahead & 1 Week Ahead & 2 Week Ahead \\ 
  \hline
ARGO & 29.409 & 40.961 & \textbf{18.659} & 29.301 & 0.919 & 0.855 \\ 
   \hline
AR7 & 27.686 & 40.538 & 21.209 & 32.168 & 0.927 & 0.848 \\ 
   \hline
COVIDhub-ensemble & 33.475 & 41.036 & 23.115 & 30.019 & 0.908 & 0.876 \\ 
   \hline
Naive & \textbf{26.823} & \textbf{36.228} & 18.962 & \textbf{28.115} & \textbf{0.932} & \textbf{0.881} \\ 
   \hline
MOBS-GLEAM\_COVID & 72.103 & 89.302 & 43.313 & 51.992 & 0.701 & 0.577 \\ 
   \hline
COVIDhub-baseline & 171.870 & 172.841 & 56.981 & 63.788 & 0.235 & 0.224 \\ 
   \hline
GT-DeepCOVID & 36.435 & 52.665 & 27.316 & 38.952 & 0.870 & 0.736 \\ 
   \hline
JHUAPL-Bucky & 69.894 & 116.933 & 43.682 & 66.638 & 0.812 & 0.765 \\ 
   \hline
\end{tabular}
\caption{Comparison of different methods for state-level COVID-19 1 to 2 weeks ahead hospitalizations predictions in Maine (ME). The MSE, MAE, and correlation are reported and best performed method is highlighted in boldface.} 
\end{table}
\begin{figure}[!h] 
  \centering 
\includegraphics[width=.8\linewidth, page=14]{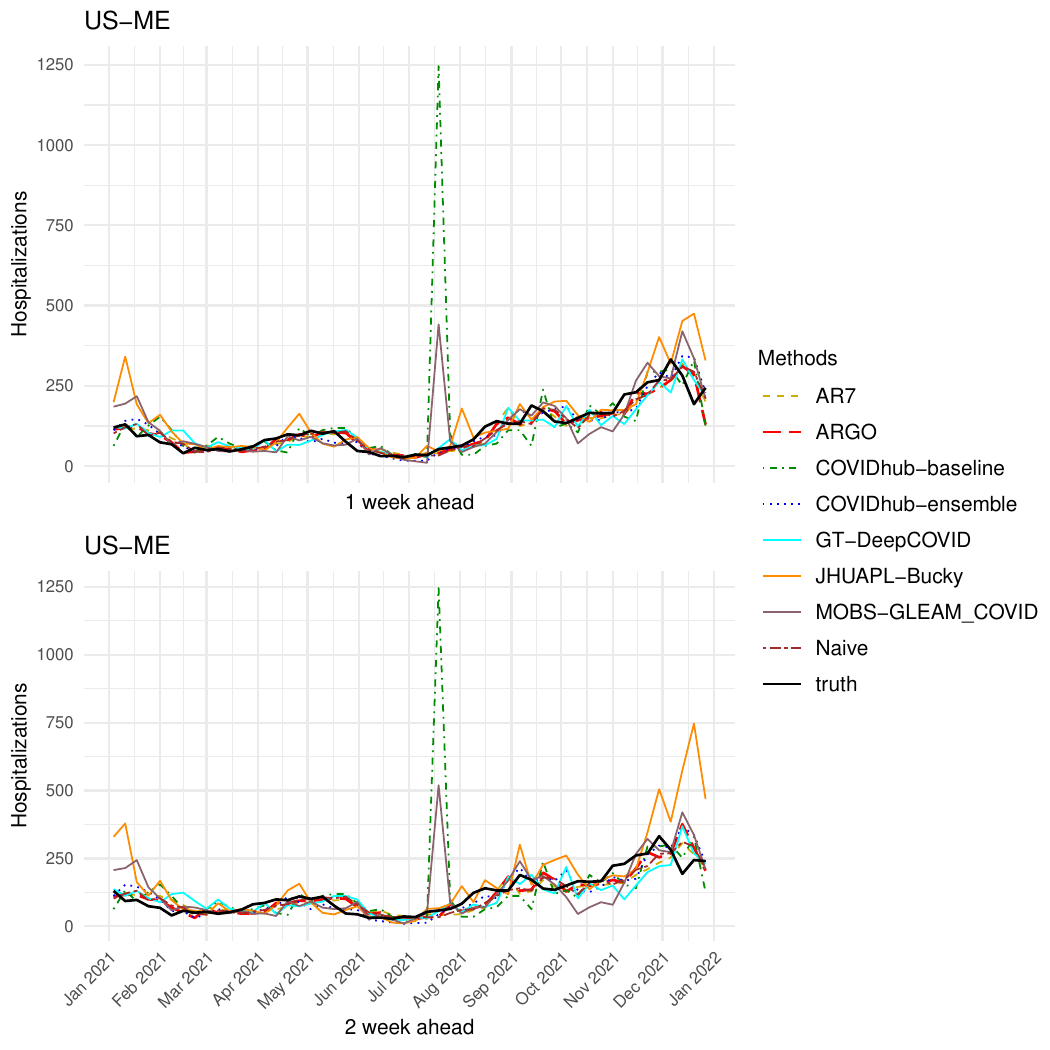} 
\caption{Plots of the COVID-19 hospitalizations 1 week (top), 2 weeks (bottom) ahead estimates of all compared models for New Maine (ME). }
\end{figure}
\newpage 
\begin{table}[ht]
\footnotesize
\centering
\begin{tabular}{|l|r|r|r|r|r|r|}
   \hline\multicolumn{1}{|c}{Methods} & \multicolumn{2}{|c}{RMSE} & \multicolumn{2}{|c}{MAE}& \multicolumn{2}{|c|}{Cor} \\  \cline{2-7} & 1 Week Ahead & 2 Week Ahead & 1 Week Ahead & 2 Week Ahead & 1 Week Ahead & 2 Week Ahead \\ 
  \hline
ARGO & \textbf{41.129} & \textbf{86.389} & \textbf{33.519} & \textbf{65.204} & \textbf{0.984} & \textbf{0.934} \\ 
   \hline
AR7 & 55.254 & 101.240 & 43.556 & 76.814 & 0.970 & 0.908 \\ 
   \hline
COVIDhub-ensemble & 60.696 & 102.870 & 50.000 & 79.885 & 0.972 & 0.926 \\ 
   \hline
Naive & 54.954 & 101.728 & 44.154 & 79.442 & 0.969 & 0.910 \\ 
   \hline
MOBS-GLEAM\_COVID & 78.540 & 114.548 & 63.017 & 87.265 & 0.957 & 0.908 \\ 
   \hline
COVIDhub-baseline & 73.244 & 109.439 & 58.154 & 82.981 & 0.948 & 0.893 \\ 
   \hline
GT-DeepCOVID & 106.453 & 181.291 & 85.626 & 132.804 & 0.904 & 0.771 \\ 
   \hline
JHUAPL-Bucky & 183.416 & 200.434 & 89.515 & 141.903 & 0.748 & 0.810 \\ 
   \hline
\end{tabular}
\caption{Comparison of different methods for state-level COVID-19 1 to 2 weeks ahead hospitalizations predictions in Iowa (IA). The MSE, MAE, and correlation are reported and best performed method is highlighted in boldface.} 
\end{table}
\begin{figure}[!h] 
  \centering 
\includegraphics[width=.8\linewidth, page=15]{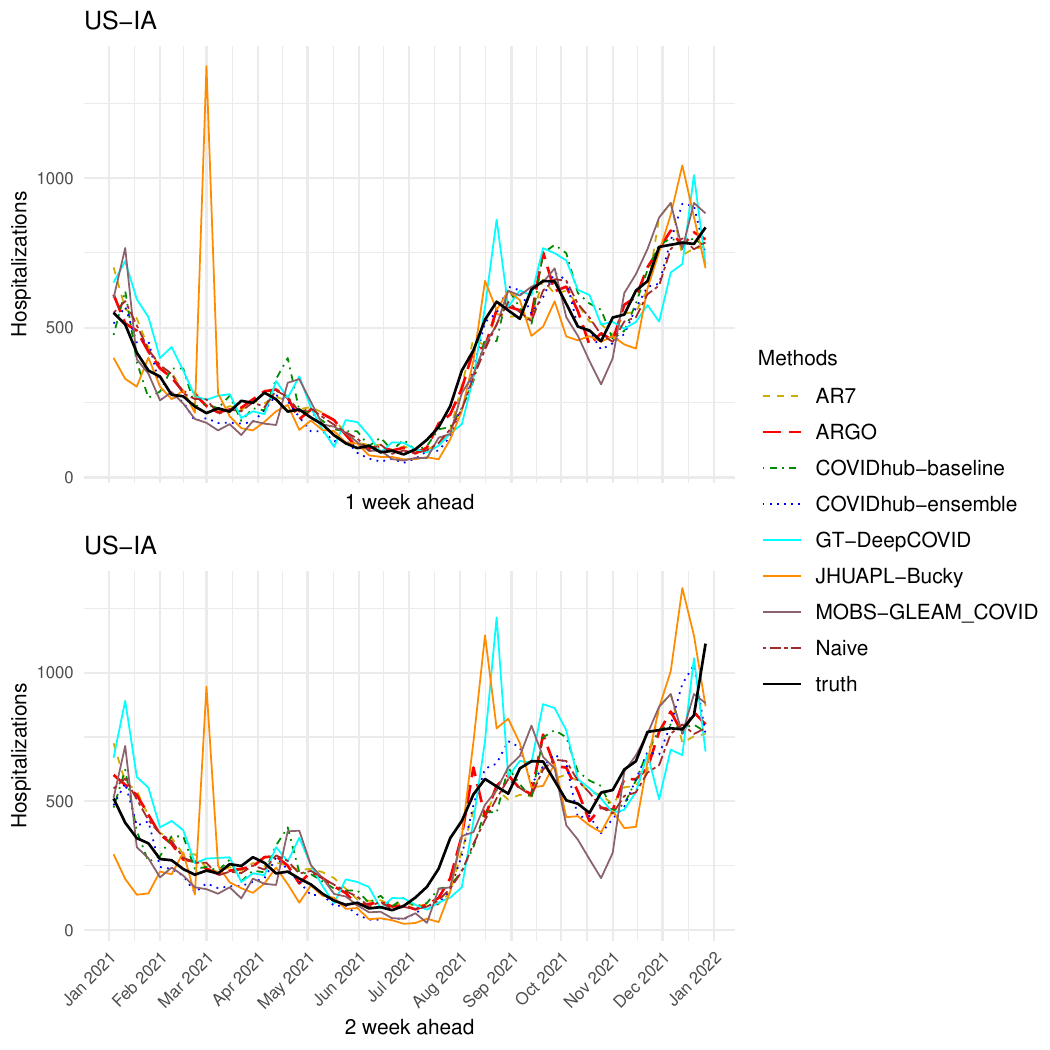} 
\caption{Plots of the COVID-19 hospitalizations 1 week (top), 2 weeks (bottom) ahead estimates of all compared models for New Iowa (IA).  }
\end{figure}
\newpage 
\begin{table}[ht]
\footnotesize
\centering
\begin{tabular}{|l|r|r|r|r|r|r|}
   \hline\multicolumn{1}{|c}{Methods} & \multicolumn{2}{|c}{RMSE} & \multicolumn{2}{|c}{MAE}& \multicolumn{2}{|c|}{Cor} \\  \cline{2-7} & 1 Week Ahead & 2 Week Ahead & 1 Week Ahead & 2 Week Ahead & 1 Week Ahead & 2 Week Ahead \\ 
  \hline
ARGO & \textbf{24.851} & 48.109 & \textbf{19.009} & 34.219 & \textbf{0.957} & 0.850 \\ 
   \hline
AR7 & 29.183 & 46.975 & 21.934 & 35.373 & 0.935 & 0.840 \\ 
   \hline
COVIDhub-ensemble & 36.679 & 55.077 & 25.288 & 35.712 & 0.916 & 0.843 \\ 
   \hline
Naive & 29.893 & \textbf{43.140} & 22.962 & \textbf{30.808} & 0.933 & \textbf{0.869} \\ 
   \hline
MOBS-GLEAM\_COVID & 56.555 & 84.251 & 39.035 & 53.683 & 0.858 & 0.730 \\ 
   \hline
COVIDhub-baseline & 37.925 & 48.689 & 26.923 & 35.846 & 0.909 & 0.849 \\ 
   \hline
GT-DeepCOVID & 35.867 & 55.116 & 27.970 & 42.391 & 0.902 & 0.787 \\ 
   \hline
JHUAPL-Bucky & 40.575 & 70.234 & 28.860 & 45.971 & 0.885 & 0.777 \\ 
   \hline
\end{tabular}
\caption{Comparison of different methods for state-level COVID-19 1 to 2 weeks ahead hospitalizations predictions in South Dakota (SD). The MSE, MAE, and correlation are reported and best performed method is highlighted in boldface.} 
\end{table}
\begin{figure}[!h] 
  \centering 
\includegraphics[width=.8\linewidth, page=16]{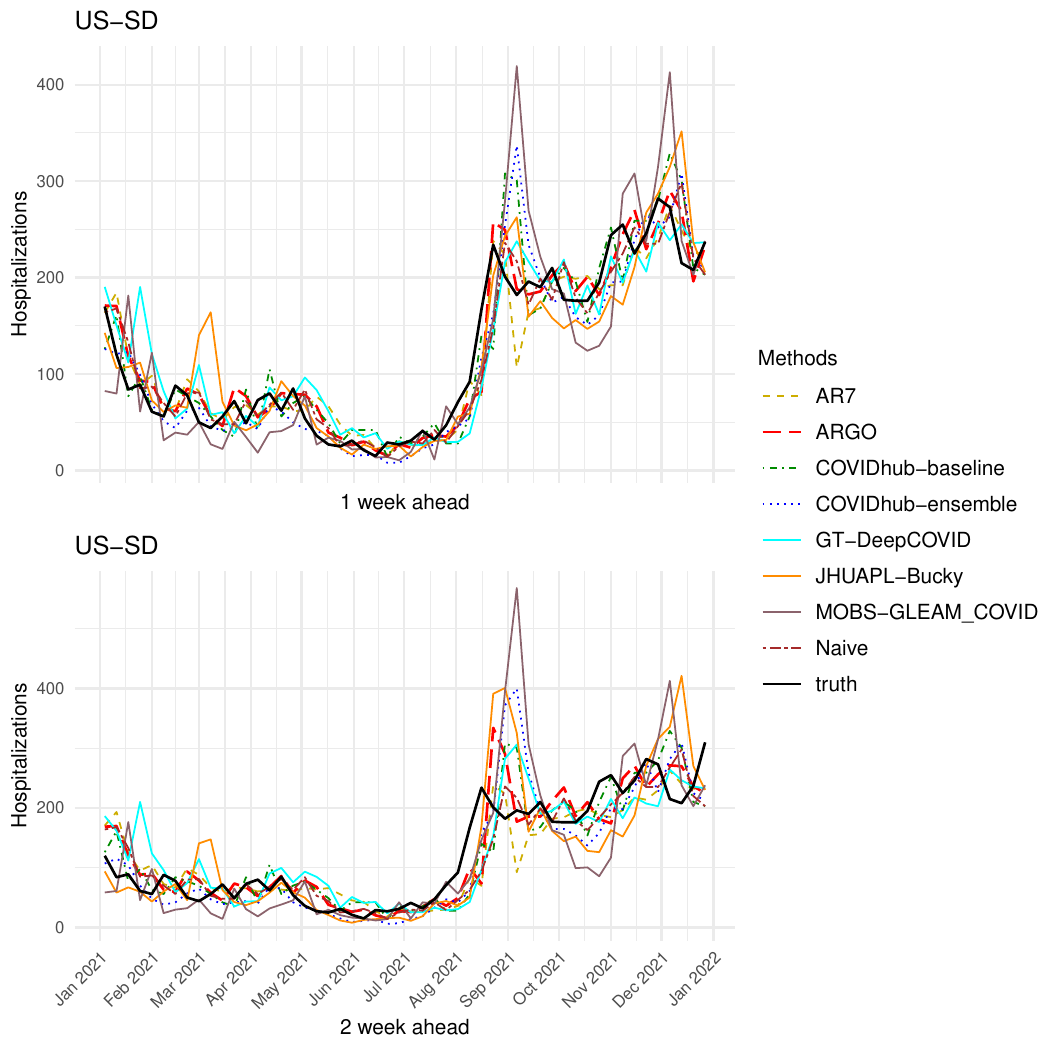} 
\caption{Plots of the COVID-19 hospitalizations 1 week (top), 2 weeks (bottom) ahead estimates of all compared models for New South Dakota (SD).  }
\end{figure}
\newpage 
\begin{table}[ht]
\footnotesize
\centering
\begin{tabular}{|l|r|r|r|r|r|r|}
   \hline\multicolumn{1}{|c}{Methods} & \multicolumn{2}{|c}{RMSE} & \multicolumn{2}{|c}{MAE}& \multicolumn{2}{|c|}{Cor} \\  \cline{2-7} & 1 Week Ahead & 2 Week Ahead & 1 Week Ahead & 2 Week Ahead & 1 Week Ahead & 2 Week Ahead \\ 
  \hline
ARGO & \textbf{155.610} & \textbf{327.579} & \textbf{112.695} & \textbf{241.830} & \textbf{0.981} & \textbf{0.920} \\ 
   \hline
AR7 & 181.258 & 351.439 & 137.329 & 267.276 & 0.972 & 0.879 \\ 
   \hline
COVIDhub-ensemble & 220.531 & 365.955 & 155.288 & 256.962 & 0.967 & 0.909 \\ 
   \hline
Naive & 221.351 & 400.950 & 161.058 & 296.923 & 0.955 & 0.845 \\ 
   \hline
MOBS-GLEAM\_COVID & 391.987 & 513.663 & 235.409 & 348.162 & 0.924 & 0.869 \\ 
   \hline
COVIDhub-baseline & 225.083 & 397.098 & 172.481 & 303.385 & 0.952 & 0.844 \\ 
   \hline
GT-DeepCOVID & 315.873 & 530.474 & 230.539 & 382.922 & 0.947 & 0.852 \\ 
   \hline
JHUAPL-Bucky & 461.762 & 484.641 & 316.105 & 338.143 & 0.876 & 0.808 \\ 
   \hline
\end{tabular}
\caption{Comparison of different methods for state-level COVID-19 1 to 2 weeks ahead hospitalizations predictions in Oklahoma (OK). The MSE, MAE, and correlation are reported and best performed method is highlighted in boldface.} 
\end{table}
\begin{figure}[!h] 
  \centering 
\includegraphics[width=.8\linewidth, page=17]{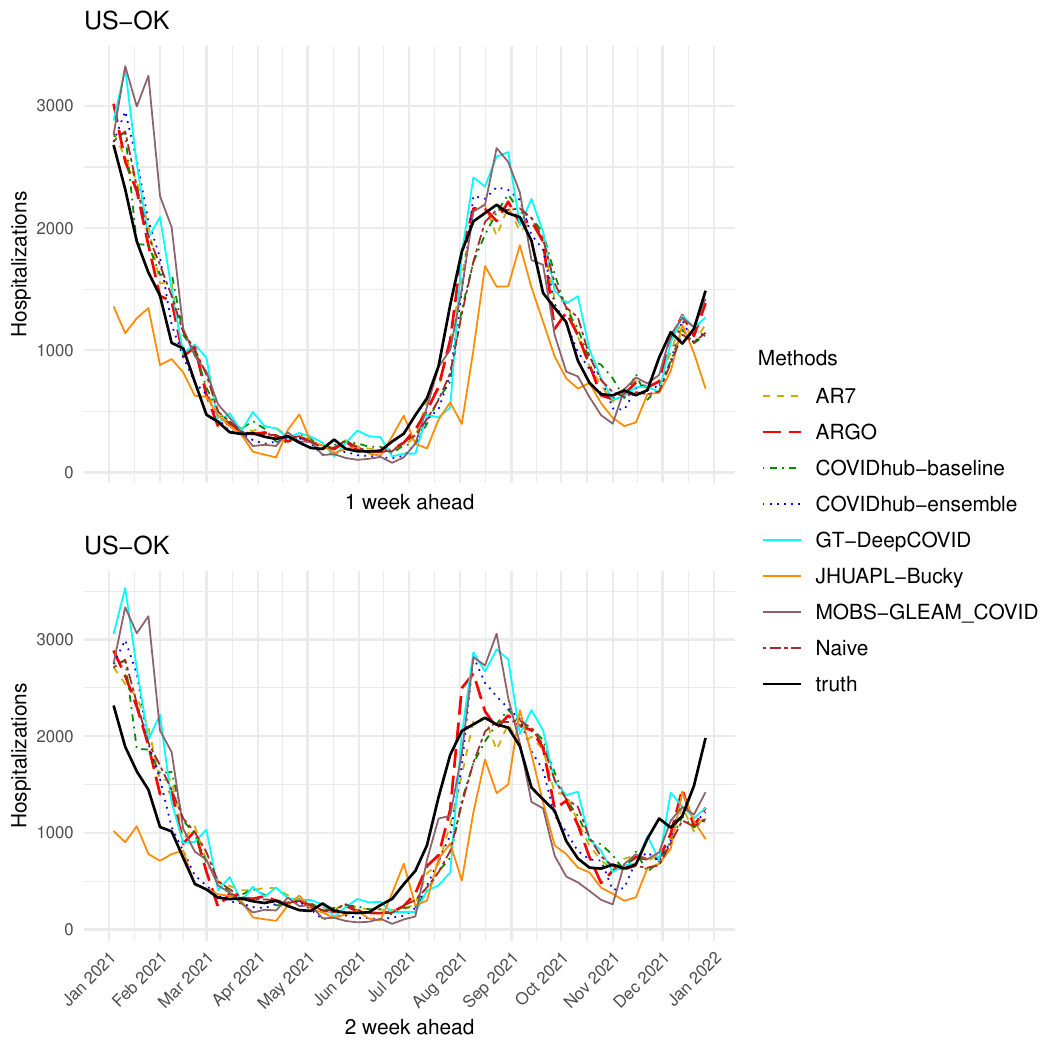} 
\caption{Plots of the COVID-19 hospitalizations 1 week (top), 2 weeks (bottom) ahead estimates of all compared models for New Oklahoma (OK).  }
\end{figure}
\newpage
\begin{table}[ht]
\footnotesize
\centering
\begin{tabular}{|l|r|r|r|r|r|r|}
   \hline\multicolumn{1}{|c}{Methods} & \multicolumn{2}{|c}{RMSE} & \multicolumn{2}{|c}{MAE}& \multicolumn{2}{|c|}{Cor} \\  \cline{2-7} & 1 Week Ahead & 2 Week Ahead & 1 Week Ahead & 2 Week Ahead & 1 Week Ahead & 2 Week Ahead \\ 
  \hline
ARGO & \textbf{405.668} & \textbf{837.908} & 291.744 & \textbf{572.543} & \textbf{0.971} & \textbf{0.878} \\ 
   \hline
AR7 & 421.467 & 917.717 & \textbf{288.420} & 646.876 & 0.965 & 0.830 \\ 
   \hline
COVIDhub-ensemble & 667.843 & 1073.718 & 445.942 & 688.173 & 0.942 & 0.860 \\ 
   \hline
Naive & 635.860 & 1145.531 & 431.481 & 789.596 & 0.920 & 0.741 \\ 
   \hline
MOBS-GLEAM\_COVID & 806.039 & 1428.043 & 498.694 & 882.558 & 0.939 & 0.844 \\ 
   \hline
COVIDhub-baseline & 950.898 & 1298.161 & 778.288 & 1073.135 & 0.892 & 0.750 \\ 
   \hline
GT-DeepCOVID & 712.933 & 1132.578 & 528.956 & 849.849 & 0.914 & 0.798 \\ 
   \hline
JHUAPL-Bucky & 977.750 & 1247.234 & 654.741 & 925.129 & 0.822 & 0.732 \\ 
   \hline
\end{tabular}
\caption{Comparison of different methods for state-level COVID-19 1 to 2 weeks ahead hospitalizations predictions in Georgia (GA). The MSE, MAE, and correlation are reported and best performed method is highlighted in boldface.} 
\end{table}
\begin{figure}[!h] 
  \centering 
\includegraphics[width=.8\linewidth, page=18]{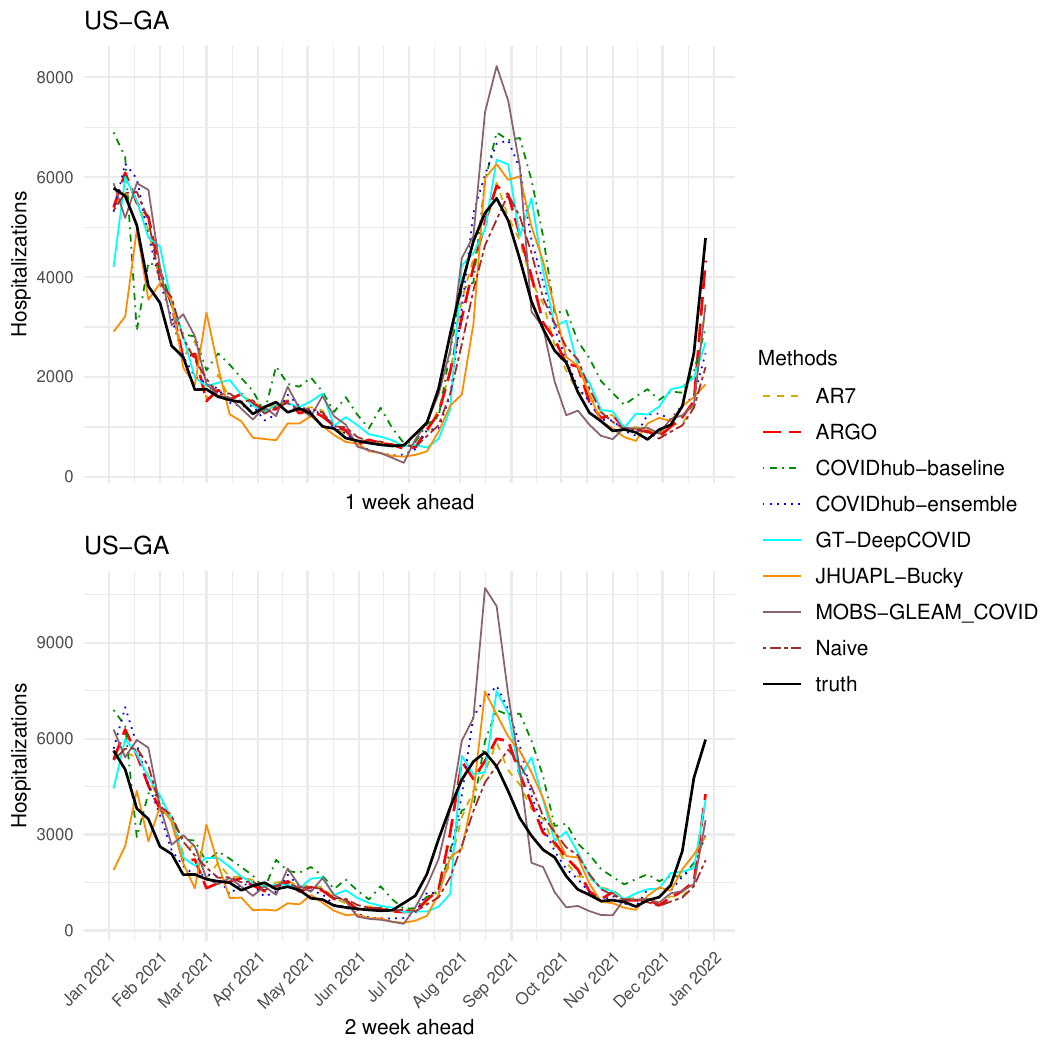} 
\caption{Plots of the COVID-19 hospitalizations 1 week (top), 2 weeks (bottom) ahead estimates of all compared models for New Georgia (GA).  }
\end{figure}
\newpage
\begin{table}[ht]
\footnotesize
\centering
\begin{tabular}{|l|r|r|r|r|r|r|}
   \hline\multicolumn{1}{|c}{Methods} & \multicolumn{2}{|c}{RMSE} & \multicolumn{2}{|c}{MAE}& \multicolumn{2}{|c|}{Cor} \\  \cline{2-7} & 1 Week Ahead & 2 Week Ahead & 1 Week Ahead & 2 Week Ahead & 1 Week Ahead & 2 Week Ahead \\ 
  \hline
ARGO & \textbf{32.251} & \textbf{65.927} & \textbf{25.713} & 44.635 & \textbf{0.952} & 0.857 \\ 
   \hline
AR7 & 44.365 & 84.970 & 32.412 & 55.567 & 0.907 & 0.743 \\ 
   \hline
COVIDhub-ensemble & 37.879 & 66.684 & 26.269 & \textbf{43.442} & 0.937 & 0.860 \\ 
   \hline
Naive & 47.366 & 84.928 & 31.577 & 54.404 & 0.893 & 0.745 \\ 
   \hline
MOBS-GLEAM\_COVID & 44.520 & 87.684 & 32.072 & 59.179 & 0.918 & 0.754 \\ 
   \hline
COVIDhub-baseline & 56.645 & 89.753 & 43.269 & 64.058 & 0.852 & 0.713 \\ 
   \hline
GT-DeepCOVID & 46.662 & 84.060 & 34.262 & 63.956 & 0.904 & 0.761 \\ 
   \hline
JHUAPL-Bucky & 44.880 & 78.692 & 33.632 & 52.272 & 0.928 & \textbf{0.877} \\ 
   \hline
\end{tabular}
\caption{Comparison of different methods for state-level COVID-19 1 to 2 weeks ahead hospitalizations predictions in Delaware (DE). The MSE, MAE, and correlation are reported and best performed method is highlighted in boldface.} 
\end{table}
\begin{figure}[!h] 
  \centering 
\includegraphics[width=.8\linewidth, page=19]{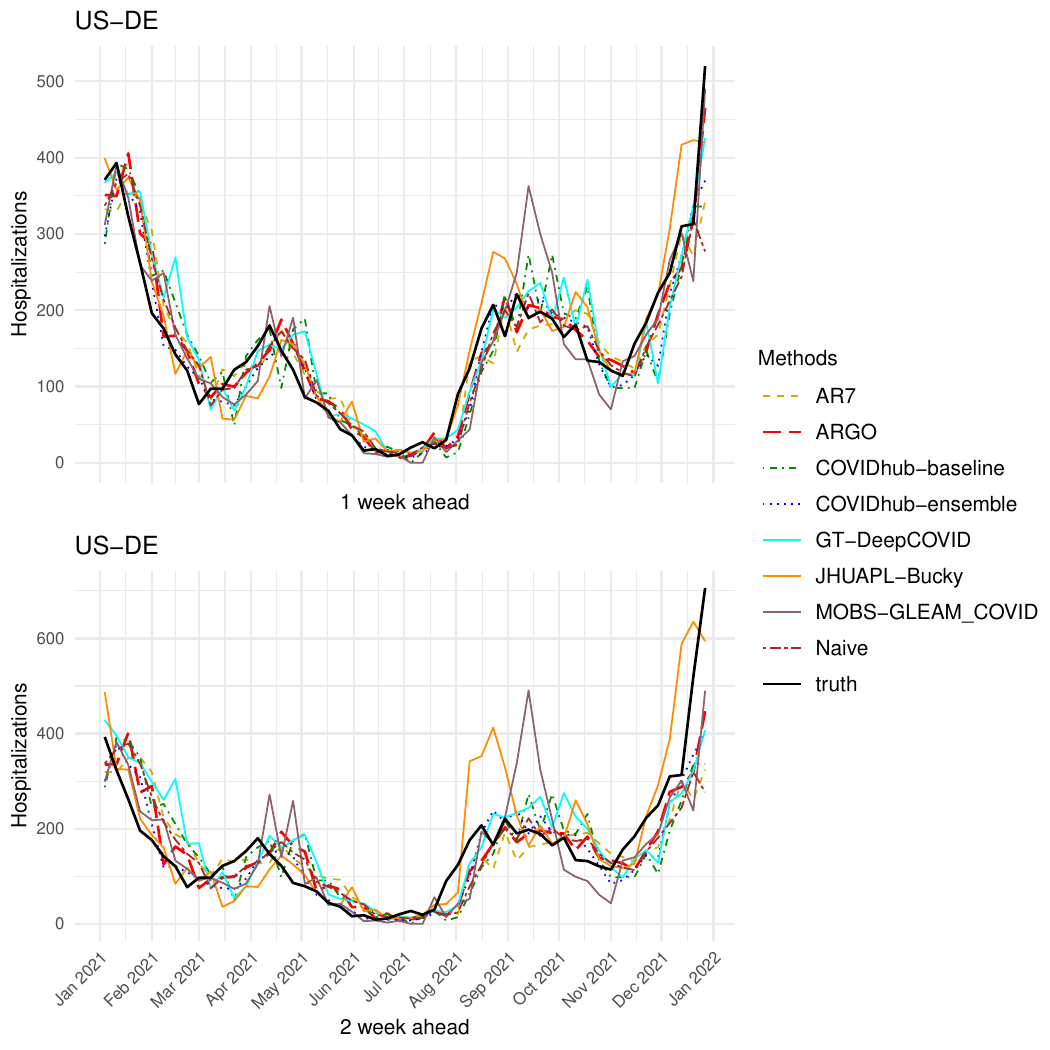} 
\caption{Plots of the COVID-19 hospitalizations 1 week (top), 2 weeks (bottom) ahead estimates of all compared models for New Delaware (DE).  }
\end{figure}
\newpage
\begin{table}[ht]
\footnotesize
\centering
\begin{tabular}{|l|r|r|r|r|r|r|}
   \hline\multicolumn{1}{|c}{Methods} & \multicolumn{2}{|c}{RMSE} & \multicolumn{2}{|c}{MAE}& \multicolumn{2}{|c|}{Cor} \\  \cline{2-7} & 1 Week Ahead & 2 Week Ahead & 1 Week Ahead & 2 Week Ahead & 1 Week Ahead & 2 Week Ahead \\ 
  \hline
ARGO & 138.307 & 277.839 & \textbf{96.234} & 181.331 & \textbf{0.951} & 0.826 \\ 
   \hline
AR7 & \textbf{136.376} & \textbf{237.028} & 98.253 & 167.551 & 0.948 & 0.866 \\ 
   \hline
COVIDhub-ensemble & 150.712 & 245.002 & 100.058 & \textbf{159.212} & 0.943 & \textbf{0.874} \\ 
   \hline
Naive & 156.975 & 270.490 & 108.077 & 188.096 & 0.930 & 0.827 \\ 
   \hline
MOBS-GLEAM\_COVID & 237.516 & 372.914 & 174.290 & 265.896 & 0.878 & 0.735 \\ 
   \hline
COVIDhub-baseline & 175.407 & 279.481 & 121.385 & 173.750 & 0.912 & 0.814 \\ 
   \hline
GT-DeepCOVID & 173.893 & 297.692 & 129.830 & 222.592 & 0.916 & 0.791 \\ 
   \hline
JHUAPL-Bucky & 254.007 & 350.913 & 183.530 & 270.130 & 0.842 & 0.776 \\ 
   \hline
\end{tabular}
\caption{Comparison of different methods for state-level COVID-19 1 to 2 weeks ahead hospitalizations predictions in Colorado (CO). The MSE, MAE, and correlation are reported and best performed method is highlighted in boldface.} 
\end{table}
\begin{figure}[!h] 
  \centering 
\includegraphics[width=.8\linewidth, page=20]{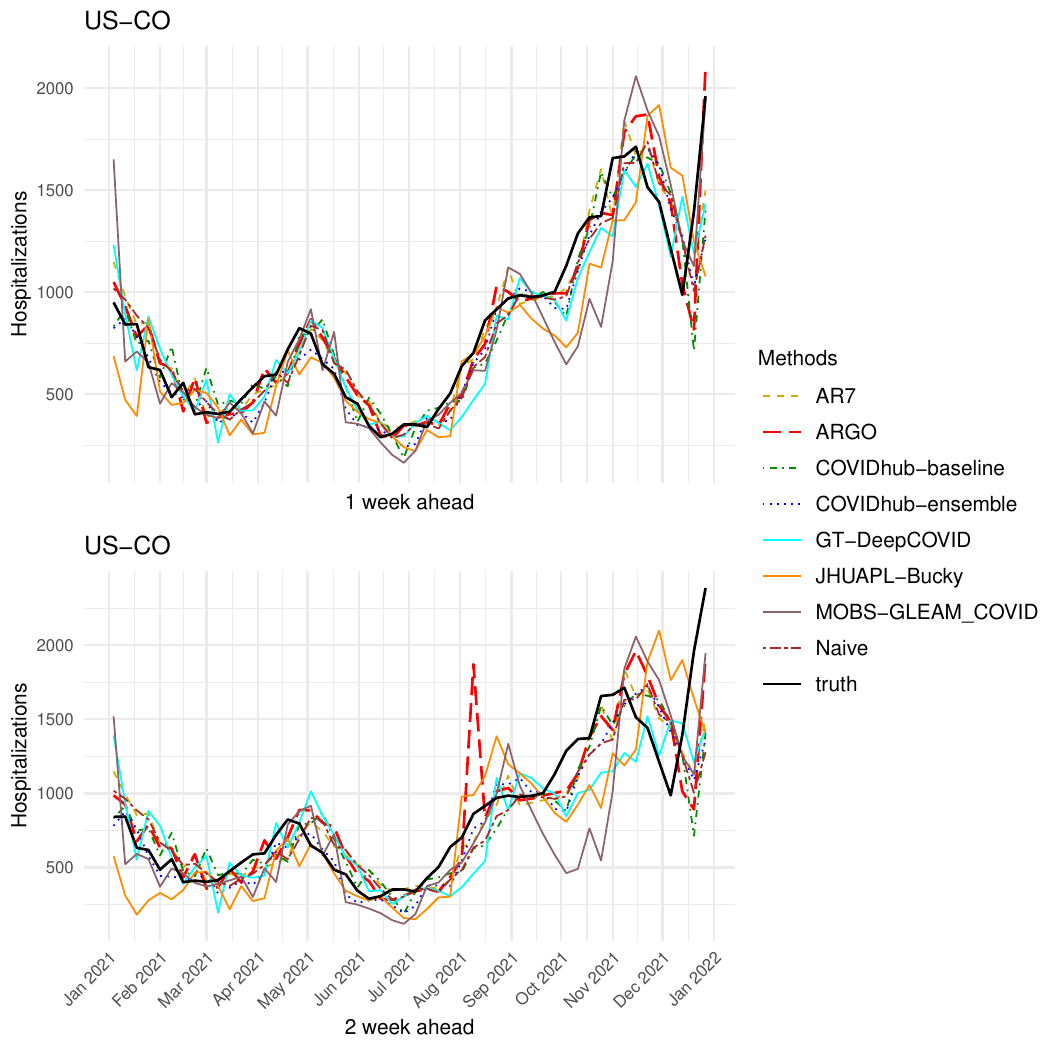} 
\caption{Plots of the COVID-19 hospitalizations 1 week (top), 2 weeks (bottom) ahead estimates of all compared models for New Colorado (CO).   }
\end{figure}
\newpage
\begin{table}[ht]
\footnotesize
\centering
\begin{tabular}{|l|r|r|r|r|r|r|}
   \hline\multicolumn{1}{|c}{Methods} & \multicolumn{2}{|c}{RMSE} & \multicolumn{2}{|c}{MAE}& \multicolumn{2}{|c|}{Cor} \\  \cline{2-7} & 1 Week Ahead & 2 Week Ahead & 1 Week Ahead & 2 Week Ahead & 1 Week Ahead & 2 Week Ahead \\ 
  \hline
ARGO & \textbf{47.255} & 83.070 & \textbf{34.050} & 61.010 & \textbf{0.976} & 0.927 \\ 
   \hline
AR7 & 51.519 & 95.348 & 38.094 & 74.136 & 0.972 & 0.896 \\ 
   \hline
COVIDhub-ensemble & 58.557 & \textbf{75.546} & 42.577 & \textbf{56.308} & 0.965 & \textbf{0.942} \\ 
   \hline
Naive & 60.159 & 96.335 & 45.096 & 76.173 & 0.957 & 0.890 \\ 
   \hline
MOBS-GLEAM\_COVID & 78.844 & 124.249 & 50.910 & 84.301 & 0.932 & 0.820 \\ 
   \hline
COVIDhub-baseline & 75.745 & 107.546 & 54.673 & 80.058 & 0.936 & 0.871 \\ 
   \hline
GT-DeepCOVID & 71.182 & 92.594 & 53.644 & 70.154 & 0.939 & 0.897 \\ 
   \hline
JHUAPL-Bucky & 134.789 & 133.093 & 91.570 & 100.631 & 0.891 & 0.909 \\ 
   \hline
\end{tabular}
\caption{Comparison of different methods for state-level COVID-19 1 to 2 weeks ahead hospitalizations predictions in Montana (MT). The MSE, MAE, and correlation are reported and best performed method is highlighted in boldface.} 
\end{table}
\begin{figure}[!h] 
  \centering 
\includegraphics[width=.8\linewidth, page=21]{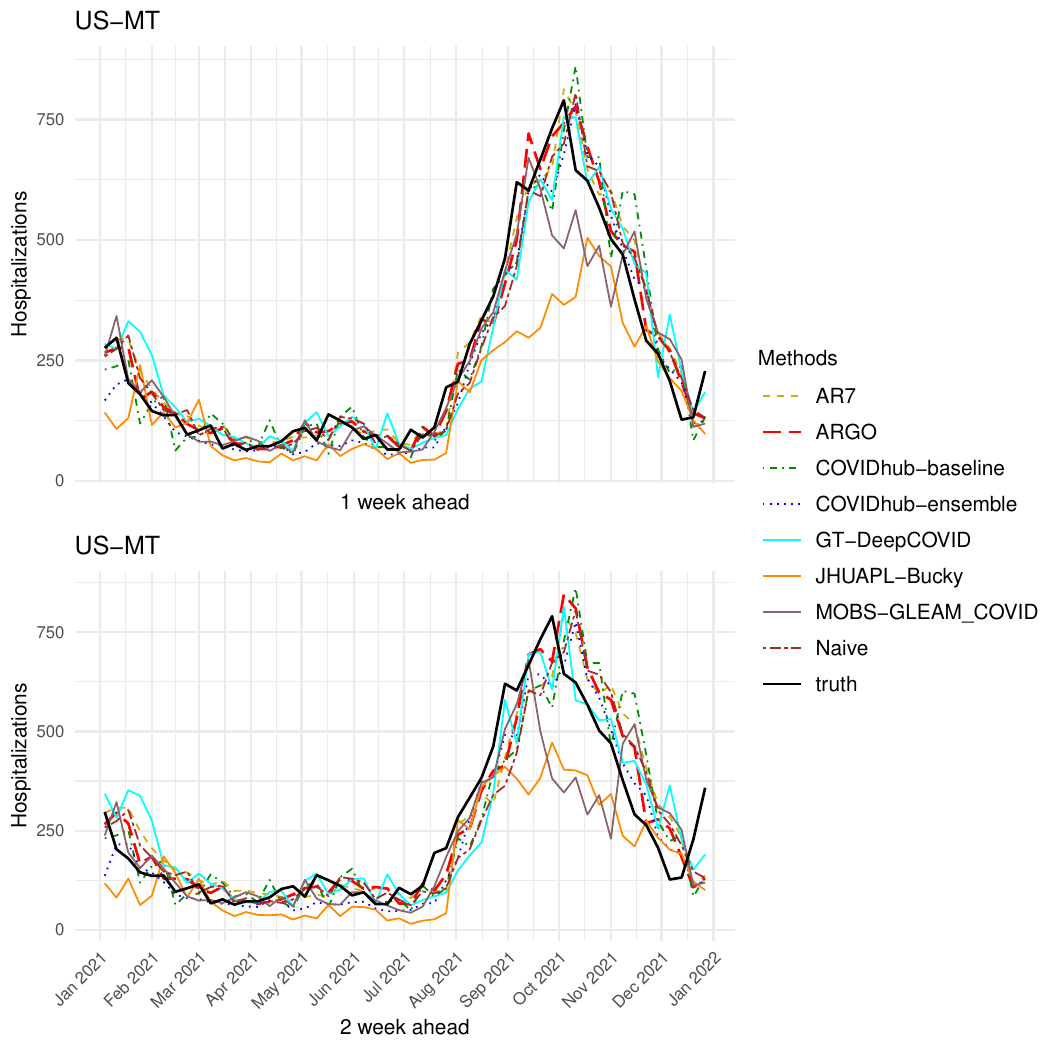} 
\caption{Plots of the COVID-19 hospitalizations 1 week (top), 2 weeks (bottom) ahead estimates of all compared models for New Montana (MT).   }
\end{figure}
\newpage
\begin{table}[ht]
\footnotesize
\centering
\begin{tabular}{|l|r|r|r|r|r|r|}
   \hline\multicolumn{1}{|c}{Methods} & \multicolumn{2}{|c}{RMSE} & \multicolumn{2}{|c}{MAE}& \multicolumn{2}{|c|}{Cor} \\  \cline{2-7} & 1 Week Ahead & 2 Week Ahead & 1 Week Ahead & 2 Week Ahead & 1 Week Ahead & 2 Week Ahead \\ 
  \hline
ARGO & \textbf{86.307} & \textbf{165.545} & \textbf{62.981} & \textbf{106.442} & \textbf{0.950} & \textbf{0.822} \\ 
   \hline
AR7 & 99.743 & 196.085 & 72.532 & 126.137 & 0.930 & 0.741 \\ 
   \hline
COVIDhub-ensemble & 112.266 & 195.618 & 73.942 & 121.769 & 0.927 & 0.788 \\ 
   \hline
Naive & 98.141 & 191.360 & 71.827 & 126.385 & 0.933 & 0.759 \\ 
   \hline
MOBS-GLEAM\_COVID & 126.663 & 216.742 & 101.806 & 163.029 & 0.927 & 0.788 \\ 
   \hline
COVIDhub-baseline & 147.090 & 223.786 & 99.846 & 151.981 & 0.857 & 0.682 \\ 
   \hline
GT-DeepCOVID & 130.237 & 225.229 & 96.337 & 165.418 & 0.920 & 0.748 \\ 
   \hline
JHUAPL-Bucky & 262.864 & 381.213 & 178.968 & 251.930 & 0.894 & 0.762 \\ 
   \hline
\end{tabular}
\caption{Comparison of different methods for state-level COVID-19 1 to 2 weeks ahead hospitalizations predictions in Nevada (NV). The MSE, MAE, and correlation are reported and best performed method is highlighted in boldface.} 
\end{table}
\begin{figure}[!h] 
  \centering 
\includegraphics[width=.8\linewidth, page=22]{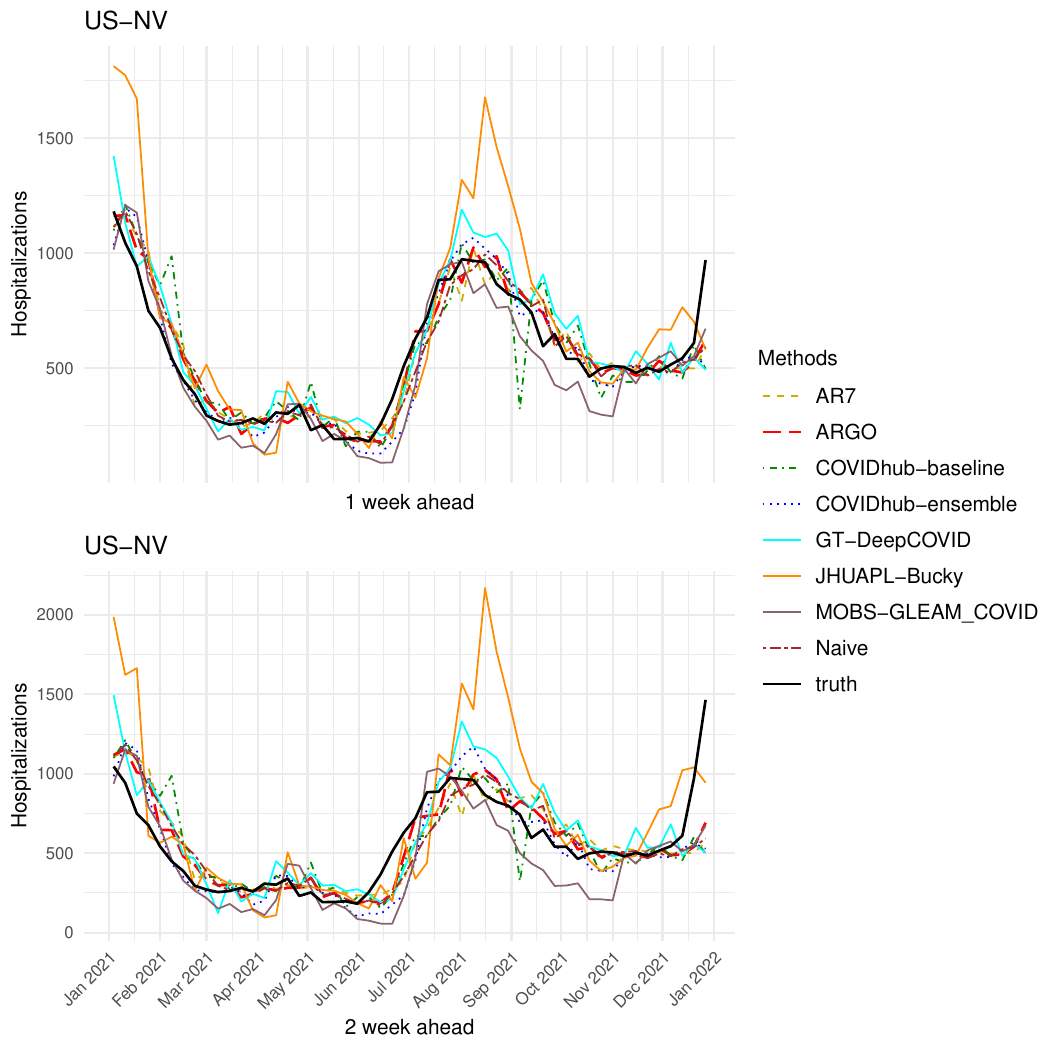} 
\caption{Plots of the COVID-19 hospitalizations 1 week (top), 2 weeks (bottom) ahead estimates of all compared models for New Nevada (NV).   }
\end{figure}
\newpage
\begin{table}[ht]
\footnotesize
\centering
\begin{tabular}{|l|r|r|r|r|r|r|}
   \hline\multicolumn{1}{|c}{Methods} & \multicolumn{2}{|c}{RMSE} & \multicolumn{2}{|c}{MAE}& \multicolumn{2}{|c|}{Cor} \\  \cline{2-7} & 1 Week Ahead & 2 Week Ahead & 1 Week Ahead & 2 Week Ahead & 1 Week Ahead & 2 Week Ahead \\ 
  \hline
ARGO & 325.881 & 499.206 & 183.123 & 355.240 & 0.930 & 0.841 \\ 
   \hline
AR7 & \textbf{218.309} & \textbf{456.009} & \textbf{163.815} & \textbf{337.945} & \textbf{0.966} & 0.843 \\ 
   \hline
COVIDhub-ensemble & 302.950 & 512.788 & 220.942 & 356.865 & 0.937 & 0.845 \\ 
   \hline
Naive & 299.536 & 558.426 & 219.904 & 401.288 & 0.930 & 0.762 \\ 
   \hline
MOBS-GLEAM\_COVID & 307.819 & 481.155 & 229.158 & 350.777 & 0.936 & \textbf{0.860} \\ 
   \hline
COVIDhub-baseline & 334.135 & 575.388 & 243.308 & 413.692 & 0.909 & 0.739 \\ 
   \hline
GT-DeepCOVID & 351.024 & 621.756 & 269.025 & 465.009 & 0.913 & 0.784 \\ 
   \hline
JHUAPL-Bucky & 535.927 & 822.458 & 379.204 & 566.136 & 0.892 & 0.827 \\ 
   \hline
\end{tabular}
\caption{Comparison of different methods for state-level COVID-19 1 to 2 weeks ahead hospitalizations predictions in North Carolina (NC). The MSE, MAE, and correlation are reported and best performed method is highlighted in boldface.} 
\end{table}
\begin{figure}[!h] 
  \centering 
\includegraphics[width=.8\linewidth, page=23]{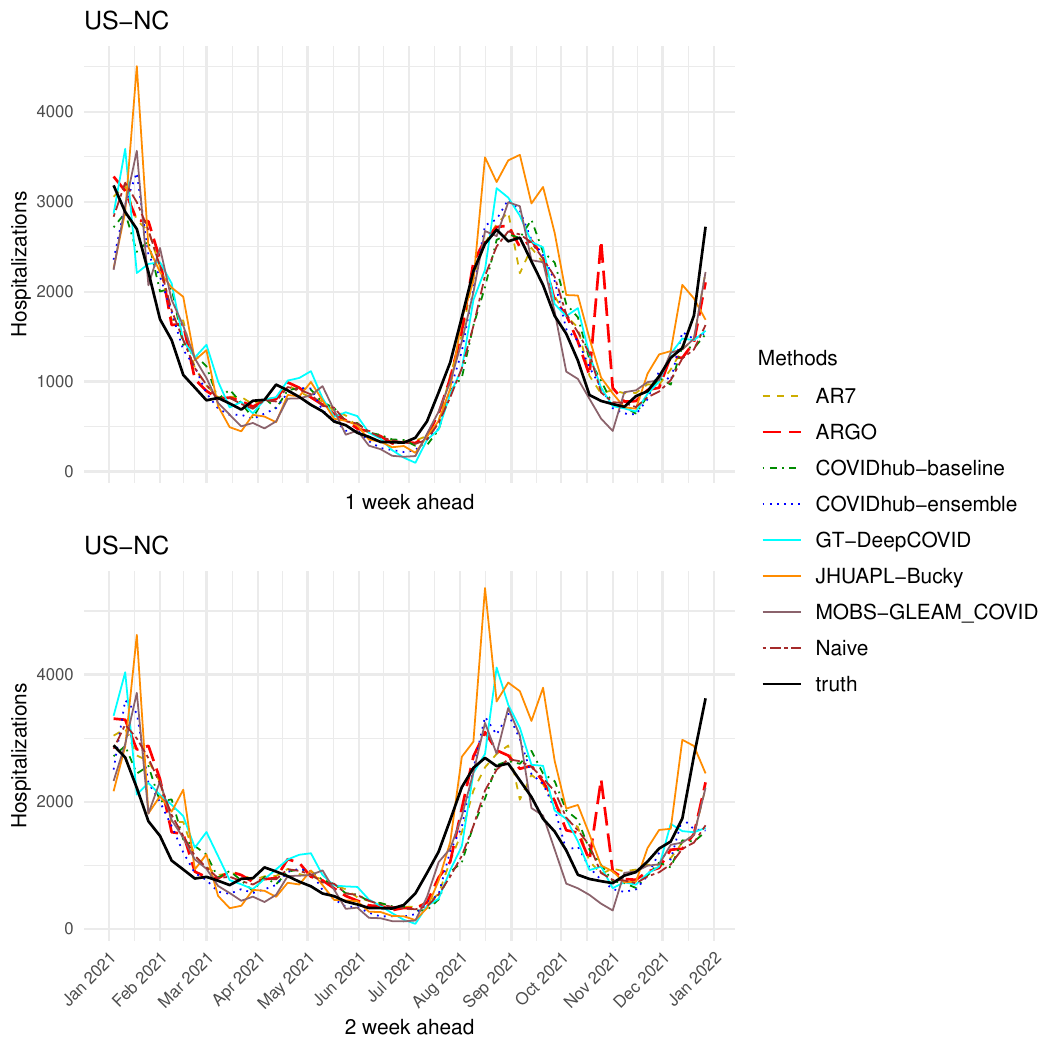} 
\caption{Plots of the COVID-19 hospitalizations 1 week (top), 2 weeks (bottom) ahead estimates of all compared models for New North Carolina (NC).   }
\end{figure}
\newpage
\begin{table}[ht]
\footnotesize
\centering
\begin{tabular}{|l|r|r|r|r|r|r|}
   \hline\multicolumn{1}{|c}{Methods} & \multicolumn{2}{|c}{RMSE} & \multicolumn{2}{|c}{MAE}& \multicolumn{2}{|c|}{Cor} \\  \cline{2-7} & 1 Week Ahead & 2 Week Ahead & 1 Week Ahead & 2 Week Ahead & 1 Week Ahead & 2 Week Ahead \\ 
  \hline
ARGO & \textbf{42.379} & \textbf{72.701} & \textbf{32.991} & \textbf{53.479} & \textbf{0.965} & 0.899 \\ 
   \hline
AR7 & 50.401 & 85.358 & 36.492 & 62.053 & 0.952 & 0.862 \\ 
   \hline
COVIDhub-ensemble & 53.282 & 77.730 & 43.288 & 60.173 & 0.954 & \textbf{0.909} \\ 
   \hline
Naive & 48.898 & 78.454 & 38.365 & 56.577 & 0.953 & 0.881 \\ 
   \hline
MOBS-GLEAM\_COVID & 112.130 & 155.510 & 81.188 & 111.434 & 0.795 & 0.641 \\ 
   \hline
COVIDhub-baseline & 75.107 & 92.617 & 57.365 & 67.077 & 0.892 & 0.838 \\ 
   \hline
GT-DeepCOVID & 64.564 & 79.838 & 48.014 & 62.257 & 0.920 & 0.887 \\ 
   \hline
JHUAPL-Bucky & 107.763 & 154.243 & 80.663 & 123.174 & 0.839 & 0.762 \\ 
   \hline
\end{tabular}
\caption{Comparison of different methods for state-level COVID-19 1 to 2 weeks ahead hospitalizations predictions in Utah (UT). The MSE, MAE, and correlation are reported and best performed method is highlighted in boldface.} 
\end{table}
\begin{figure}[!h] 
  \centering 
\includegraphics[width=.8\linewidth, page=24]{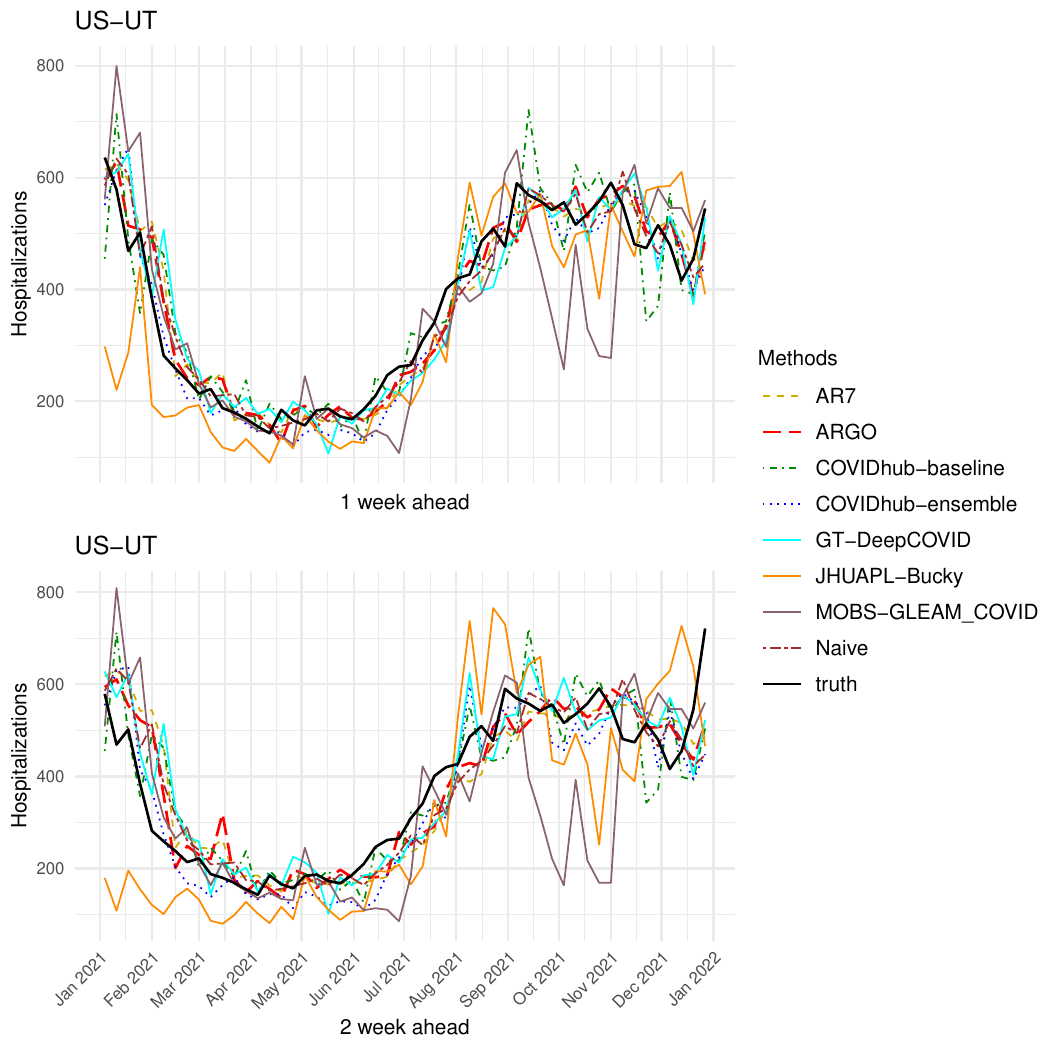} 
\caption{Plots of the COVID-19 hospitalizations 1 week (top), 2 weeks (bottom) ahead estimates of all compared models for New Utah (UT).   }
\end{figure}
\newpage
\begin{table}[ht]
\footnotesize
\centering
\begin{tabular}{|l|r|r|r|r|r|r|}
   \hline\multicolumn{1}{|c}{Methods} & \multicolumn{2}{|c}{RMSE} & \multicolumn{2}{|c}{MAE}& \multicolumn{2}{|c|}{Cor} \\  \cline{2-7} & 1 Week Ahead & 2 Week Ahead & 1 Week Ahead & 2 Week Ahead & 1 Week Ahead & 2 Week Ahead \\ 
  \hline
ARGO & \textbf{42.130} & 63.930 & \textbf{30.963} & 47.644 & \textbf{0.954} & 0.901 \\ 
   \hline
AR7 & 44.237 & 66.137 & 35.017 & 51.867 & 0.947 & 0.883 \\ 
   \hline
COVIDhub-ensemble & 44.898 & \textbf{63.320} & 33.769 & \textbf{44.596} & 0.952 & \textbf{0.910} \\ 
   \hline
Naive & 43.943 & 65.188 & 34.423 & 50.692 & 0.948 & 0.889 \\ 
   \hline
MOBS-GLEAM\_COVID & 60.825 & 91.525 & 42.365 & 65.689 & 0.922 & 0.822 \\ 
   \hline
COVIDhub-baseline & 106.052 & 120.721 & 50.942 & 68.058 & 0.826 & 0.758 \\ 
   \hline
GT-DeepCOVID & 62.474 & 87.314 & 46.571 & 66.467 & 0.897 & 0.819 \\ 
   \hline
JHUAPL-Bucky & 100.201 & 192.865 & 65.111 & 112.632 & 0.861 & 0.711 \\ 
   \hline
\end{tabular}
\caption{Comparison of different methods for state-level COVID-19 1 to 2 weeks ahead hospitalizations predictions in Nebraska (NE). The MSE, MAE, and correlation are reported and best performed method is highlighted in boldface.} 
\end{table}
\begin{figure}[!h] 
  \centering 
\includegraphics[width=.8\linewidth, page=25]{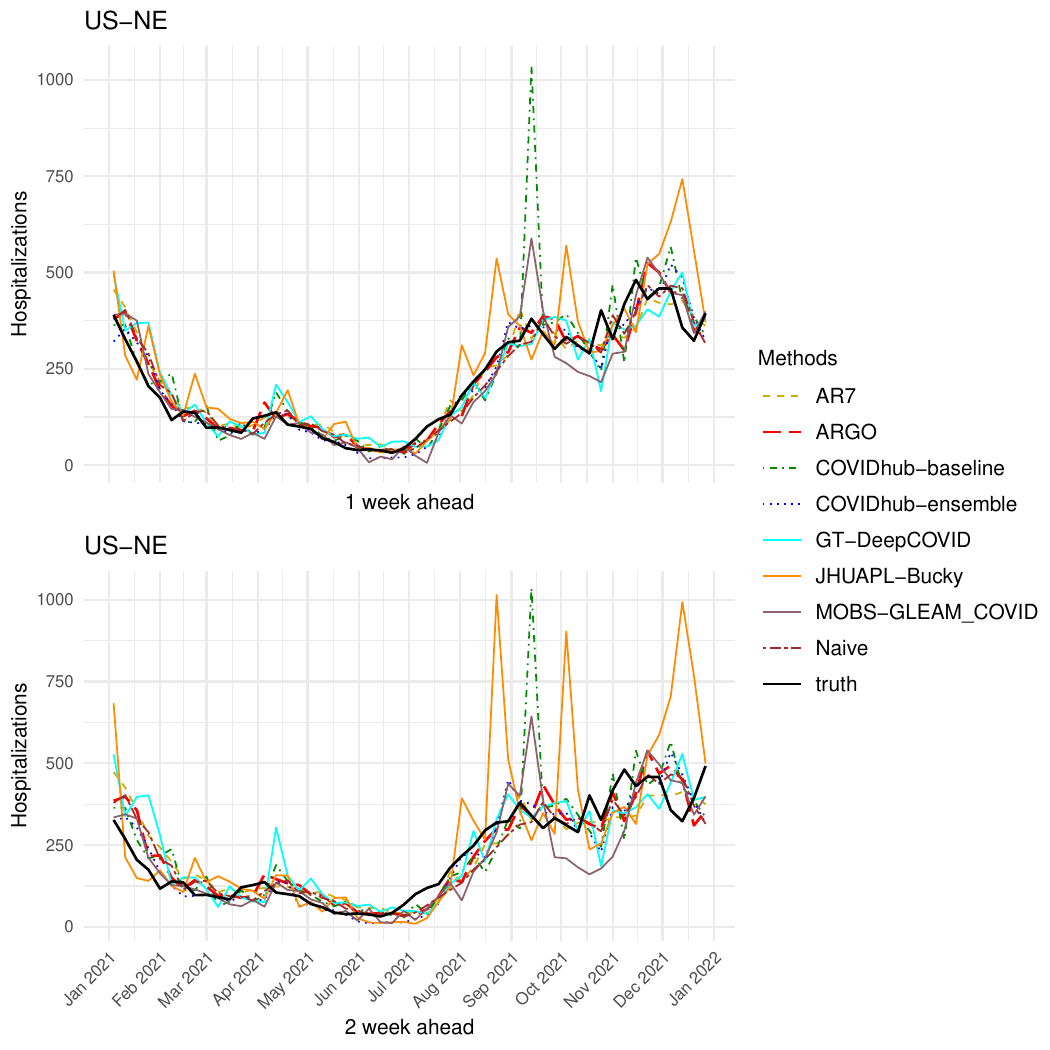} 
\caption{Plots of the COVID-19 hospitalizations 1 week (top), 2 weeks (bottom) ahead estimates of all compared models for New Nebraska (NE).   }
\end{figure}
\newpage
\begin{table}[ht]
\footnotesize
\centering
\begin{tabular}{|l|r|r|r|r|r|r|}
   \hline\multicolumn{1}{|c}{Methods} & \multicolumn{2}{|c}{RMSE} & \multicolumn{2}{|c}{MAE}& \multicolumn{2}{|c|}{Cor} \\  \cline{2-7} & 1 Week Ahead & 2 Week Ahead & 1 Week Ahead & 2 Week Ahead & 1 Week Ahead & 2 Week Ahead \\ 
  \hline
ARGO & \textbf{480.564} & 1253.041 & \textbf{294.576} & \textbf{653.258} & \textbf{0.975} & 0.850 \\ 
   \hline
AR7 & 484.782 & \textbf{1176.086} & 304.829 & 659.807 & 0.974 & \textbf{0.869} \\ 
   \hline
COVIDhub-ensemble & 832.408 & 1338.327 & 396.269 & 702.096 & 0.922 & 0.833 \\ 
   \hline
Naive & 876.339 & 1519.967 & 431.654 & 809.596 & 0.911 & 0.772 \\ 
   \hline
MOBS-GLEAM\_COVID & 688.959 & 1351.754 & 486.538 & 865.310 & 0.949 & 0.831 \\ 
   \hline
COVIDhub-baseline & 971.889 & 1572.613 & 507.250 & 830.692 & 0.891 & 0.754 \\ 
   \hline
GT-DeepCOVID & 660.281 & 1269.201 & 456.087 & 713.004 & 0.951 & 0.848 \\ 
   \hline
JHUAPL-Bucky & 1127.248 & 1564.264 & 673.512 & 941.596 & 0.903 & \textbf{0.869} \\ 
   \hline
\end{tabular}
\caption{Comparison of different methods for state-level COVID-19 1 to 2 weeks ahead hospitalizations predictions in New York (NY). The MSE, MAE, and correlation are reported and best performed method is highlighted in boldface.} 
\end{table}
\begin{figure}[!h] 
  \centering 
\includegraphics[width=.8\linewidth, page=26]{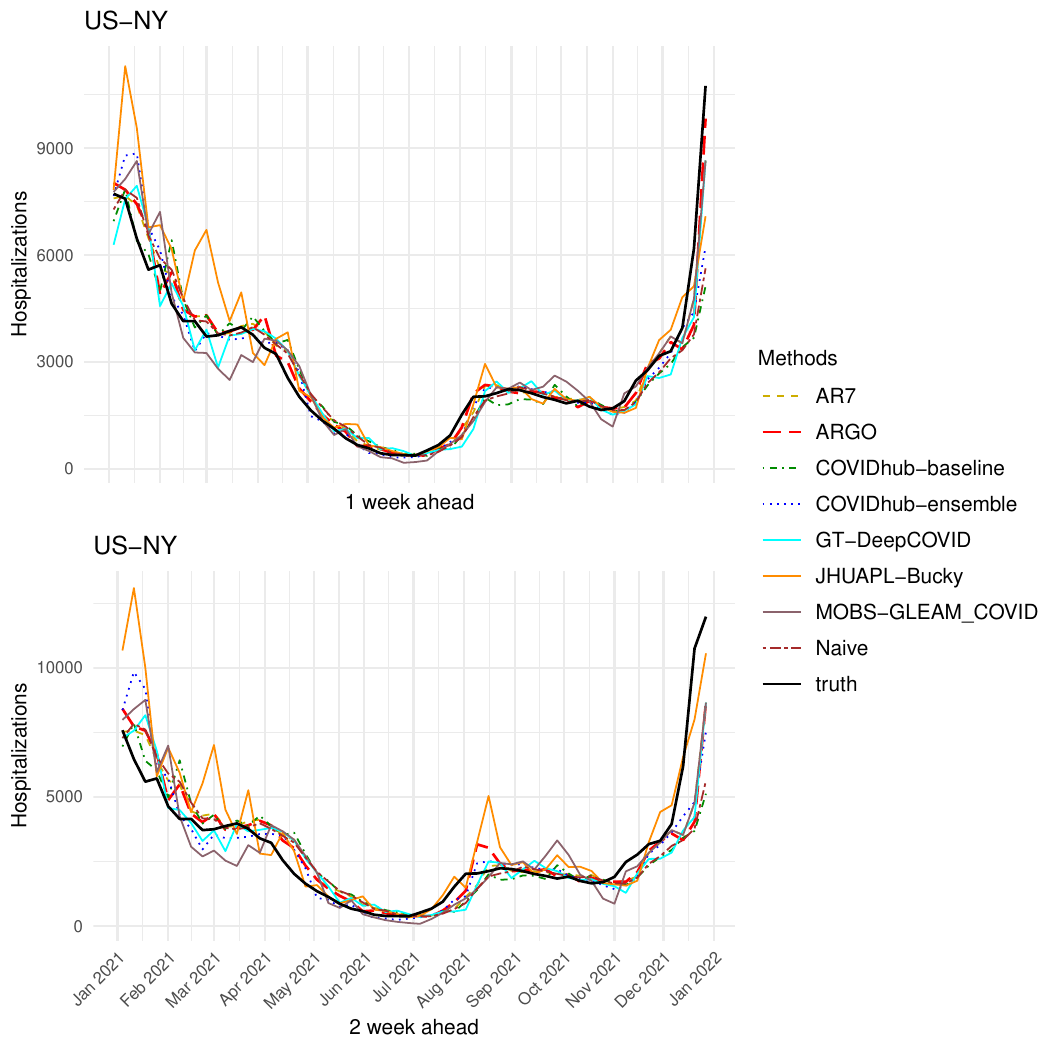} 
\caption{Plots of the COVID-19 hospitalizations 1 week (top), 2 weeks (bottom) ahead estimates of all compared models for New New York (NY).   }
\end{figure}
\newpage
\begin{table}[ht]
\footnotesize
\centering
\begin{tabular}{|l|r|r|r|r|r|r|}
   \hline\multicolumn{1}{|c}{Methods} & \multicolumn{2}{|c}{RMSE} & \multicolumn{2}{|c}{MAE}& \multicolumn{2}{|c|}{Cor} \\  \cline{2-7} & 1 Week Ahead & 2 Week Ahead & 1 Week Ahead & 2 Week Ahead & 1 Week Ahead & 2 Week Ahead \\ 
  \hline
ARGO & 70.956 & 114.018 & 49.118 & 82.550 & 0.945 & 0.859 \\ 
   \hline
AR7 & \textbf{68.576} & 121.906 & 49.623 & 89.501 & 0.945 & 0.823 \\ 
   \hline
COVIDhub-ensemble & 68.721 & \textbf{110.588} & \textbf{49.096} & \textbf{77.308} & \textbf{0.955} & 0.892 \\ 
   \hline
Naive & 78.339 & 136.946 & 57.019 & 95.731 & 0.925 & 0.774 \\ 
   \hline
MOBS-GLEAM\_COVID & 83.624 & 120.248 & 63.678 & 91.975 & 0.920 & 0.845 \\ 
   \hline
COVIDhub-baseline & 82.469 & 139.727 & 59.058 & 101.000 & 0.916 & 0.763 \\ 
   \hline
GT-DeepCOVID & 99.027 & 177.124 & 75.166 & 127.312 & 0.894 & 0.762 \\ 
   \hline
JHUAPL-Bucky & 161.293 & 354.549 & 112.632 & 211.784 & 0.949 & \textbf{0.901} \\ 
   \hline
\end{tabular}
\caption{Comparison of different methods for state-level COVID-19 1 to 2 weeks ahead hospitalizations predictions in Oregon (OR). The MSE, MAE, and correlation are reported and best performed method is highlighted in boldface.} 
\end{table}
\begin{figure}[!h] 
  \centering 
\includegraphics[width=.8\linewidth, page=27]{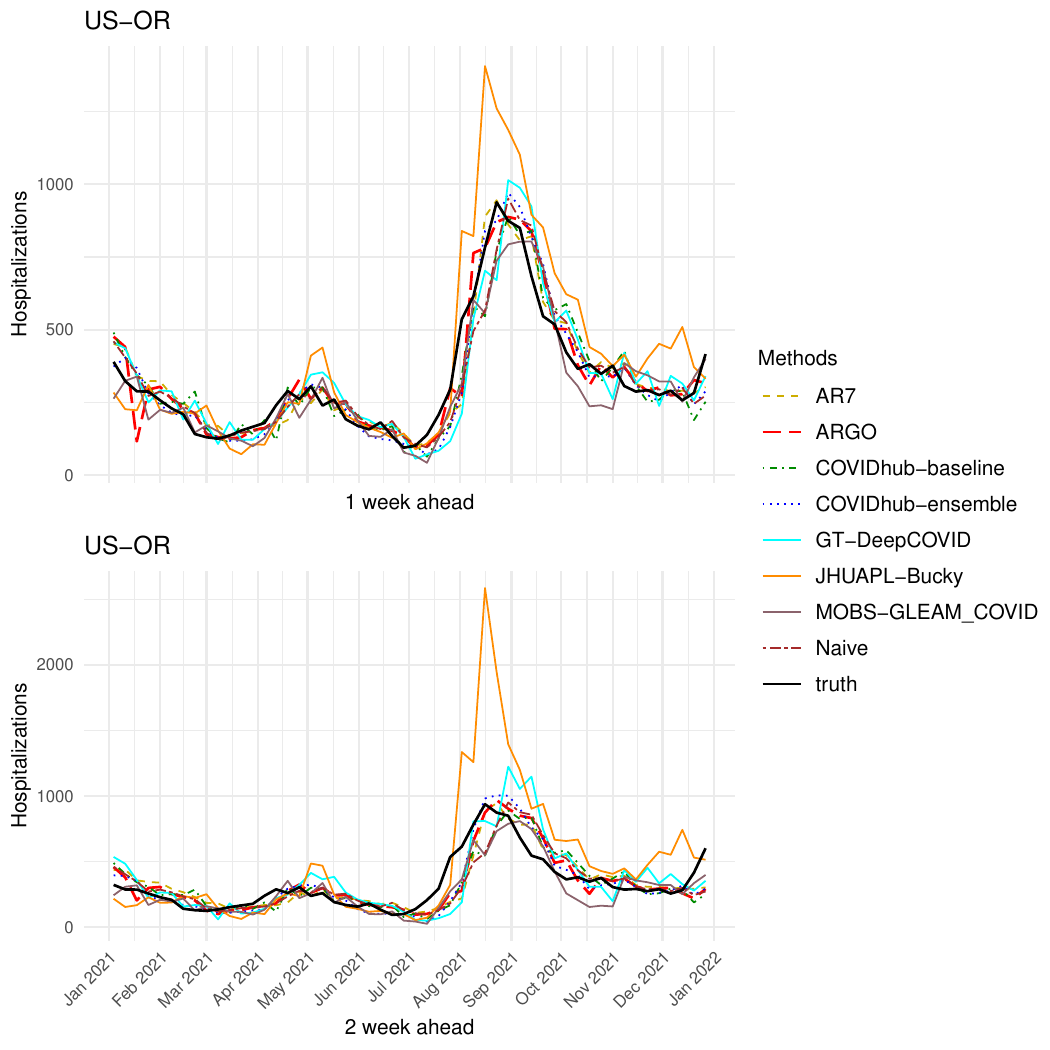} 
\caption{Plots of the COVID-19 hospitalizations 1 week (top), 2 weeks (bottom) ahead estimates of all compared models for New Oregon (OR).    }
\end{figure}
\newpage
\begin{table}[ht]
\footnotesize
\centering
\begin{tabular}{|l|r|r|r|r|r|r|}
   \hline\multicolumn{1}{|c}{Methods} & \multicolumn{2}{|c}{RMSE} & \multicolumn{2}{|c}{MAE}& \multicolumn{2}{|c|}{Cor} \\  \cline{2-7} & 1 Week Ahead & 2 Week Ahead & 1 Week Ahead & 2 Week Ahead & 1 Week Ahead & 2 Week Ahead \\ 
  \hline
ARGO & 140.275 & 253.954 & 84.862 & 158.993 & 0.959 & 0.889 \\ 
   \hline
AR7 & 145.483 & 270.938 & 82.568 & 159.024 & 0.955 & 0.870 \\ 
   \hline
COVIDhub-ensemble & \textbf{132.464} & \textbf{233.235} & \textbf{74.058} & \textbf{142.731} & \textbf{0.964} & \textbf{0.908} \\ 
   \hline
Naive & 175.455 & 321.447 & 105.115 & 197.462 & 0.935 & 0.814 \\ 
   \hline
MOBS-GLEAM\_COVID & 178.676 & 290.679 & 122.435 & 194.646 & 0.933 & 0.853 \\ 
   \hline
COVIDhub-baseline & 169.419 & 300.656 & 104.519 & 191.904 & 0.938 & 0.837 \\ 
   \hline
GT-DeepCOVID & 205.979 & 340.541 & 129.320 & 230.928 & 0.915 & 0.803 \\ 
   \hline
JHUAPL-Bucky & 157.056 & 259.875 & 91.521 & 159.747 & 0.948 & 0.891 \\ 
   \hline
\end{tabular}
\caption{Comparison of different methods for state-level COVID-19 1 to 2 weeks ahead hospitalizations predictions in Massachusetts (MA). The MSE, MAE, and correlation are reported and best performed method is highlighted in boldface.} 
\end{table}
\begin{figure}[!h] 
  \centering 
\includegraphics[width=.8\linewidth, page=28]{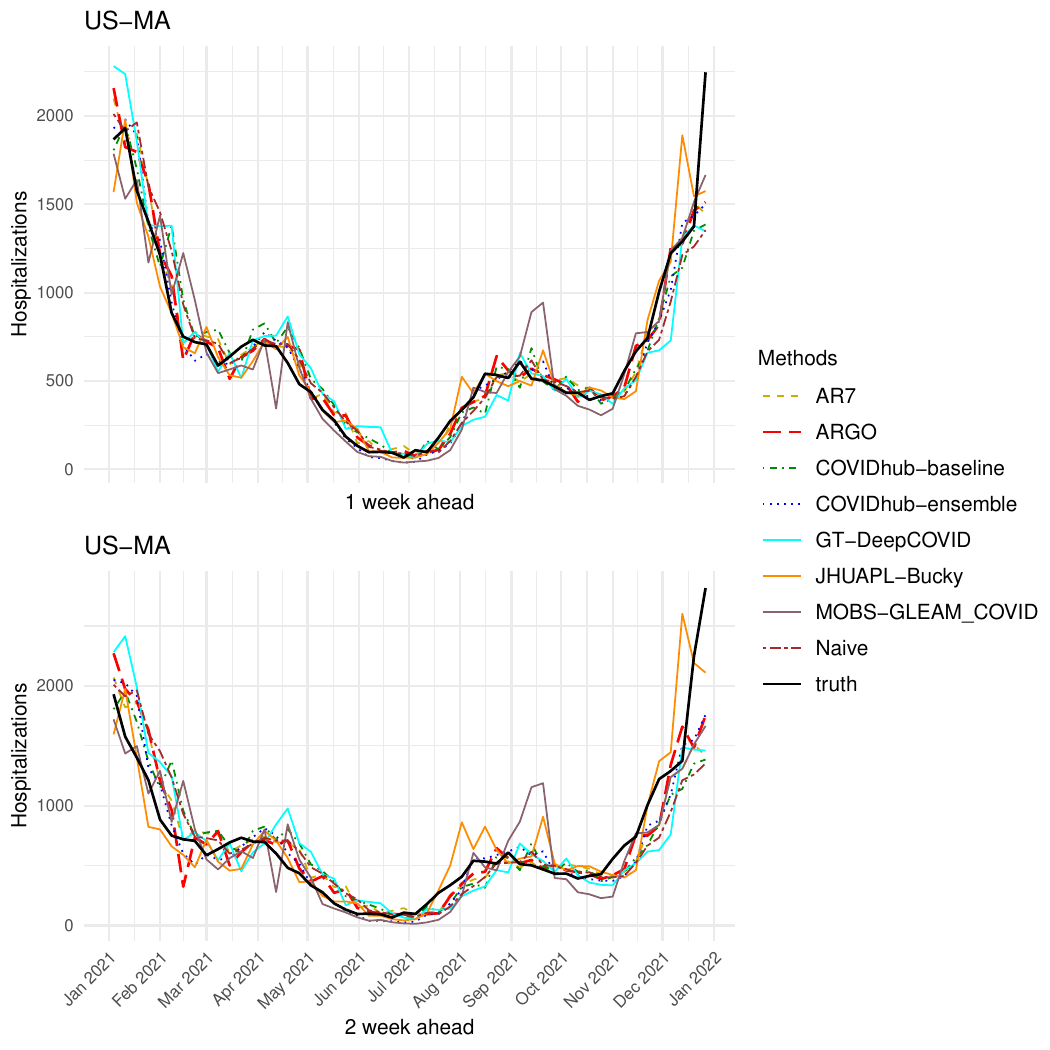} 
\caption{Plots of the COVID-19 hospitalizations 1 week (top), 2 weeks (bottom) ahead estimates of all compared models for New Massachusetts (MA).     }
\end{figure}
\newpage
\begin{table}[ht]
\footnotesize
\centering
\begin{tabular}{|l|r|r|r|r|r|r|}
   \hline\multicolumn{1}{|c}{Methods} & \multicolumn{2}{|c}{RMSE} & \multicolumn{2}{|c}{MAE}& \multicolumn{2}{|c|}{Cor} \\  \cline{2-7} & 1 Week Ahead & 2 Week Ahead & 1 Week Ahead & 2 Week Ahead & 1 Week Ahead & 2 Week Ahead \\ 
  \hline
ARGO & \textbf{93.288} & \textbf{160.449} & \textbf{69.217} & \textbf{131.633} & \textbf{0.973} & \textbf{0.917} \\ 
   \hline
AR7 & 102.632 & 198.858 & 81.008 & 150.721 & 0.964 & 0.855 \\ 
   \hline
COVIDhub-ensemble & 116.690 & 191.548 & 85.654 & 132.365 & 0.960 & 0.892 \\ 
   \hline
Naive & 119.640 & 220.201 & 89.846 & 169.481 & 0.946 & 0.816 \\ 
   \hline
MOBS-GLEAM\_COVID & 186.978 & 274.706 & 132.295 & 195.205 & 0.896 & 0.771 \\ 
   \hline
COVIDhub-baseline & 152.599 & 239.035 & 113.769 & 186.442 & 0.920 & 0.796 \\ 
   \hline
GT-DeepCOVID & 173.042 & 279.126 & 131.280 & 206.161 & 0.915 & 0.788 \\ 
   \hline
JHUAPL-Bucky & 215.347 & 297.893 & 156.615 & 220.588 & 0.880 & 0.800 \\ 
   \hline
\end{tabular}
\caption{Comparison of different methods for state-level COVID-19 1 to 2 weeks ahead hospitalizations predictions in Arkansas (AR). The MSE, MAE, and correlation are reported and best performed method is highlighted in boldface.} 
\end{table}
\begin{figure}[!h] 
  \centering 
\includegraphics[width=.8\linewidth, page=29]{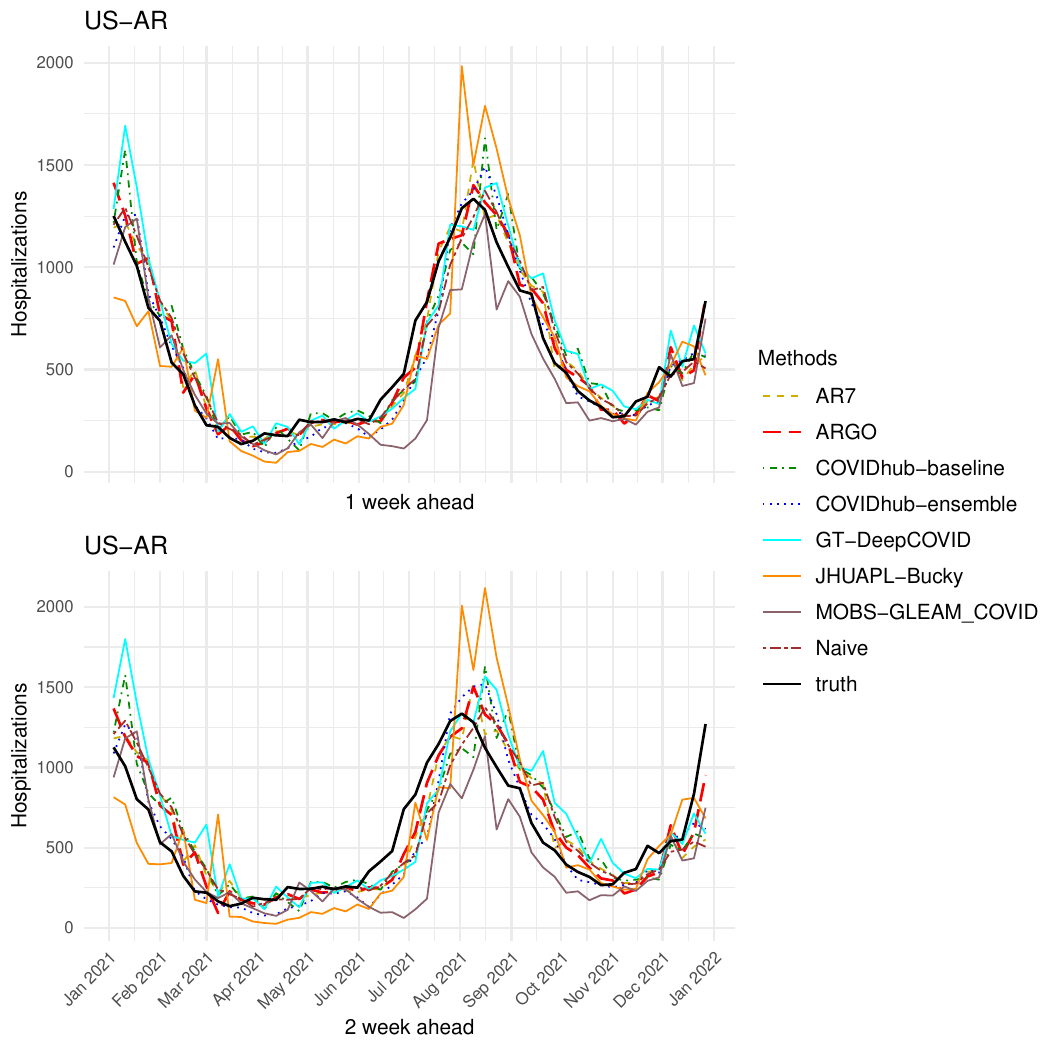} 
\caption{Plots of the COVID-19 hospitalizations 1 week (top), 2 weeks (bottom) ahead estimates of all compared models for New Arkansas (AR).     }
\end{figure}
\newpage
\begin{table}[ht]
\footnotesize
\centering
\begin{tabular}{|l|r|r|r|r|r|r|}
   \hline\multicolumn{1}{|c}{Methods} & \multicolumn{2}{|c}{RMSE} & \multicolumn{2}{|c}{MAE}& \multicolumn{2}{|c|}{Cor} \\  \cline{2-7} & 1 Week Ahead & 2 Week Ahead & 1 Week Ahead & 2 Week Ahead & 1 Week Ahead & 2 Week Ahead \\ 
  \hline
ARGO & 18.832 & \textbf{25.268} & \textbf{13.442} & \textbf{19.389} & 0.944 & 0.899 \\ 
   \hline
AR7 & 19.934 & 30.782 & 14.759 & 24.012 & 0.939 & 0.848 \\ 
   \hline
COVIDhub-ensemble & 19.049 & 26.665 & 15.577 & 19.962 & 0.948 & \textbf{0.907} \\ 
   \hline
Naive & \textbf{17.975} & 27.653 & 13.750 & 21.288 & \textbf{0.949} & 0.879 \\ 
   \hline
MOBS-GLEAM\_COVID & 36.580 & 48.489 & 27.378 & 35.389 & 0.773 & 0.598 \\ 
   \hline
COVIDhub-baseline & 27.480 & 31.138 & 20.635 & 22.788 & 0.882 & 0.848 \\ 
   \hline
GT-DeepCOVID & 33.029 & 45.123 & 24.355 & 34.253 & 0.833 & 0.737 \\ 
   \hline
JHUAPL-Bucky & 40.895 & 62.014 & 26.846 & 40.007 & 0.916 & 0.853 \\ 
   \hline
\end{tabular}
\caption{Comparison of different methods for state-level COVID-19 1 to 2 weeks ahead hospitalizations predictions in Alaska (AK). The MSE, MAE, and correlation are reported and best performed method is highlighted in boldface.} 
\end{table}
\begin{figure}[!h] 
  \centering 
\includegraphics[width=.8\linewidth, page=30]{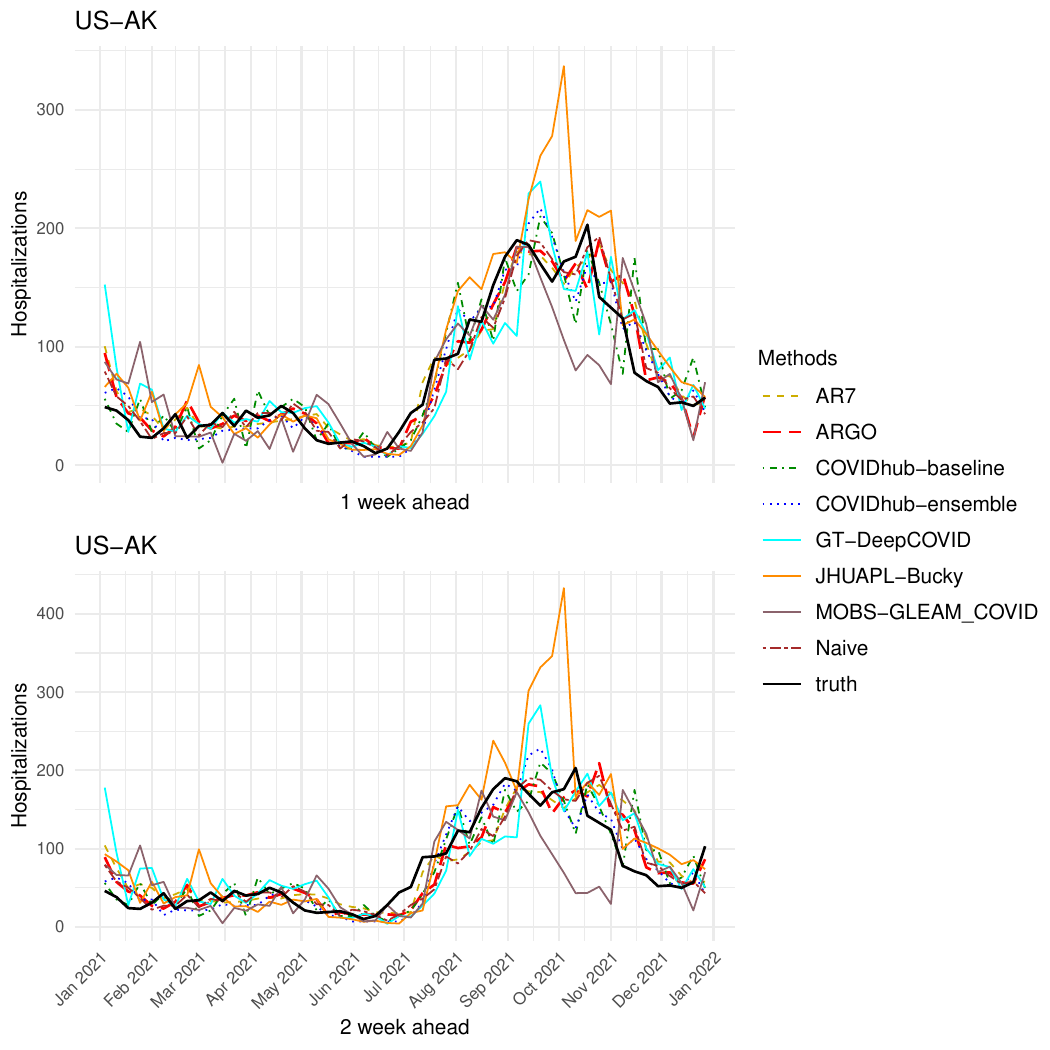} 
\caption{Plots of the COVID-19 hospitalizations 1 week (top), 2 weeks (bottom) ahead estimates of all compared models for New Alaska (AK).     }
\end{figure}
\newpage
\begin{table}[ht]
\footnotesize
\centering
\begin{tabular}{|l|r|r|r|r|r|r|}
   \hline\multicolumn{1}{|c}{Methods} & \multicolumn{2}{|c}{RMSE} & \multicolumn{2}{|c}{MAE}& \multicolumn{2}{|c|}{Cor} \\  \cline{2-7} & 1 Week Ahead & 2 Week Ahead & 1 Week Ahead & 2 Week Ahead & 1 Week Ahead & 2 Week Ahead \\ 
  \hline
ARGO & \textbf{241.217} & \textbf{559.846} & \textbf{182.339} & \textbf{408.646} & \textbf{0.985} & \textbf{0.930} \\ 
   \hline
AR7 & 301.667 & 640.456 & 214.845 & 470.570 & 0.976 & 0.910 \\ 
   \hline
COVIDhub-ensemble & 408.214 & 634.251 & 255.462 & 419.615 & 0.958 & 0.917 \\ 
   \hline
Naive & 422.981 & 747.374 & 306.115 & 564.692 & 0.953 & 0.876 \\ 
   \hline
MOBS-GLEAM\_COVID & 538.346 & 816.229 & 395.590 & 627.383 & 0.930 & 0.869 \\ 
   \hline
COVIDhub-baseline & 473.253 & 769.653 & 334.769 & 555.308 & 0.941 & 0.871 \\ 
   \hline
GT-DeepCOVID & 404.838 & 768.588 & 274.394 & 543.376 & 0.956 & 0.873 \\ 
   \hline
JHUAPL-Bucky & 560.093 & 756.705 & 430.132 & 552.097 & 0.929 & 0.917 \\ 
   \hline
\end{tabular}
\caption{Comparison of different methods for state-level COVID-19 1 to 2 weeks ahead hospitalizations predictions in Ohio (OH). The MSE, MAE, and correlation are reported and best performed method is highlighted in boldface.} 
\end{table}
\begin{figure}[!h] 
  \centering 
\includegraphics[width=.8\linewidth, page=31]{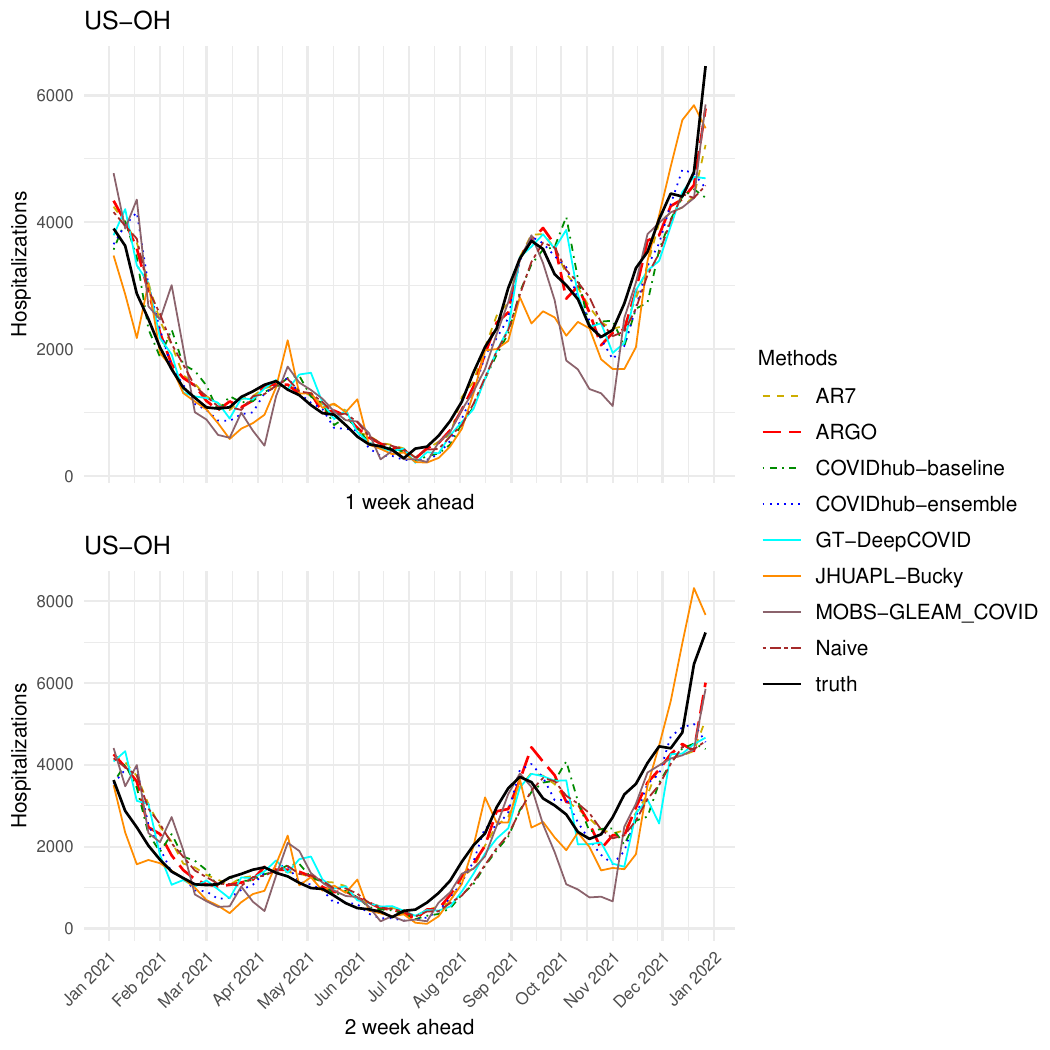} 
\caption{Plots of the COVID-19 hospitalizations 1 week (top), 2 weeks (bottom) ahead estimates of all compared models for New Ohio (OH).     }
\end{figure}
\newpage
\begin{table}[ht]
\footnotesize
\centering
\begin{tabular}{|l|r|r|r|r|r|r|}
   \hline\multicolumn{1}{|c}{Methods} & \multicolumn{2}{|c}{RMSE} & \multicolumn{2}{|c}{MAE}& \multicolumn{2}{|c|}{Cor} \\  \cline{2-7} & 1 Week Ahead & 2 Week Ahead & 1 Week Ahead & 2 Week Ahead & 1 Week Ahead & 2 Week Ahead \\ 
  \hline
ARGO & \textbf{47.537} & \textbf{90.427} & \textbf{36.538} & \textbf{73.412} & \textbf{0.974} & 0.909 \\ 
   \hline
AR7 & 55.714 & 100.311 & 45.173 & 78.422 & 0.960 & 0.871 \\ 
   \hline
COVIDhub-ensemble & 65.042 & 107.526 & 46.923 & 77.712 & 0.965 & \textbf{0.912} \\ 
   \hline
Naive & 67.355 & 114.326 & 52.981 & 90.750 & 0.943 & 0.839 \\ 
   \hline
MOBS-GLEAM\_COVID & 73.086 & 106.901 & 55.463 & 80.139 & 0.947 & 0.899 \\ 
   \hline
COVIDhub-baseline & 89.307 & 130.608 & 65.962 & 101.923 & 0.923 & 0.823 \\ 
   \hline
GT-DeepCOVID & 83.356 & 130.505 & 67.665 & 102.860 & 0.917 & 0.826 \\ 
   \hline
JHUAPL-Bucky & 161.839 & 248.272 & 110.168 & 171.911 & 0.884 & 0.844 \\ 
   \hline
\end{tabular}
\caption{Comparison of different methods for state-level COVID-19 1 to 2 weeks ahead hospitalizations predictions in West Virginia (WV). The MSE, MAE, and correlation are reported and best performed method is highlighted in boldface.} 
\end{table}
\begin{figure}[!h] 
  \centering 
\includegraphics[width=.8\linewidth, page=32]{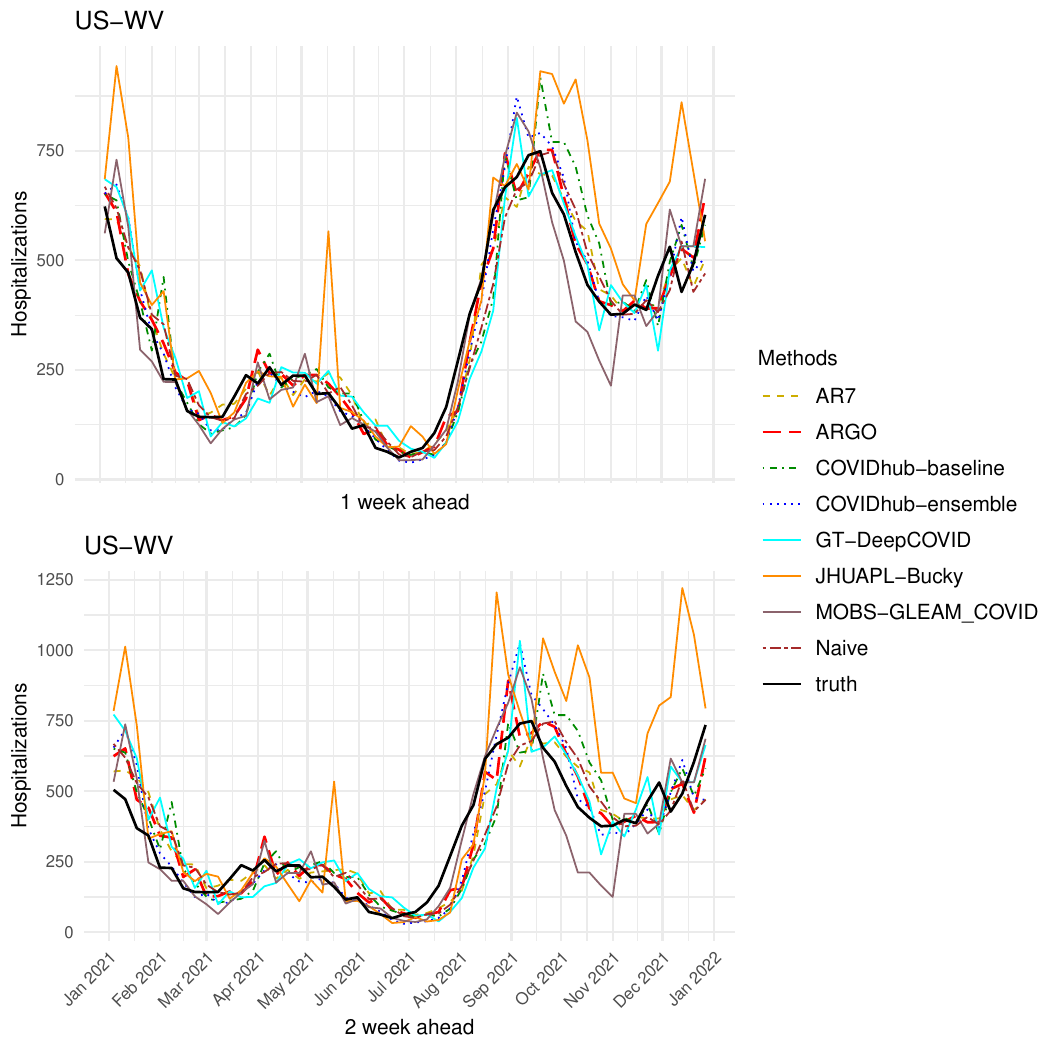} 
\caption{Plots of the COVID-19 hospitalizations 1 week (top), 2 weeks (bottom) ahead estimates of all compared models for New West Virginia (WV).      }
\end{figure}
\newpage
\begin{table}[ht]
\footnotesize
\centering
\begin{tabular}{|l|r|r|r|r|r|r|}
   \hline\multicolumn{1}{|c}{Methods} & \multicolumn{2}{|c}{RMSE} & \multicolumn{2}{|c}{MAE}& \multicolumn{2}{|c|}{Cor} \\  \cline{2-7} & 1 Week Ahead & 2 Week Ahead & 1 Week Ahead & 2 Week Ahead & 1 Week Ahead & 2 Week Ahead \\ 
  \hline
ARGO & 31.523 & \textbf{39.868} & 20.774 & 29.749 & \textbf{0.928} & \textbf{0.874} \\ 
   \hline
AR7 & 29.258 & 48.270 & 20.759 & 34.960 & 0.926 & 0.793 \\ 
   \hline
COVIDhub-ensemble & 33.487 & 42.151 & 24.750 & 30.827 & 0.915 & 0.865 \\ 
   \hline
Naive & \textbf{28.912} & 41.082 & \textbf{19.481} & \textbf{29.577} & \textbf{0.928} & 0.855 \\ 
   \hline
MOBS-GLEAM\_COVID & 42.033 & 56.486 & 30.655 & 41.229 & 0.844 & 0.719 \\ 
   \hline
COVIDhub-baseline & 35.713 & 45.431 & 27.654 & 33.135 & 0.905 & 0.843 \\ 
   \hline
GT-DeepCOVID & 45.661 & 61.359 & 33.846 & 47.373 & 0.836 & 0.735 \\ 
   \hline
JHUAPL-Bucky & 44.127 & 53.392 & 33.299 & 39.225 & 0.854 & 0.769 \\ 
   \hline
\end{tabular}
\caption{Comparison of different methods for state-level COVID-19 1 to 2 weeks ahead hospitalizations predictions in Wyoming (WY). The MSE, MAE, and correlation are reported and best performed method is highlighted in boldface.} 
\end{table}
\begin{figure}[!h] 
  \centering 
\includegraphics[width=.8\linewidth, page=33]{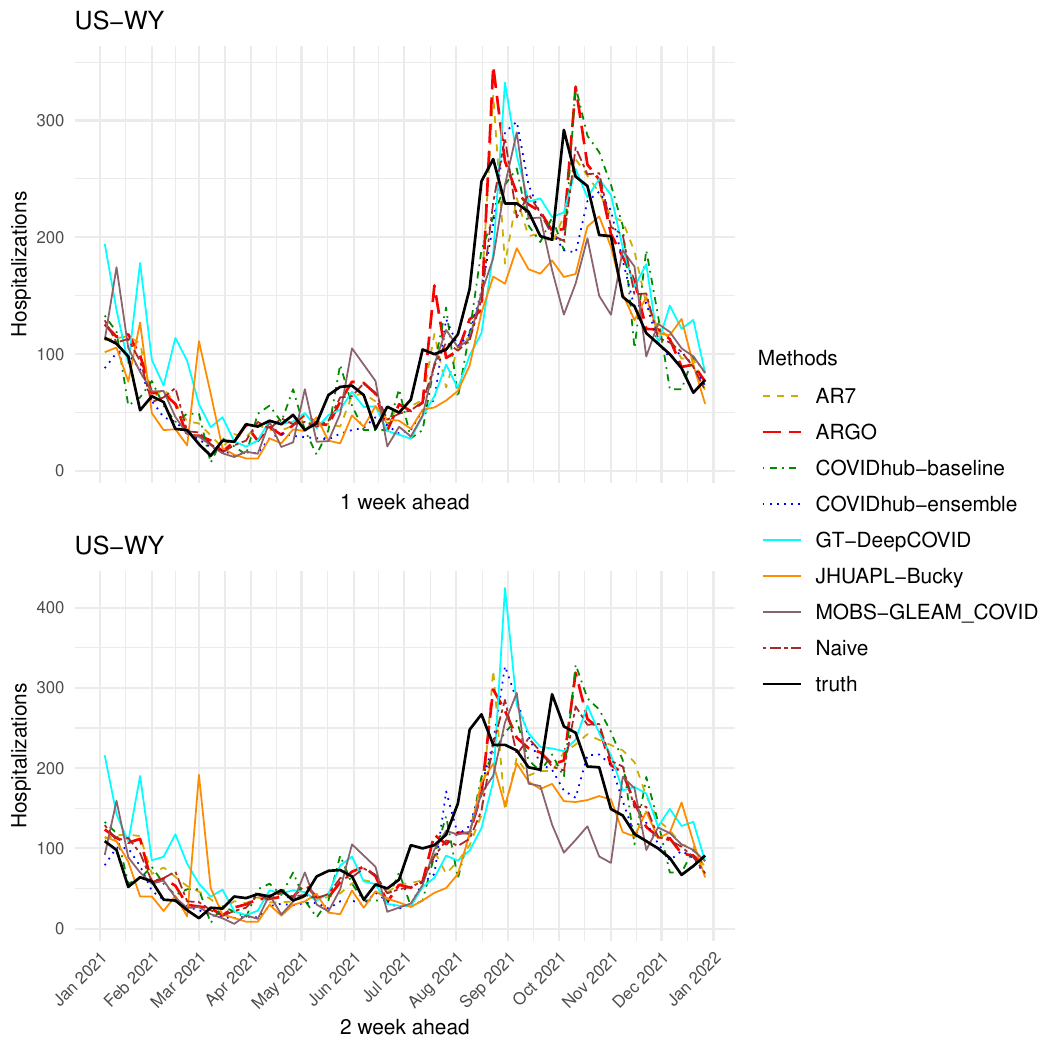} 
\caption{Plots of the COVID-19 hospitalizations 1 week (top), 2 weeks (bottom) ahead estimates of all compared models for New Wyoming (WY).       }
\end{figure}
\newpage
\begin{table}[ht]
\footnotesize
\centering
\begin{tabular}{|l|r|r|r|r|r|r|}
   \hline\multicolumn{1}{|c}{Methods} & \multicolumn{2}{|c}{RMSE} & \multicolumn{2}{|c}{MAE}& \multicolumn{2}{|c|}{Cor} \\  \cline{2-7} & 1 Week Ahead & 2 Week Ahead & 1 Week Ahead & 2 Week Ahead & 1 Week Ahead & 2 Week Ahead \\ 
  \hline
ARGO & \textbf{95.493} & \textbf{170.424} & \textbf{51.479} & \textbf{107.810} & 0.955 & 0.892 \\ 
   \hline
AR7 & 133.967 & 240.808 & 71.178 & 127.096 & 0.907 & 0.768 \\ 
   \hline
COVIDhub-ensemble & 126.391 & 192.531 & 69.904 & 116.577 & 0.923 & 0.871 \\ 
   \hline
Naive & 150.763 & 253.019 & 78.731 & 145.019 & 0.882 & 0.744 \\ 
   \hline
MOBS-GLEAM\_COVID & 95.790 & 175.794 & 75.205 & 113.623 & \textbf{0.956} & \textbf{0.895} \\ 
   \hline
COVIDhub-baseline & 168.271 & 259.005 & 99.231 & 158.096 & 0.851 & 0.727 \\ 
   \hline
GT-DeepCOVID & 174.089 & 266.976 & 99.549 & 158.615 & 0.842 & 0.715 \\ 
   \hline
JHUAPL-Bucky & 139.367 & 188.020 & 85.753 & 129.321 & 0.904 & 0.882 \\ 
   \hline
\end{tabular}
\caption{Comparison of different methods for state-level COVID-19 1 to 2 weeks ahead hospitalizations predictions in Connecticut (CT). The MSE, MAE, and correlation are reported and best performed method is highlighted in boldface.} 
\end{table}
\begin{figure}[!h] 
  \centering 
\includegraphics[width=.8\linewidth, page=34]{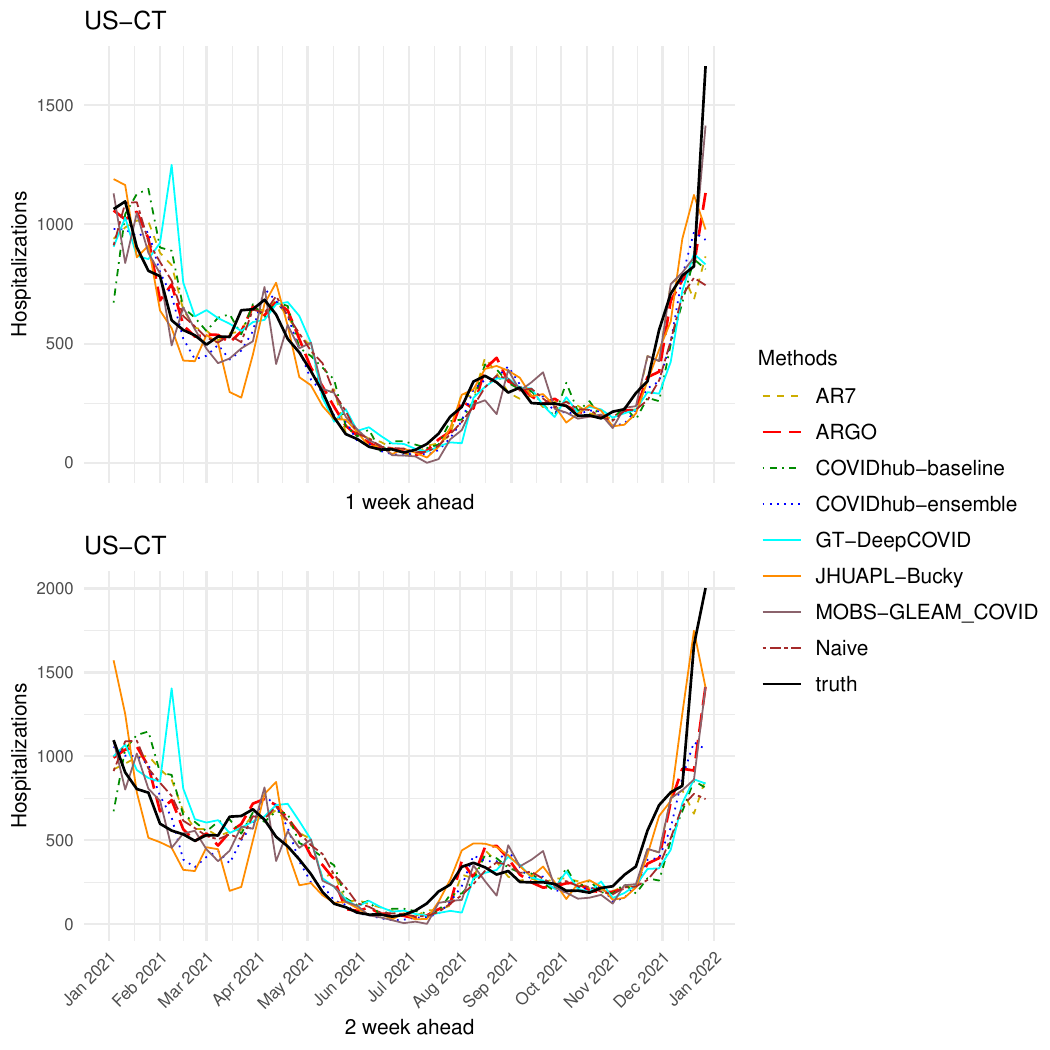} 
\caption{Plots of the COVID-19 hospitalizations 1 week (top), 2 weeks (bottom) ahead estimates of all compared models for New Connecticut (CT).       }
\end{figure}
\newpage
\begin{table}[ht]
\footnotesize
\centering
\begin{tabular}{|l|r|r|r|r|r|r|}
   \hline\multicolumn{1}{|c}{Methods} & \multicolumn{2}{|c}{RMSE} & \multicolumn{2}{|c}{MAE}& \multicolumn{2}{|c|}{Cor} \\  \cline{2-7} & 1 Week Ahead & 2 Week Ahead & 1 Week Ahead & 2 Week Ahead & 1 Week Ahead & 2 Week Ahead \\ 
  \hline
ARGO & \textbf{294.078} & 768.804 & \textbf{201.508} & \textbf{480.730} & \textbf{0.974} & \textbf{0.875} \\ 
   \hline
AR7 & 298.662 & \textbf{650.740} & 223.916 & 505.674 & 0.969 & 0.847 \\ 
   \hline
COVIDhub-ensemble & 477.592 & 780.033 & 305.250 & 500.654 & 0.929 & 0.825 \\ 
   \hline
Naive & 437.563 & 752.901 & 313.827 & 571.346 & 0.926 & 0.790 \\ 
   \hline
MOBS-GLEAM\_COVID & 685.047 & 1351.964 & 426.187 & 750.286 & 0.893 & 0.706 \\ 
   \hline
COVIDhub-baseline & 485.242 & 794.124 & 335.038 & 580.865 & 0.910 & 0.768 \\ 
   \hline
GT-DeepCOVID & 490.180 & 857.363 & 356.471 & 641.452 & 0.919 & 0.752 \\ 
   \hline
JHUAPL-Bucky & 868.019 & 1386.654 & 525.806 & 786.233 & 0.868 & 0.769 \\ 
   \hline
\end{tabular}
\caption{Comparison of different methods for state-level COVID-19 1 to 2 weeks ahead hospitalizations predictions in Michigan (MI). The MSE, MAE, and correlation are reported and best performed method is highlighted in boldface.} 
\end{table}
\begin{figure}[!h] 
  \centering 
\includegraphics[width=.8\linewidth, page=35]{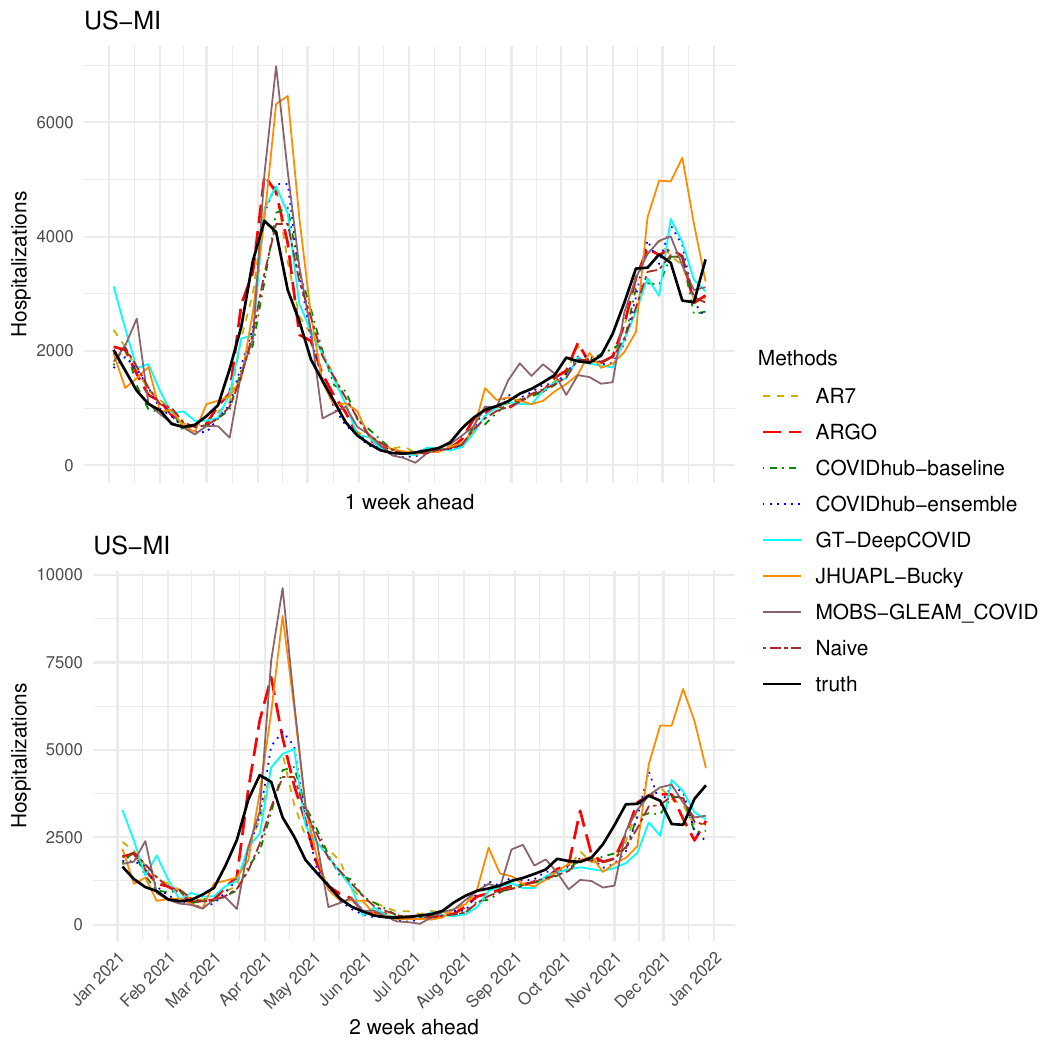} 
\caption{Plots of the COVID-19 hospitalizations 1 week (top), 2 weeks (bottom) ahead estimates of all compared models for New Michigan (MI).        }
\end{figure}
\newpage
\begin{table}[ht]
\footnotesize
\centering
\begin{tabular}{|l|r|r|r|r|r|r|}
   \hline\multicolumn{1}{|c}{Methods} & \multicolumn{2}{|c}{RMSE} & \multicolumn{2}{|c}{MAE}& \multicolumn{2}{|c|}{Cor} \\  \cline{2-7} & 1 Week Ahead & 2 Week Ahead & 1 Week Ahead & 2 Week Ahead & 1 Week Ahead & 2 Week Ahead \\ 
  \hline
ARGO & \textbf{234.507} & 558.415 & \textbf{162.766} & \textbf{297.625} & \textbf{0.971} & 0.881 \\ 
   \hline
AR7 & 260.853 & 628.973 & 165.250 & 348.374 & 0.965 & 0.849 \\ 
   \hline
COVIDhub-ensemble & 409.691 & 625.172 & 208.250 & 344.827 & 0.913 & 0.858 \\ 
   \hline
Naive & 464.845 & 804.344 & 231.269 & 432.423 & 0.886 & 0.739 \\ 
   \hline
MOBS-GLEAM\_COVID & 365.702 & 736.589 & 261.375 & 464.496 & 0.931 & 0.785 \\ 
   \hline
COVIDhub-baseline & 497.509 & 813.902 & 266.000 & 451.769 & 0.866 & 0.730 \\ 
   \hline
GT-DeepCOVID & 342.160 & 684.912 & 238.350 & 447.842 & 0.942 & 0.816 \\ 
   \hline
JHUAPL-Bucky & 407.962 & \textbf{546.555} & 220.364 & 334.957 & 0.910 & \textbf{0.886} \\ 
   \hline
\end{tabular}
\caption{Comparison of different methods for state-level COVID-19 1 to 2 weeks ahead hospitalizations predictions in New Jersey (NJ). The MSE, MAE, and correlation are reported and best performed method is highlighted in boldface.} 
\end{table}
\begin{figure}[!h] 
  \centering 
\includegraphics[width=.8\linewidth, page=36]{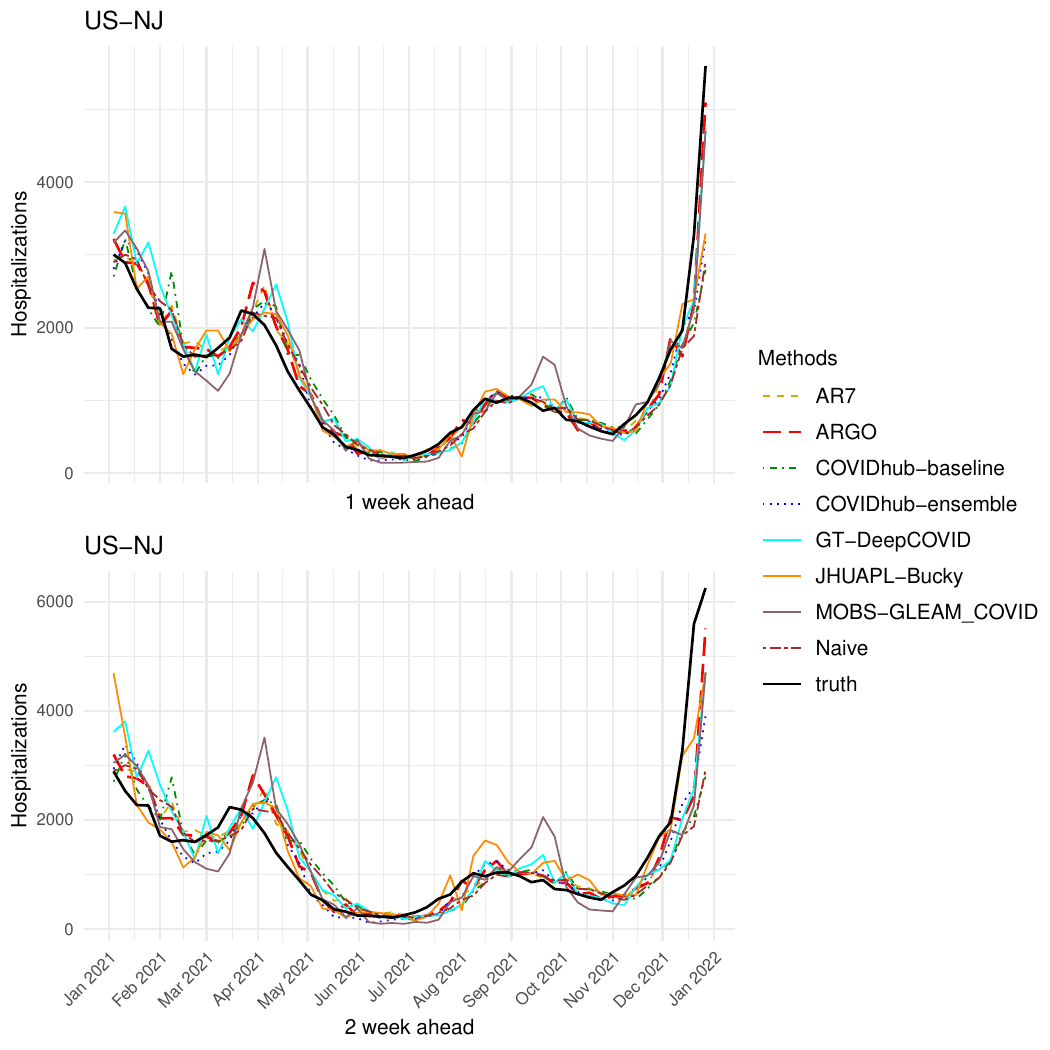} 
\caption{Plots of the COVID-19 hospitalizations 1 week (top), 2 weeks (bottom) ahead estimates of all compared models for New New Jersey (NJ).   }
\end{figure}
\newpage
\begin{table}[ht]
\footnotesize
\centering
\begin{tabular}{|l|r|r|r|r|r|r|}
   \hline\multicolumn{1}{|c}{Methods} & \multicolumn{2}{|c}{RMSE} & \multicolumn{2}{|c}{MAE}& \multicolumn{2}{|c|}{Cor} \\  \cline{2-7} & 1 Week Ahead & 2 Week Ahead & 1 Week Ahead & 2 Week Ahead & 1 Week Ahead & 2 Week Ahead \\ 
  \hline
ARGO & 205.911 & 432.673 & \textbf{148.668} & 310.243 & \textbf{0.969} & 0.862 \\ 
   \hline
AR7 & \textbf{205.837} & 435.616 & 156.040 & 333.226 & \textbf{0.969} & 0.832 \\ 
   \hline
COVIDhub-ensemble & 259.060 & 429.272 & 175.692 & \textbf{286.577} & 0.951 & \textbf{0.872} \\ 
   \hline
Naive & 292.678 & 525.504 & 216.058 & 387.519 & 0.919 & 0.740 \\ 
   \hline
MOBS-GLEAM\_COVID & 270.064 & \textbf{406.380} & 206.227 & 307.255 & 0.938 & 0.870 \\ 
   \hline
COVIDhub-baseline & 572.630 & 600.462 & 329.173 & 394.788 & 0.661 & 0.622 \\ 
   \hline
GT-DeepCOVID & 342.450 & 595.482 & 256.920 & 430.579 & 0.903 & 0.746 \\ 
   \hline
JHUAPL-Bucky & 463.969 & 598.101 & 334.789 & 399.626 & 0.885 & 0.859 \\ 
   \hline
\end{tabular}
\caption{Comparison of different methods for state-level COVID-19 1 to 2 weeks ahead hospitalizations predictions in Tennessee (TN). The MSE, MAE, and correlation are reported and best performed method is highlighted in boldface.} 
\end{table}
\begin{figure}[!h] 
  \centering 
\includegraphics[width=.8\linewidth, page=37]{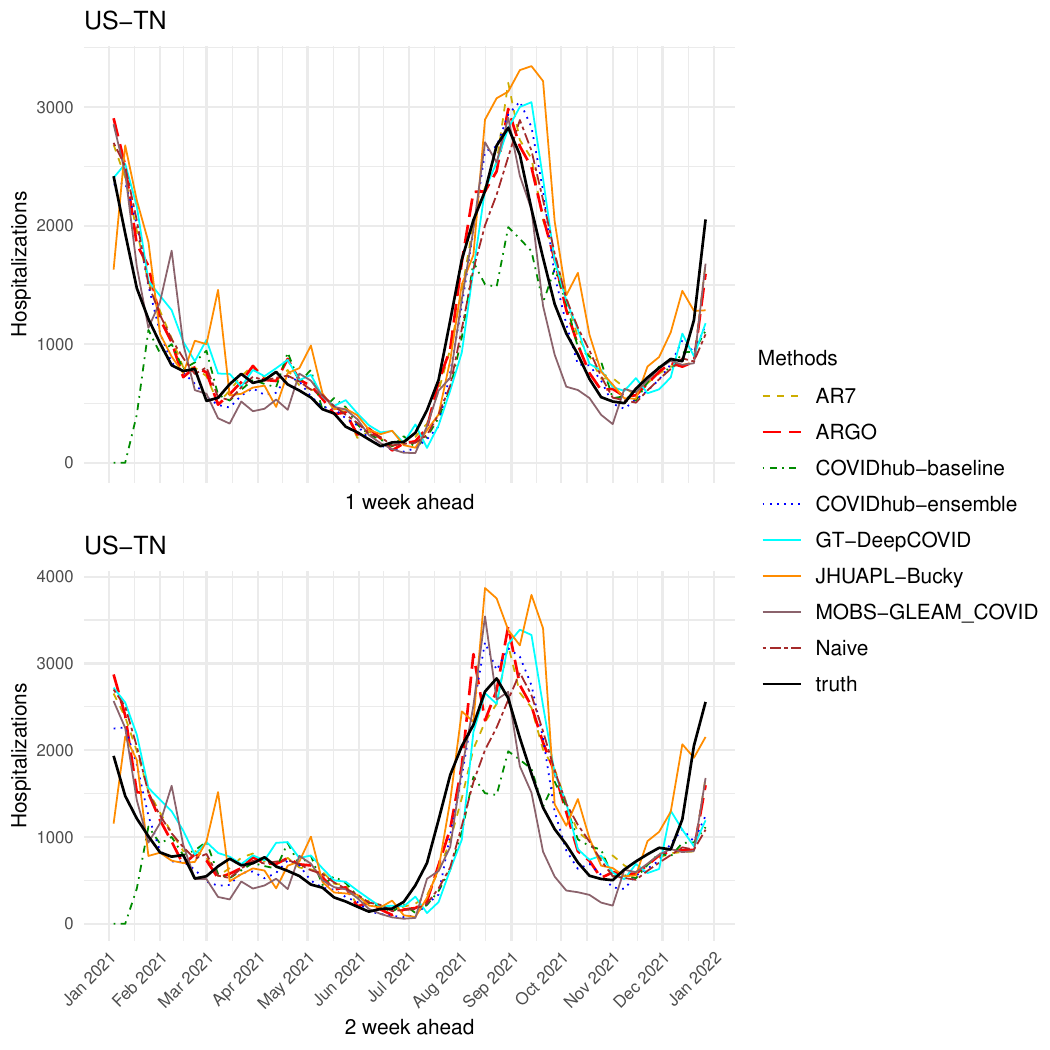} 
\caption{Plots of the COVID-19 hospitalizations 1 week (top), 2 weeks (bottom) ahead estimates of all compared models for New Tennessee (TN).   }
\end{figure}
\newpage
\begin{table}[ht]
\footnotesize
\centering
\begin{tabular}{|l|r|r|r|r|r|r|}
   \hline\multicolumn{1}{|c}{Methods} & \multicolumn{2}{|c}{RMSE} & \multicolumn{2}{|c}{MAE}& \multicolumn{2}{|c|}{Cor} \\  \cline{2-7} & 1 Week Ahead & 2 Week Ahead & 1 Week Ahead & 2 Week Ahead & 1 Week Ahead & 2 Week Ahead \\ 
  \hline
ARGO & \textbf{164.989} & 379.966 & \textbf{109.513} & \textbf{238.075} & \textbf{0.987} & \textbf{0.927} \\ 
   \hline
AR7 & 175.915 & \textbf{375.124} & 123.950 & 246.940 & 0.983 & 0.912 \\ 
   \hline
COVIDhub-ensemble & 258.440 & 461.810 & 160.558 & 266.481 & 0.974 & 0.915 \\ 
   \hline
Naive & 252.887 & 463.828 & 171.808 & 303.385 & 0.967 & 0.881 \\ 
   \hline
MOBS-GLEAM\_COVID & 360.336 & 586.952 & 257.392 & 417.970 & 0.949 & 0.864 \\ 
   \hline
COVIDhub-baseline & 261.054 & 445.984 & 179.038 & 296.962 & 0.963 & 0.887 \\ 
   \hline
GT-DeepCOVID & 228.603 & 435.147 & 183.521 & 332.371 & 0.971 & 0.900 \\ 
   \hline
JHUAPL-Bucky & 440.951 & 693.221 & 286.634 & 437.574 & 0.947 & 0.901 \\ 
   \hline
\end{tabular}
\caption{Comparison of different methods for state-level COVID-19 1 to 2 weeks ahead hospitalizations predictions in Arizona (AZ). The MSE, MAE, and correlation are reported and best performed method is highlighted in boldface.} 
\end{table}
\begin{figure}[!h] 
  \centering 
\includegraphics[width=.8\linewidth, page=38]{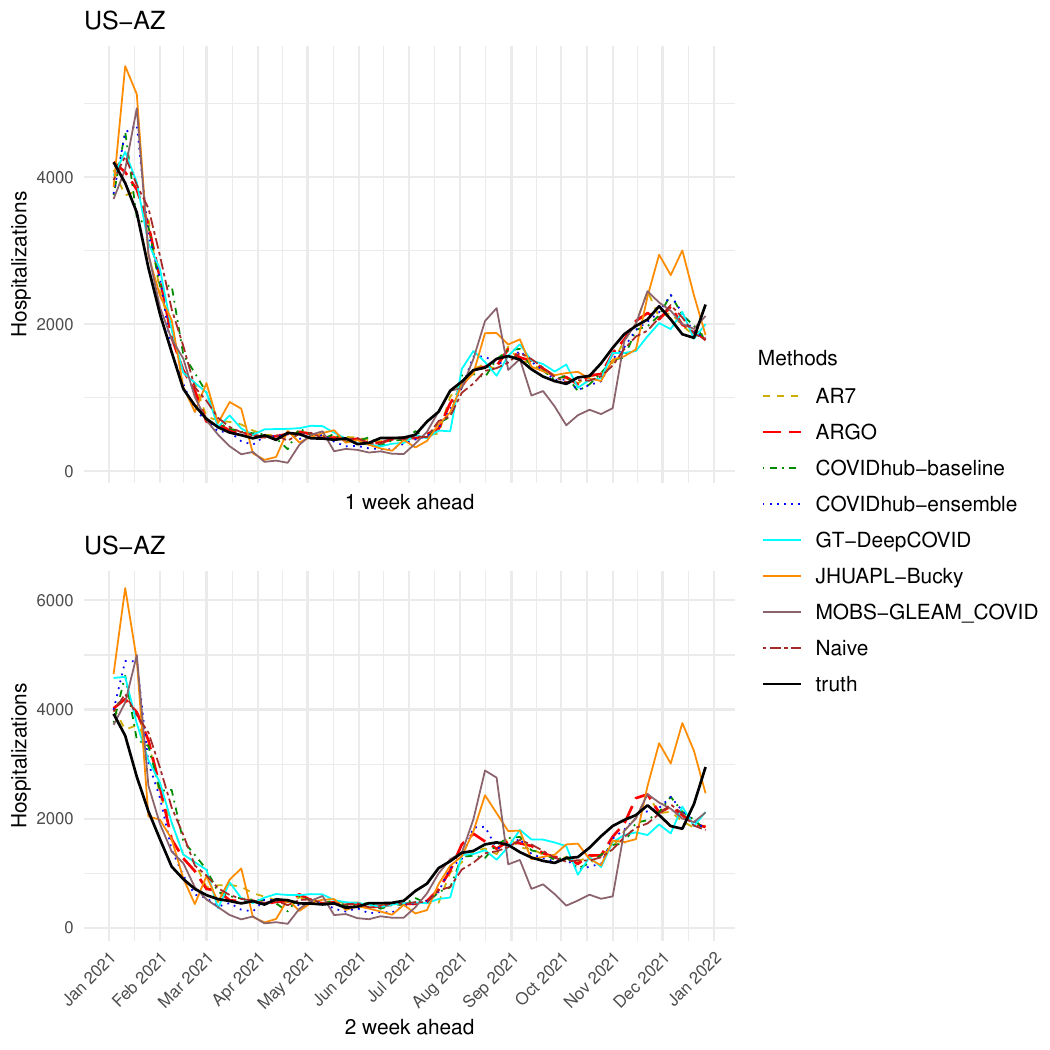} 
\caption{Plots of the COVID-19 hospitalizations 1 week (top), 2 weeks (bottom) ahead estimates of all compared models for New Arizona (AZ).    }
\end{figure}
\newpage
\begin{table}[ht]
\footnotesize
\centering
\begin{tabular}{|l|r|r|r|r|r|r|}
   \hline\multicolumn{1}{|c}{Methods} & \multicolumn{2}{|c}{RMSE} & \multicolumn{2}{|c}{MAE}& \multicolumn{2}{|c|}{Cor} \\  \cline{2-7} & 1 Week Ahead & 2 Week Ahead & 1 Week Ahead & 2 Week Ahead & 1 Week Ahead & 2 Week Ahead \\ 
  \hline
ARGO & \textbf{694.280} & 1926.407 & \textbf{475.535} & \textbf{1275.348} & \textbf{0.989} & \textbf{0.893} \\ 
   \hline
AR7 & 803.686 & 1936.963 & 610.086 & 1350.937 & 0.978 & 0.862 \\ 
   \hline
COVIDhub-ensemble & 1420.437 & 2426.737 & 932.019 & 1570.846 & 0.941 & 0.844 \\ 
   \hline
Naive & 1350.872 & 2510.225 & 924.769 & 1709.962 & 0.933 & 0.769 \\ 
   \hline
MOBS-GLEAM\_COVID & 1017.099 & \textbf{1843.362} & 749.273 & 1309.635 & 0.963 & 0.883 \\ 
   \hline
COVIDhub-baseline & 1487.790 & 2609.230 & 1012.923 & 1809.846 & 0.920 & 0.752 \\ 
   \hline
GT-DeepCOVID & 1409.907 & 2480.871 & 986.914 & 1726.428 & 0.927 & 0.792 \\ 
   \hline
JHUAPL-Bucky & 2758.649 & 3793.091 & 1942.424 & 2685.922 & 0.836 & 0.770 \\ 
   \hline
\end{tabular}
\caption{Comparison of different methods for state-level COVID-19 1 to 2 weeks ahead hospitalizations predictions in Texas (TX). The MSE, MAE, and correlation are reported and best performed method is highlighted in boldface.} 
\end{table}
\begin{figure}[!h] 
  \centering 
\includegraphics[width=.8\linewidth, page=39]{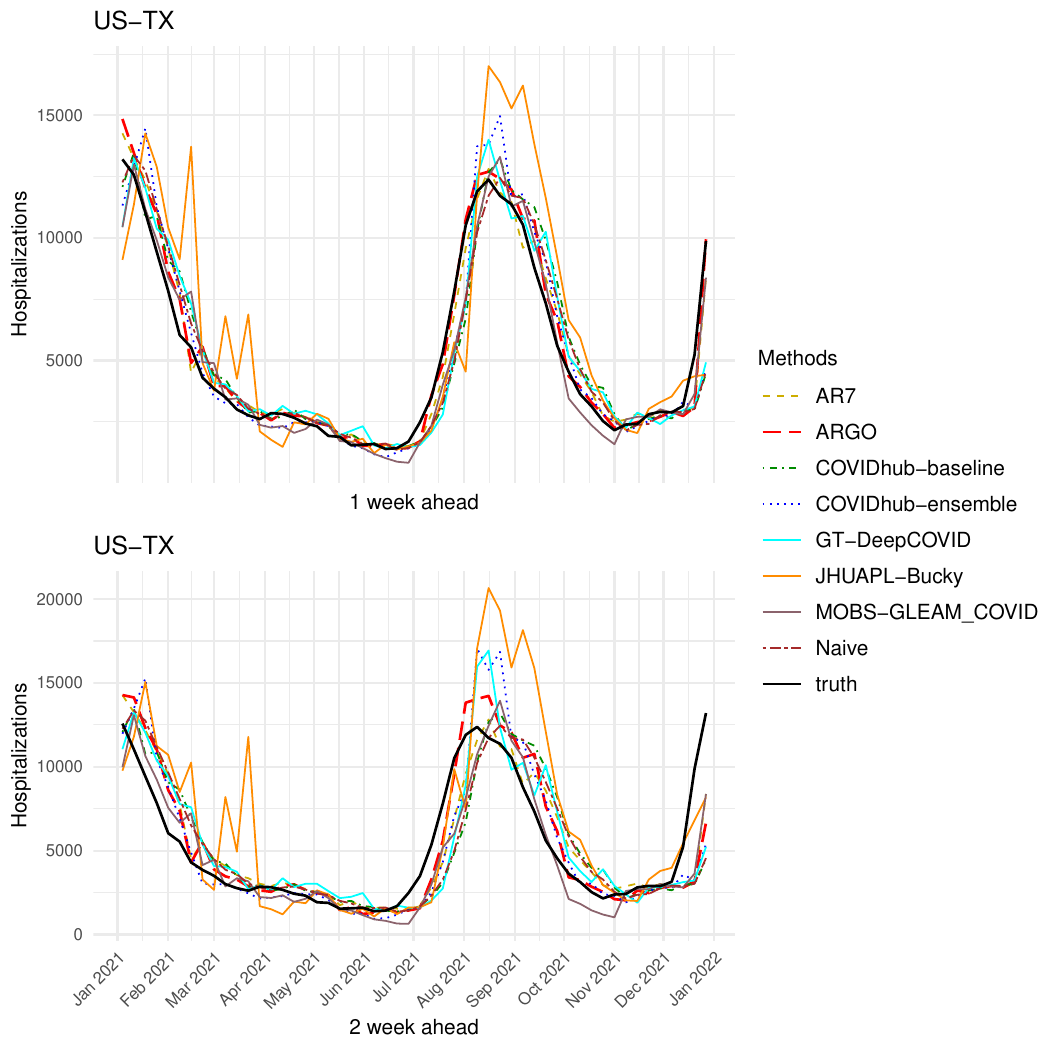} 
\caption{Plots of the COVID-19 hospitalizations 1 week (top), 2 weeks (bottom) ahead estimates of all compared models for New Texas (TX).    }
\end{figure}
\newpage
\begin{table}[ht]
\footnotesize
\centering
\begin{tabular}{|l|r|r|r|r|r|r|}
   \hline\multicolumn{1}{|c}{Methods} & \multicolumn{2}{|c}{RMSE} & \multicolumn{2}{|c}{MAE}& \multicolumn{2}{|c|}{Cor} \\  \cline{2-7} & 1 Week Ahead & 2 Week Ahead & 1 Week Ahead & 2 Week Ahead & 1 Week Ahead & 2 Week Ahead \\ 
  \hline
ARGO & \textbf{558.546} & \textbf{1630.819} & \textbf{330.790} & \textbf{896.587} & \textbf{0.990} & \textbf{0.895} \\ 
   \hline
AR7 & 713.487 & 1783.270 & 460.100 & 1099.391 & 0.982 & 0.857 \\ 
   \hline
COVIDhub-ensemble & 1269.816 & 2349.889 & 725.038 & 1250.712 & 0.960 & 0.833 \\ 
   \hline
Naive & 1117.565 & 2239.017 & 698.269 & 1366.692 & 0.956 & 0.794 \\ 
   \hline
MOBS-GLEAM\_COVID & 1430.195 & 2242.263 & 1001.432 & 1500.074 & 0.929 & 0.800 \\ 
   \hline
COVIDhub-baseline & 1043.072 & 2146.711 & 667.115 & 1341.577 & 0.957 & 0.797 \\ 
   \hline
GT-DeepCOVID & 1327.192 & 2632.167 & 776.532 & 1652.832 & 0.950 & 0.801 \\ 
   \hline
JHUAPL-Bucky & 3425.902 & 4010.325 & 1636.756 & 2086.601 & 0.907 & 0.821 \\ 
   \hline
\end{tabular}
\caption{Comparison of different methods for state-level COVID-19 1 to 2 weeks ahead hospitalizations predictions in California (CA). The MSE, MAE, and correlation are reported and best performed method is highlighted in boldface.} 
\end{table}
\begin{figure}[!h] 
  \centering 
\includegraphics[width=.8\linewidth, page=40]{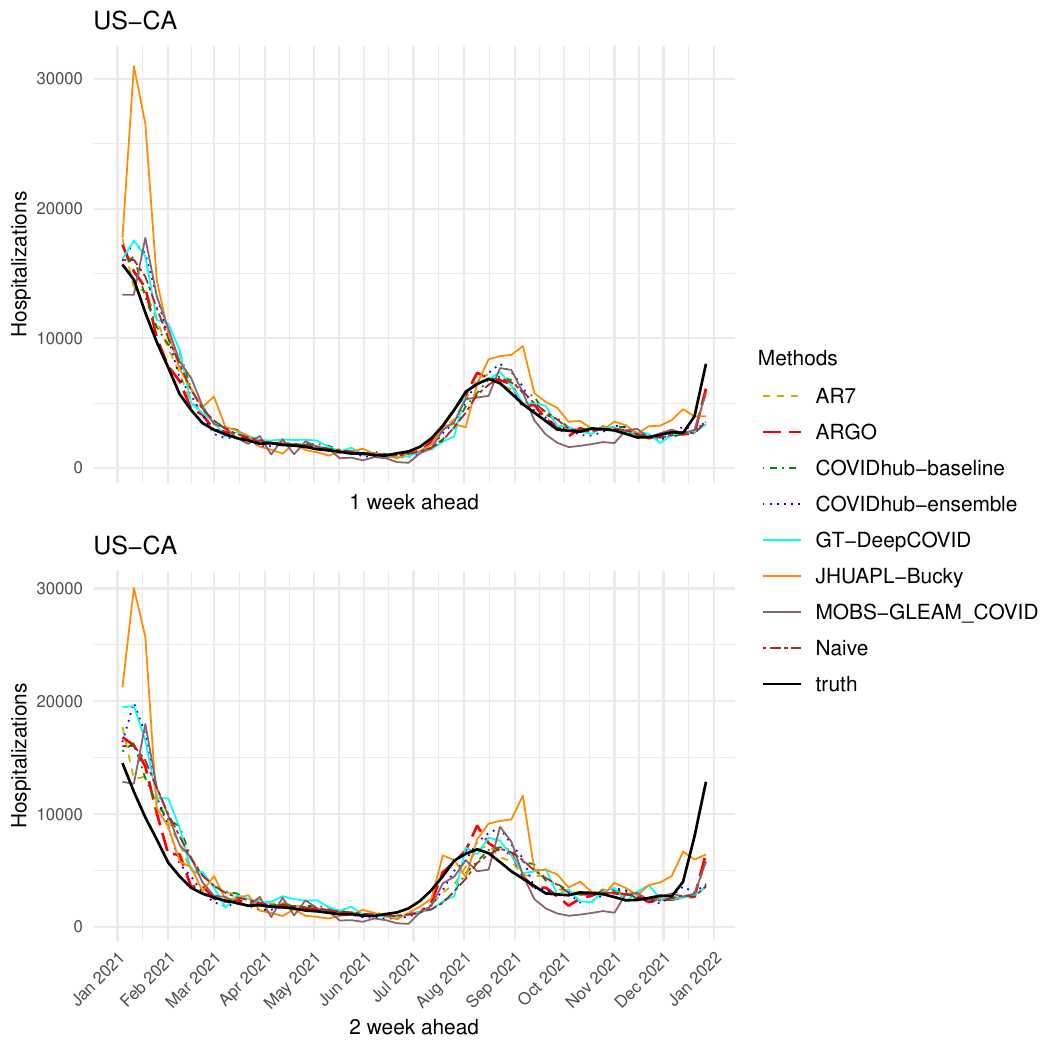} 
\caption{Plots of the COVID-19 hospitalizations 1 week (top), 2 weeks (bottom) ahead estimates of all compared models for New California (CA).     }
\end{figure}
\newpage
\begin{table}[ht]
\footnotesize
\centering
\begin{tabular}{|l|r|r|r|r|r|r|}
   \hline\multicolumn{1}{|c}{Methods} & \multicolumn{2}{|c}{RMSE} & \multicolumn{2}{|c}{MAE}& \multicolumn{2}{|c|}{Cor} \\  \cline{2-7} & 1 Week Ahead & 2 Week Ahead & 1 Week Ahead & 2 Week Ahead & 1 Week Ahead & 2 Week Ahead \\ 
  \hline
ARGO & \textbf{259.858} & \textbf{564.317} & \textbf{191.719} & 409.134 & \textbf{0.974} & \textbf{0.894} \\ 
   \hline
AR7 & 333.522 & 649.426 & 229.746 & 481.142 & 0.957 & 0.856 \\ 
   \hline
COVIDhub-ensemble & 407.388 & 646.373 & 245.327 & \textbf{394.269} & 0.935 & 0.863 \\ 
   \hline
Naive & 404.332 & 703.372 & 289.077 & 527.423 & 0.935 & 0.830 \\ 
   \hline
MOBS-GLEAM\_COVID & 392.792 & 721.467 & 298.344 & 518.342 & 0.947 & 0.844 \\ 
   \hline
COVIDhub-baseline & 467.962 & 732.576 & 308.154 & 503.038 & 0.912 & 0.815 \\ 
   \hline
GT-DeepCOVID & 560.011 & 874.646 & 365.580 & 586.065 & 0.881 & 0.752 \\ 
   \hline
JHUAPL-Bucky & 647.529 & 965.349 & 433.610 & 648.434 & 0.910 & 0.881 \\ 
   \hline
\end{tabular}
\caption{Comparison of different methods for state-level COVID-19 1 to 2 weeks ahead hospitalizations predictions in Pennsylvania (PA). The MSE, MAE, and correlation are reported and best performed method is highlighted in boldface.} 
\end{table}
\begin{figure}[!h] 
  \centering 
\includegraphics[width=.8\linewidth, page=41]{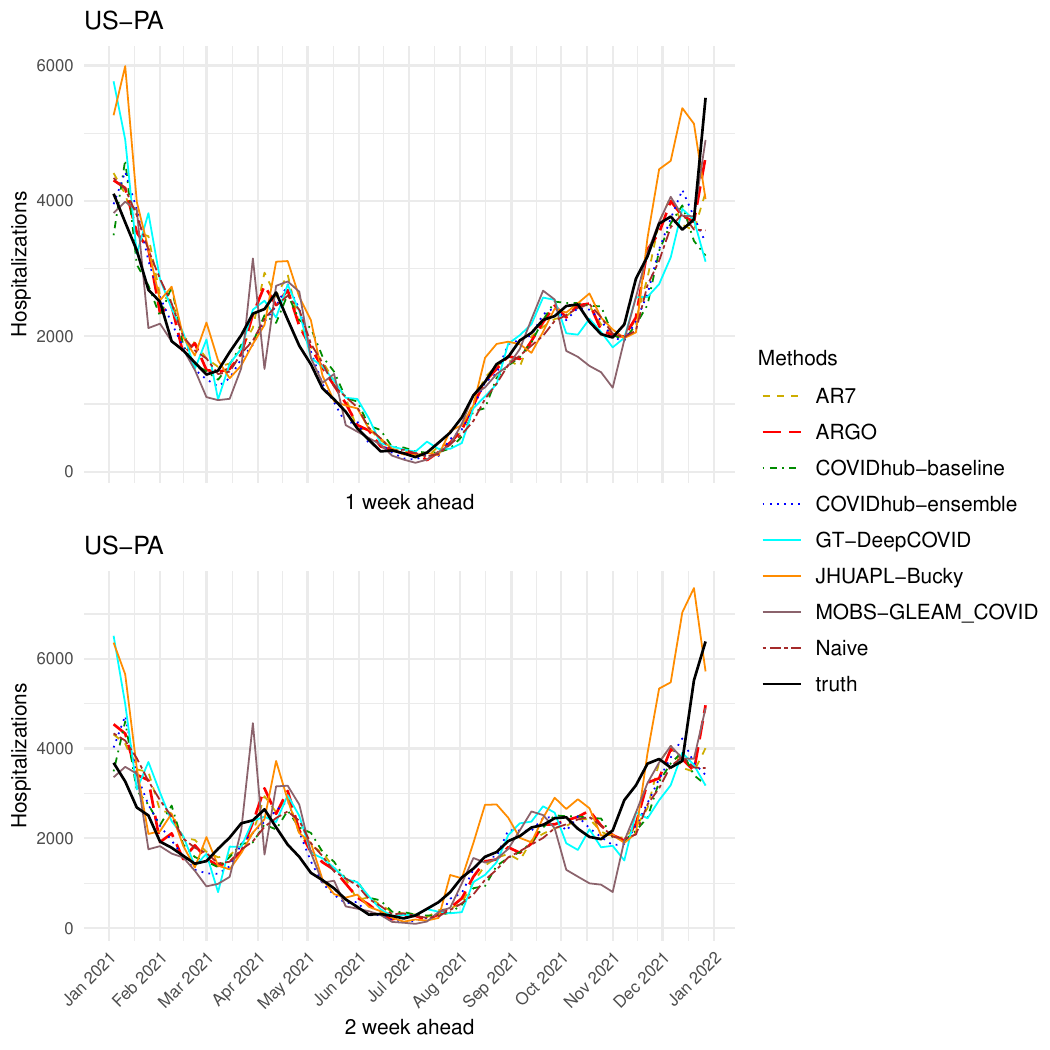} 
\caption{Plots of the COVID-19 hospitalizations 1 week (top), 2 weeks (bottom) ahead estimates of all compared models for New Pennsylvania (PA).     }
\end{figure}
\newpage
\begin{table}[ht]
\footnotesize
\centering
\begin{tabular}{|l|r|r|r|r|r|r|}
   \hline\multicolumn{1}{|c}{Methods} & \multicolumn{2}{|c}{RMSE} & \multicolumn{2}{|c}{MAE}& \multicolumn{2}{|c|}{Cor} \\  \cline{2-7} & 1 Week Ahead & 2 Week Ahead & 1 Week Ahead & 2 Week Ahead & 1 Week Ahead & 2 Week Ahead \\ 
  \hline
ARGO & 215.699 & \textbf{434.108} & \textbf{159.364} & \textbf{330.164} & 0.979 & 0.929 \\ 
   \hline
AR7 & \textbf{204.355} & 504.545 & 163.034 & 393.979 & \textbf{0.980} & 0.912 \\ 
   \hline
COVIDhub-ensemble & 312.647 & 482.454 & 219.981 & 357.577 & 0.956 & 0.924 \\ 
   \hline
Naive & 376.284 & 679.884 & 262.000 & 493.308 & 0.935 & 0.839 \\ 
   \hline
MOBS-GLEAM\_COVID & 303.902 & 564.891 & 247.513 & 439.540 & 0.963 & 0.913 \\ 
   \hline
COVIDhub-baseline & 398.236 & 670.867 & 260.673 & 473.404 & 0.925 & 0.847 \\ 
   \hline
GT-DeepCOVID & 446.429 & 718.916 & 338.989 & 576.896 & 0.891 & 0.790 \\ 
   \hline
JHUAPL-Bucky & 325.060 & 537.848 & 258.863 & 424.967 & 0.948 & \textbf{0.949} \\ 
   \hline
\end{tabular}
\caption{Comparison of different methods for state-level COVID-19 1 to 2 weeks ahead hospitalizations predictions in Illinois (IL). The MSE, MAE, and correlation are reported and best performed method is highlighted in boldface.} 
\end{table}
\begin{figure}[!h] 
  \centering 
\includegraphics[width=.8\linewidth, page=42]{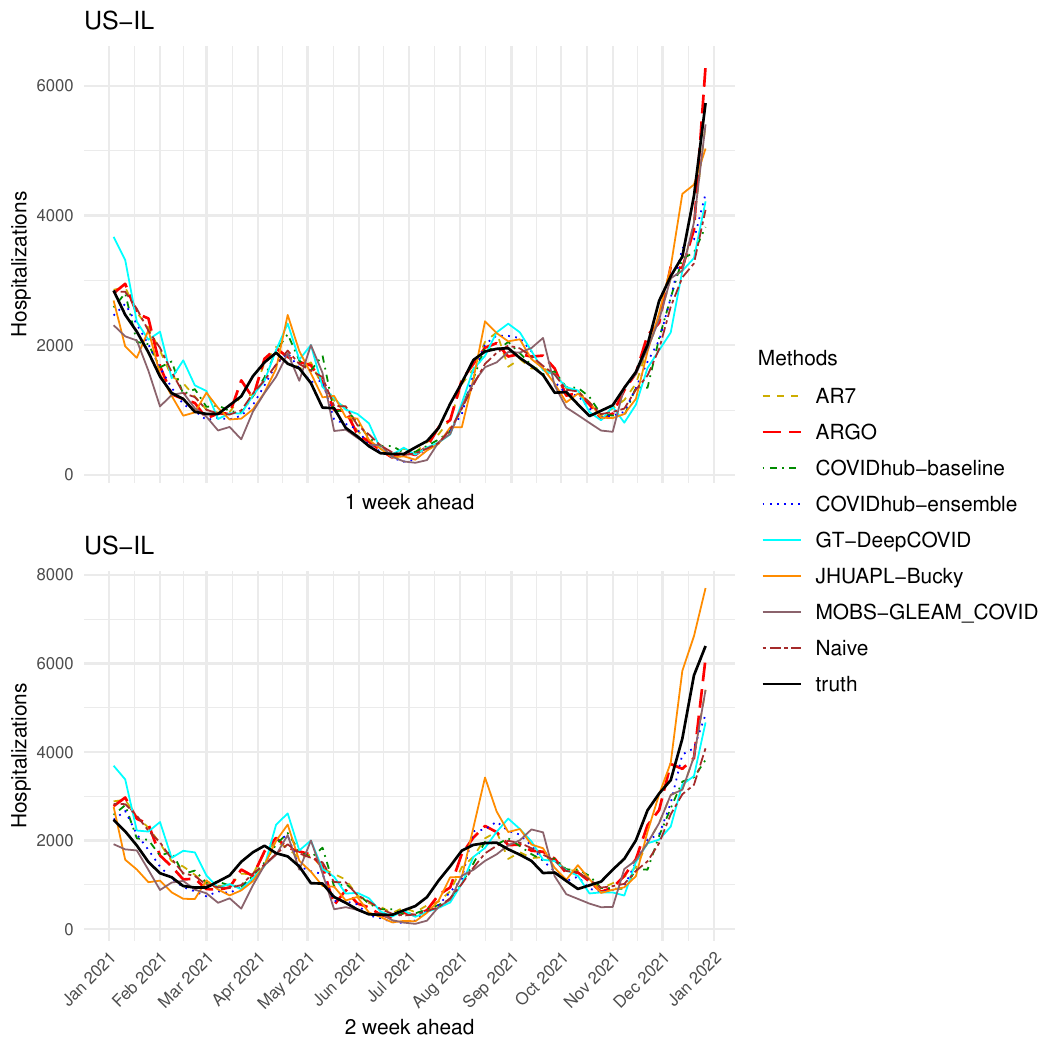} 
\caption{Plots of the COVID-19 hospitalizations 1 week (top), 2 weeks (bottom) ahead estimates of all compared models for New Illinois (IL).    }
\end{figure}
\newpage
\begin{table}[ht]
\footnotesize
\centering
\begin{tabular}{|l|r|r|r|r|r|r|}
   \hline\multicolumn{1}{|c}{Methods} & \multicolumn{2}{|c}{RMSE} & \multicolumn{2}{|c}{MAE}& \multicolumn{2}{|c|}{Cor} \\  \cline{2-7} & 1 Week Ahead & 2 Week Ahead & 1 Week Ahead & 2 Week Ahead & 1 Week Ahead & 2 Week Ahead \\ 
  \hline
ARGO & 199.557 & 487.760 & 132.192 & 307.975 & 0.971 & 0.814 \\ 
   \hline
AR7 & \textbf{169.264} & \textbf{437.288} & \textbf{117.656} & \textbf{289.906} & \textbf{0.973} & 0.812 \\ 
   \hline
COVIDhub-ensemble & 355.698 & 595.532 & 209.231 & 325.673 & 0.908 & 0.774 \\ 
   \hline
Naive & 312.898 & 567.201 & 198.788 & 360.615 & 0.905 & 0.695 \\ 
   \hline
MOBS-GLEAM\_COVID & 351.671 & 604.996 & 197.005 & 327.721 & 0.944 & \textbf{0.845} \\ 
   \hline
COVIDhub-baseline & 334.153 & 571.009 & 219.173 & 363.423 & 0.889 & 0.684 \\ 
   \hline
GT-DeepCOVID & 356.651 & 673.241 & 236.374 & 421.725 & 0.893 & 0.696 \\ 
   \hline
JHUAPL-Bucky & 461.199 & 719.550 & 277.292 & 420.004 & 0.871 & 0.734 \\ 
   \hline
\end{tabular}
\caption{Comparison of different methods for state-level COVID-19 1 to 2 weeks ahead hospitalizations predictions in Louisiana (LA). The MSE, MAE, and correlation are reported and best performed method is highlighted in boldface.} 
\end{table}
\begin{figure}[!h] 
  \centering 
\includegraphics[width=.8\linewidth, page=43]{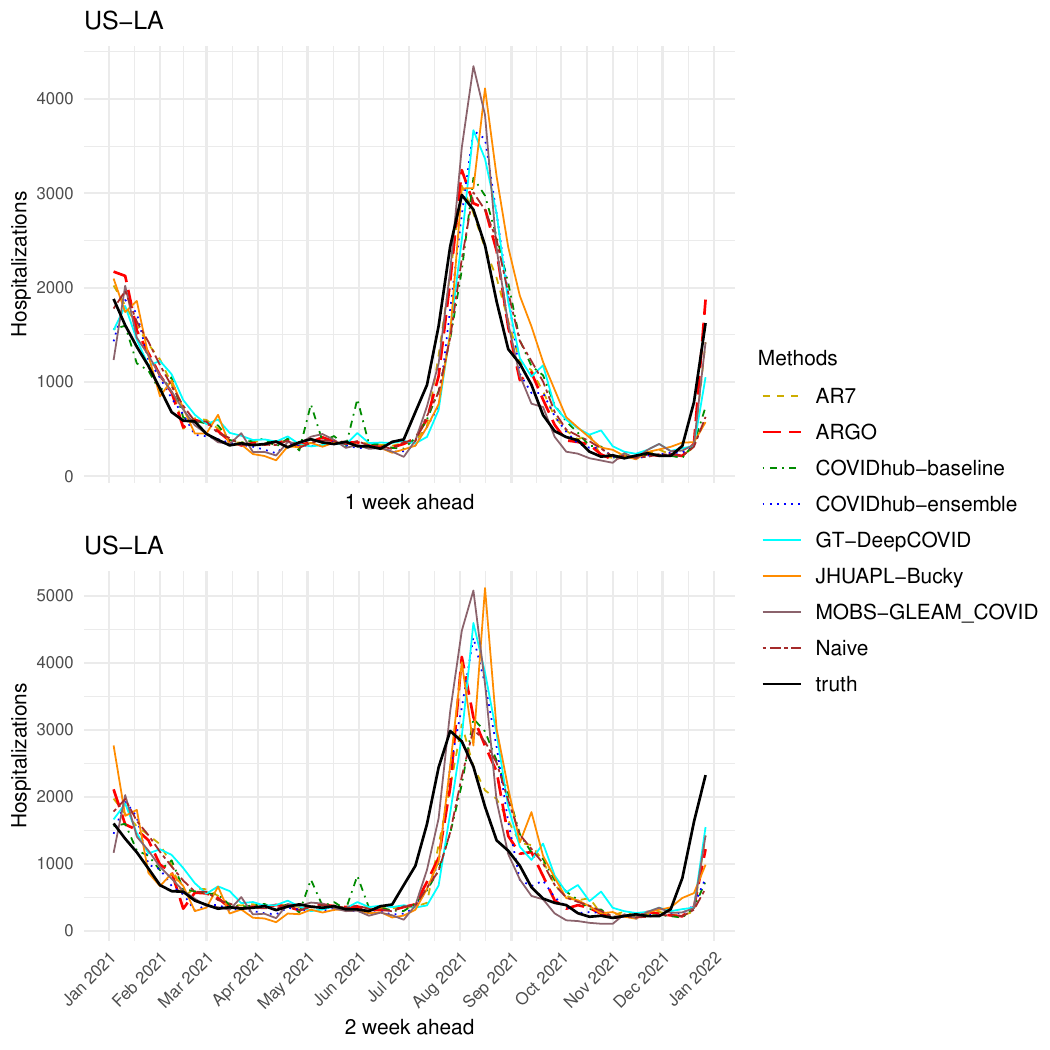} 
\caption{Plots of the COVID-19 hospitalizations 1 week (top), 2 weeks (bottom) ahead estimates of all compared models for New Louisiana (LA).   }
\end{figure}
\newpage
\begin{table}[ht]
\footnotesize
\centering
\begin{tabular}{|l|r|r|r|r|r|r|}
   \hline\multicolumn{1}{|c}{Methods} & \multicolumn{2}{|c}{RMSE} & \multicolumn{2}{|c}{MAE}& \multicolumn{2}{|c|}{Cor} \\  \cline{2-7} & 1 Week Ahead & 2 Week Ahead & 1 Week Ahead & 2 Week Ahead & 1 Week Ahead & 2 Week Ahead \\ 
  \hline
ARGO & \textbf{158.348} & 349.934 & \textbf{113.063} & 229.415 & \textbf{0.973} & 0.895 \\ 
   \hline
AR7 & 173.657 & 380.119 & 127.918 & 272.105 & 0.968 & 0.833 \\ 
   \hline
COVIDhub-ensemble & 236.207 & 396.143 & 153.269 & 248.327 & 0.946 & 0.864 \\ 
   \hline
Naive & 238.203 & 443.930 & 162.558 & 307.288 & 0.932 & 0.766 \\ 
   \hline
MOBS-GLEAM\_COVID & 198.951 & \textbf{318.386} & 134.317 & \textbf{206.750} & 0.956 & \textbf{0.896} \\ 
   \hline
COVIDhub-baseline & 263.125 & 454.578 & 185.192 & 322.231 & 0.913 & 0.742 \\ 
   \hline
GT-DeepCOVID & 297.477 & 519.232 & 207.885 & 352.984 & 0.917 & 0.788 \\ 
   \hline
JHUAPL-Bucky & 340.942 & 508.695 & 229.046 & 341.596 & 0.916 & 0.852 \\ 
   \hline
\end{tabular}
\caption{Comparison of different methods for state-level COVID-19 1 to 2 weeks ahead hospitalizations predictions in South Carolina (SC). The MSE, MAE, and correlation are reported and best performed method is highlighted in boldface.} 
\end{table}
\begin{figure}[!h] 
  \centering 
\includegraphics[width=.8\linewidth, page=44]{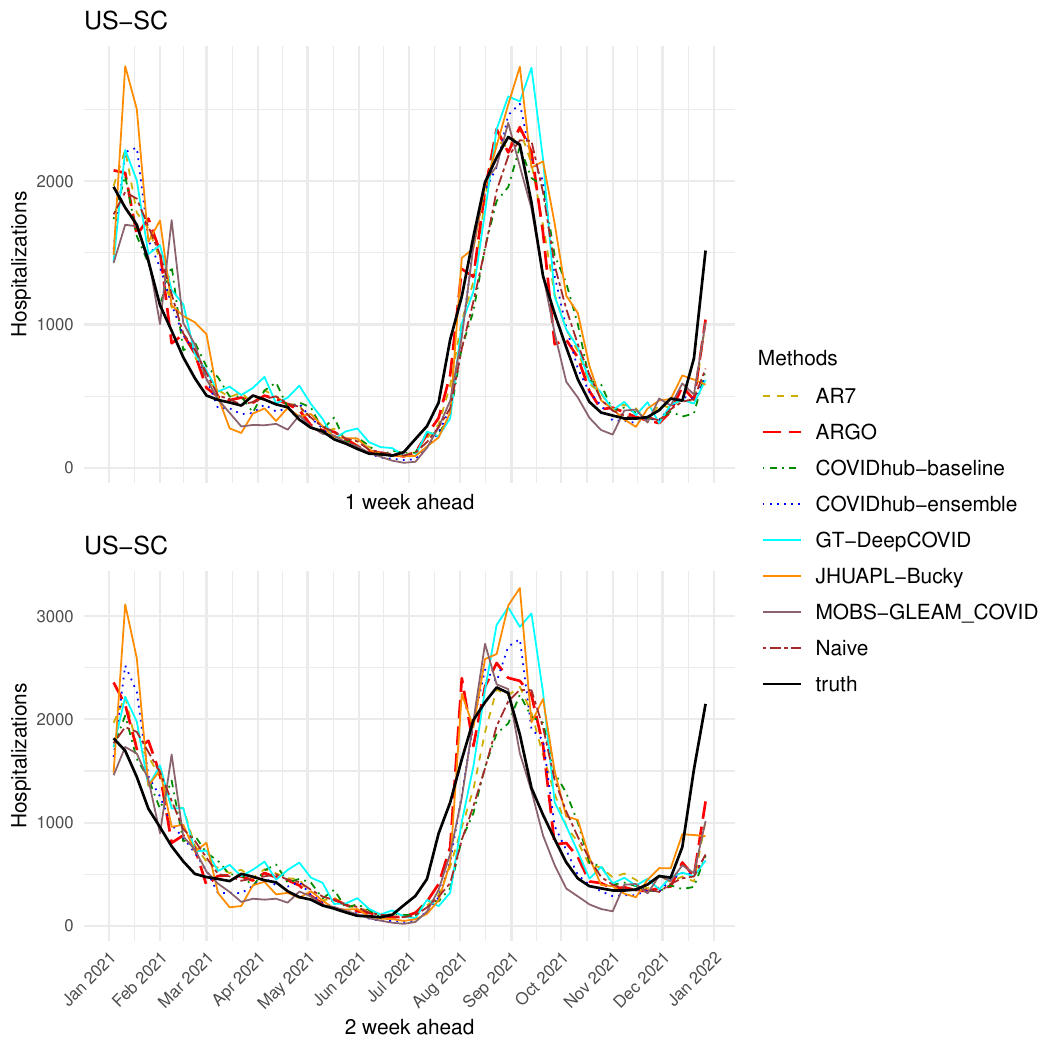} 
\caption{Plots of the COVID-19 hospitalizations 1 week (top), 2 weeks (bottom) ahead estimates of all compared models for New South Carolina (SC).   }
\end{figure}
\newpage
\begin{table}[ht]
\footnotesize
\centering
\begin{tabular}{|l|r|r|r|r|r|r|}
   \hline\multicolumn{1}{|c}{Methods} & \multicolumn{2}{|c}{RMSE} & \multicolumn{2}{|c}{MAE}& \multicolumn{2}{|c|}{Cor} \\  \cline{2-7} & 1 Week Ahead & 2 Week Ahead & 1 Week Ahead & 2 Week Ahead & 1 Week Ahead & 2 Week Ahead \\ 
  \hline
ARGO & \textbf{100.711} & \textbf{177.897} & \textbf{68.874} & \textbf{121.079} & \textbf{0.928} & \textbf{0.824} \\ 
   \hline
AR7 & 115.458 & 205.272 & 79.173 & 136.253 & 0.907 & 0.737 \\ 
   \hline
COVIDhub-ensemble & 118.433 & 199.314 & 72.231 & 123.615 & 0.904 & 0.784 \\ 
   \hline
Naive & 123.894 & 221.467 & 83.212 & 148.192 & 0.885 & 0.686 \\ 
   \hline
MOBS-GLEAM\_COVID & 200.373 & 287.254 & 148.837 & 205.179 & 0.702 & 0.500 \\ 
   \hline
COVIDhub-baseline & 146.191 & 214.758 & 106.481 & 142.038 & 0.833 & 0.695 \\ 
   \hline
GT-DeepCOVID & 138.220 & 243.580 & 103.731 & 180.514 & 0.877 & 0.710 \\ 
   \hline
JHUAPL-Bucky & 281.425 & 593.154 & 207.012 & 373.554 & 0.887 & 0.781 \\ 
   \hline
\end{tabular}
\caption{Comparison of different methods for state-level COVID-19 1 to 2 weeks ahead hospitalizations predictions in Washington (WA). The MSE, MAE, and correlation are reported and best performed method is highlighted in boldface.} 
\end{table}
\begin{figure}[!h] 
  \centering 
\includegraphics[width=.8\linewidth, page=45]{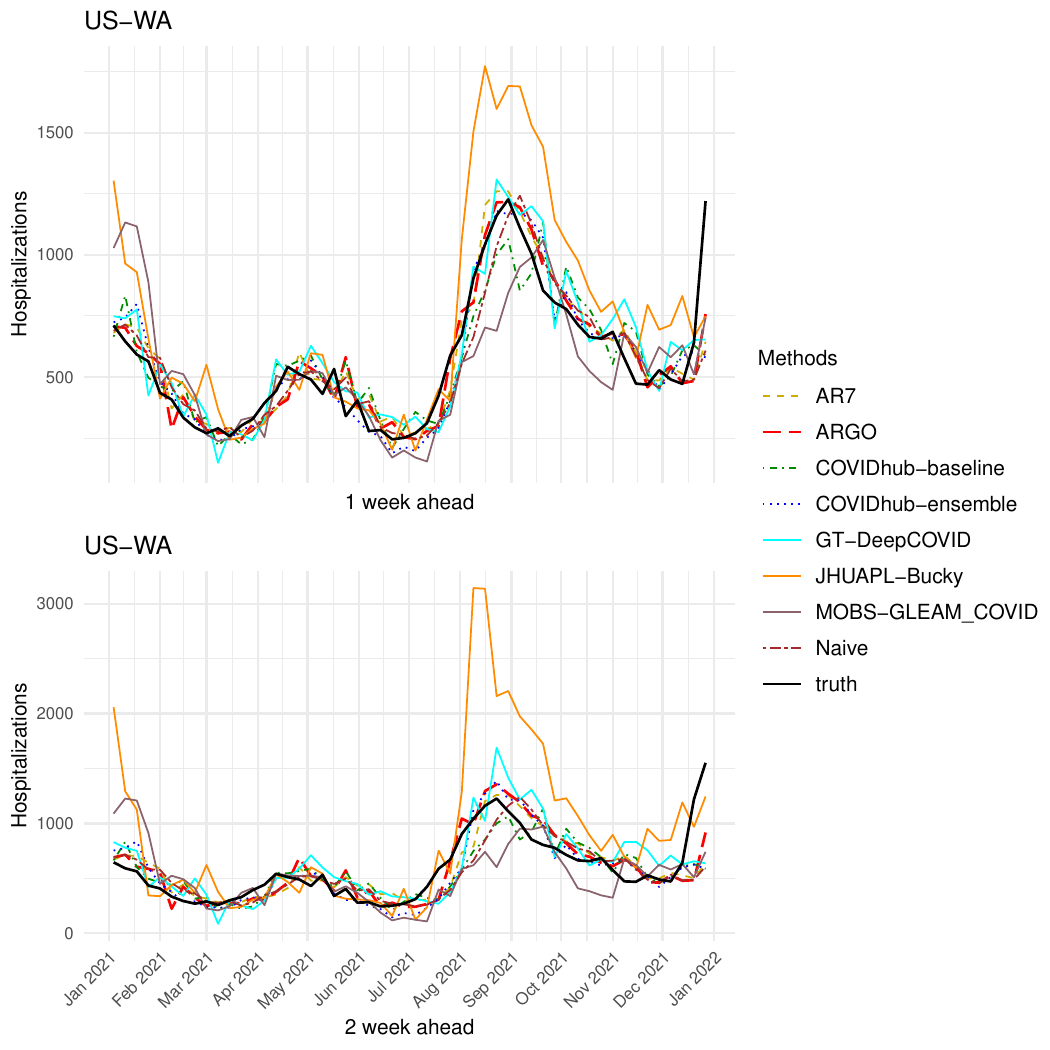} 
\caption{Plots of the COVID-19 hospitalizations 1 week (top), 2 weeks (bottom) ahead estimates of all compared models for New Washington (WA).   }
\end{figure}
\newpage
\begin{table}[ht]
\footnotesize
\centering
\begin{tabular}{|l|r|r|r|r|r|r|}
   \hline\multicolumn{1}{|c}{Methods} & \multicolumn{2}{|c}{RMSE} & \multicolumn{2}{|c}{MAE}& \multicolumn{2}{|c|}{Cor} \\  \cline{2-7} & 1 Week Ahead & 2 Week Ahead & 1 Week Ahead & 2 Week Ahead & 1 Week Ahead & 2 Week Ahead \\ 
  \hline
ARGO & \textbf{123.723} & 250.794 & \textbf{78.795} & \textbf{151.552} & \textbf{0.975} & 0.914 \\ 
   \hline
AR7 & 137.437 & 323.873 & 99.521 & 207.547 & 0.966 & 0.855 \\ 
   \hline
COVIDhub-ensemble & 237.464 & 366.832 & 124.250 & 203.788 & 0.900 & 0.818 \\ 
   \hline
Naive & 244.221 & 421.129 & 141.635 & 253.365 & 0.889 & 0.737 \\ 
   \hline
MOBS-GLEAM\_COVID & 195.585 & 359.708 & 151.764 & 250.017 & 0.927 & 0.819 \\ 
   \hline
COVIDhub-baseline & 285.742 & 447.762 & 160.885 & 271.000 & 0.838 & 0.692 \\ 
   \hline
GT-DeepCOVID & 288.326 & 447.215 & 160.238 & 284.163 & 0.834 & 0.690 \\ 
   \hline
JHUAPL-Bucky & 223.484 & \textbf{224.424} & 137.416 & 165.112 & 0.914 & \textbf{0.933} \\ 
   \hline
\end{tabular}
\caption{Comparison of different methods for state-level COVID-19 1 to 2 weeks ahead hospitalizations predictions in Maryland (MD). The MSE, MAE, and correlation are reported and best performed method is highlighted in boldface.} 
\end{table}
\begin{figure}[!h] 
  \centering 
\includegraphics[width=.8\linewidth, page=46]{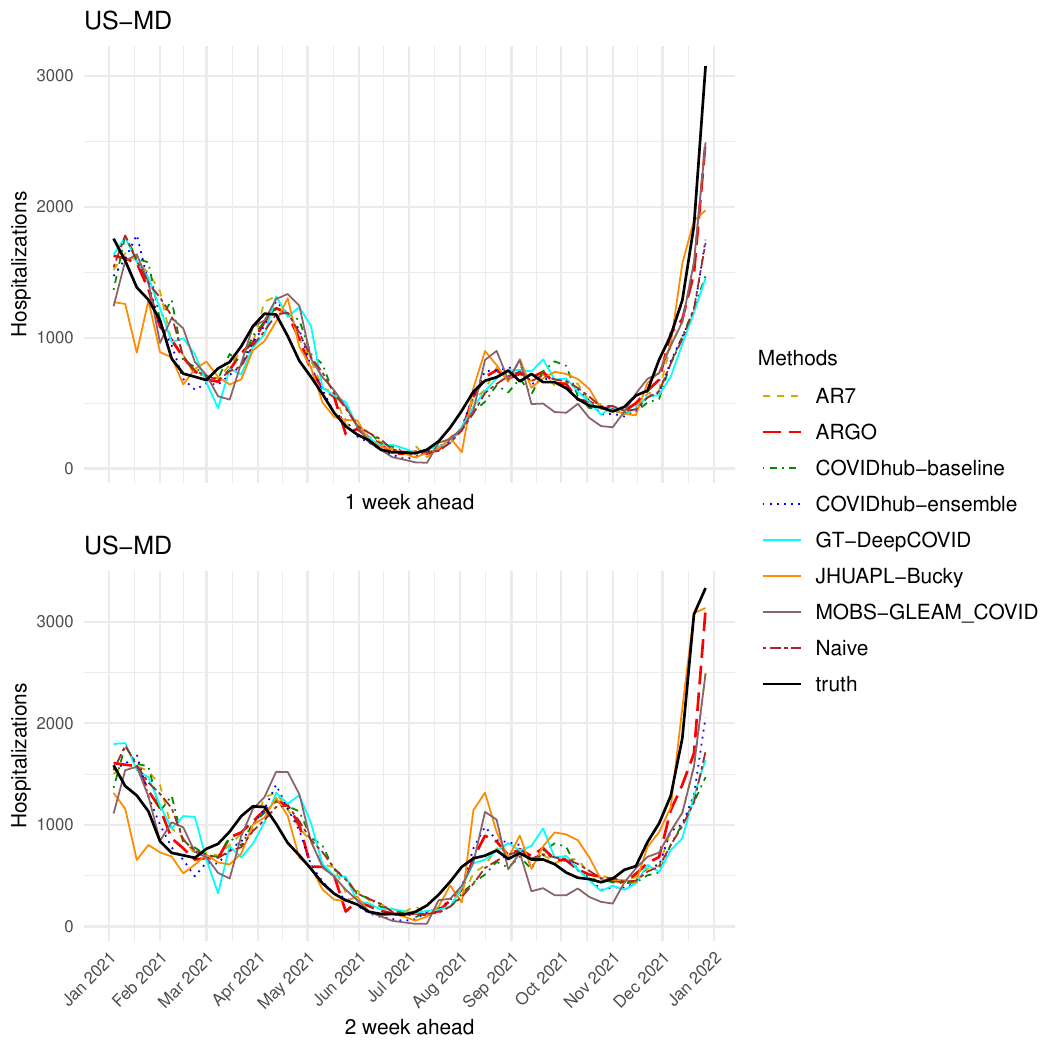} 
\caption{Plots of the COVID-19 hospitalizations 1 week (top), 2 weeks (bottom) ahead estimates of all compared models for New Maryland (MD).    }
\end{figure}
\newpage
\begin{table}[ht]
\footnotesize
\centering
\begin{tabular}{|l|r|r|r|r|r|r|}
   \hline\multicolumn{1}{|c}{Methods} & \multicolumn{2}{|c}{RMSE} & \multicolumn{2}{|c}{MAE}& \multicolumn{2}{|c|}{Cor} \\  \cline{2-7} & 1 Week Ahead & 2 Week Ahead & 1 Week Ahead & 2 Week Ahead & 1 Week Ahead & 2 Week Ahead \\ 
  \hline
ARGO & \textbf{209.962} & \textbf{463.977} & \textbf{143.836} & \textbf{319.985} & \textbf{0.974} & 0.862 \\ 
   \hline
AR7 & 223.781 & 477.016 & 156.253 & 346.971 & 0.968 & 0.835 \\ 
   \hline
COVIDhub-ensemble & 386.932 & 603.366 & 252.827 & 396.173 & 0.951 & \textbf{0.873} \\ 
   \hline
Naive & 305.156 & 551.819 & 209.654 & 375.365 & 0.939 & 0.788 \\ 
   \hline
MOBS-GLEAM\_COVID & 397.913 & 522.436 & 273.021 & 376.229 & 0.930 & 0.869 \\ 
   \hline
COVIDhub-baseline & 296.298 & 545.556 & 199.038 & 360.673 & 0.945 & 0.801 \\ 
   \hline
GT-DeepCOVID & 386.025 & 698.010 & 272.459 & 488.488 & 0.949 & 0.813 \\ 
   \hline
JHUAPL-Bucky & 469.580 & 628.119 & 339.331 & 451.020 & 0.857 & 0.756 \\ 
   \hline
\end{tabular}
\caption{Comparison of different methods for state-level COVID-19 1 to 2 weeks ahead hospitalizations predictions in Alabama (AL). The MSE, MAE, and correlation are reported and best performed method is highlighted in boldface.} 
\end{table}
\begin{figure}[!h] 
  \centering 
\includegraphics[width=.8\linewidth, page=47]{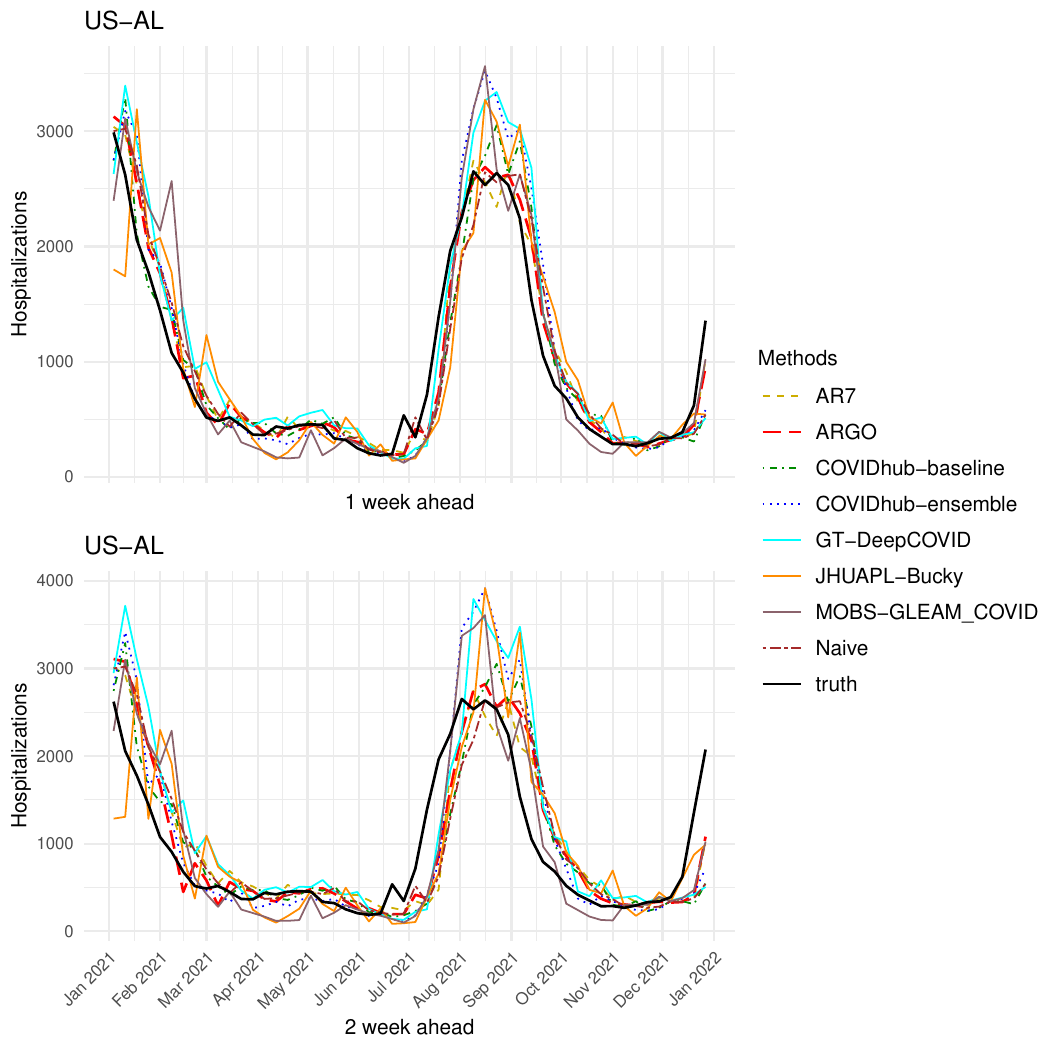} 
\caption{Plots of the COVID-19 hospitalizations 1 week (top), 2 weeks (bottom) ahead estimates of all compared models for New Alabama (AL).     }
\end{figure}
\newpage
\begin{table}[ht]
\footnotesize
\centering
\begin{tabular}{|l|r|r|r|r|r|r|}
   \hline\multicolumn{1}{|c}{Methods} & \multicolumn{2}{|c}{RMSE} & \multicolumn{2}{|c}{MAE}& \multicolumn{2}{|c|}{Cor} \\  \cline{2-7} & 1 Week Ahead & 2 Week Ahead & 1 Week Ahead & 2 Week Ahead & 1 Week Ahead & 2 Week Ahead \\ 
  \hline
ARGO & \textbf{129.526} & \textbf{272.484} & \textbf{96.264} & \textbf{199.592} & \textbf{0.972} & \textbf{0.874} \\ 
   \hline
AR7 & 163.866 & 309.442 & 113.818 & 225.714 & 0.954 & 0.835 \\ 
   \hline
COVIDhub-ensemble & 192.947 & 289.086 & 137.173 & 206.500 & 0.936 & 0.871 \\ 
   \hline
Naive & 185.534 & 336.480 & 141.596 & 257.365 & 0.932 & 0.796 \\ 
   \hline
MOBS-GLEAM\_COVID & 224.774 & 329.355 & 168.402 & 252.954 & 0.931 & 0.843 \\ 
   \hline
COVIDhub-baseline & 225.626 & 365.762 & 159.519 & 269.327 & 0.902 & 0.762 \\ 
   \hline
GT-DeepCOVID & 252.037 & 397.956 & 187.827 & 300.490 & 0.903 & 0.775 \\ 
   \hline
JHUAPL-Bucky & 765.690 & 890.613 & 437.979 & 508.849 & 0.572 & 0.505 \\ 
   \hline
\end{tabular}
\caption{Comparison of different methods for state-level COVID-19 1 to 2 weeks ahead hospitalizations predictions in Missouri (MO). The MSE, MAE, and correlation are reported and best performed method is highlighted in boldface.} 
\end{table}
\begin{figure}[!h] 
  \centering 
\includegraphics[width=.8\linewidth, page=48]{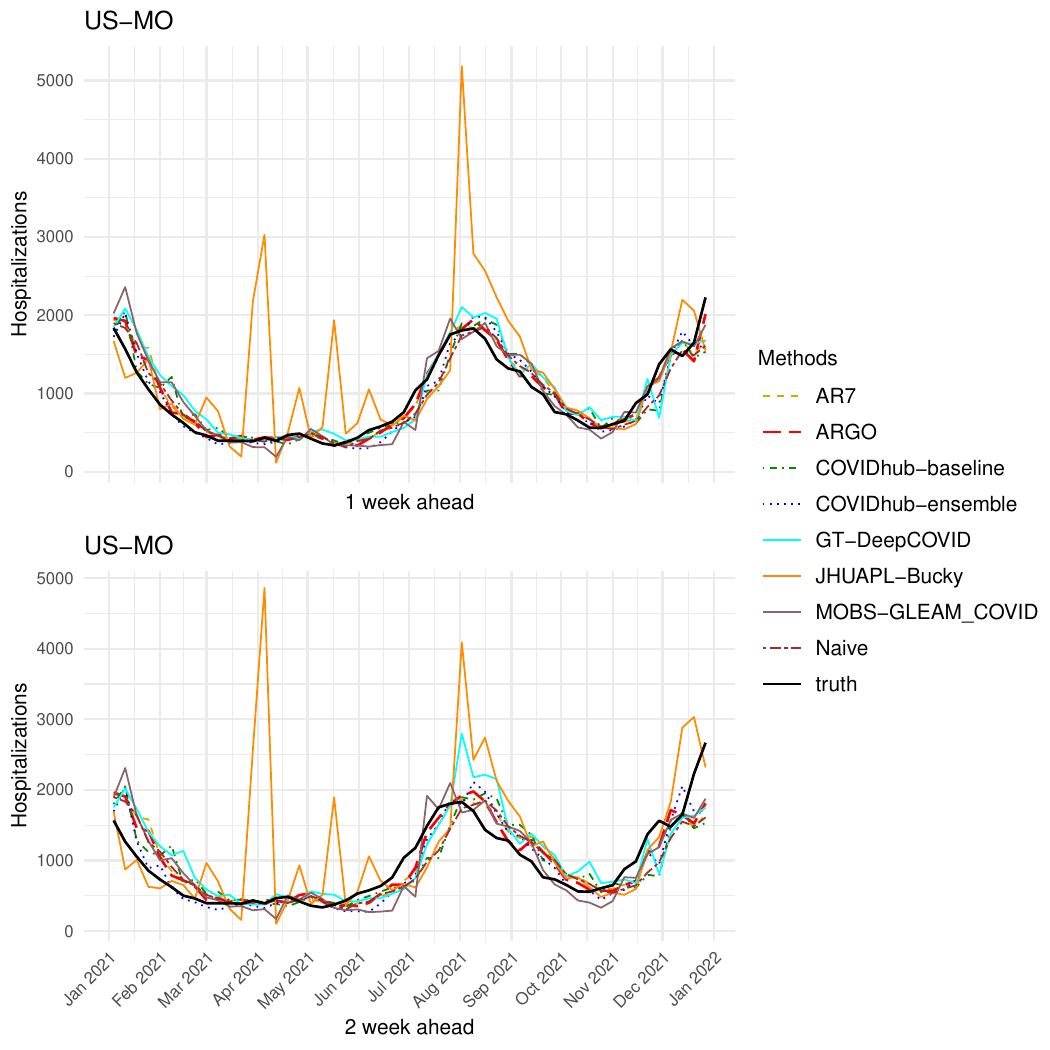} 
\caption{Plots of the COVID-19 hospitalizations 1 week (top), 2 weeks (bottom) ahead estimates of all compared models for New Missouri (MO).      }
\end{figure}
\newpage
\begin{table}[ht]
\footnotesize
\centering
\begin{tabular}{|l|r|r|r|r|r|r|}
   \hline\multicolumn{1}{|c}{Methods} & \multicolumn{2}{|c}{RMSE} & \multicolumn{2}{|c}{MAE}& \multicolumn{2}{|c|}{Cor} \\  \cline{2-7} & 1 Week Ahead & 2 Week Ahead & 1 Week Ahead & 2 Week Ahead & 1 Week Ahead & 2 Week Ahead \\ 
  \hline
ARGO & \textbf{126.048} & \textbf{251.517} & \textbf{94.747} & 190.974 & \textbf{0.978} & \textbf{0.918} \\ 
   \hline
AR7 & 133.824 & 273.208 & 106.503 & 213.514 & 0.977 & 0.902 \\ 
   \hline
COVIDhub-ensemble & 160.870 & 267.971 & 120.615 & \textbf{185.923} & 0.967 & 0.916 \\ 
   \hline
Naive & 168.777 & 293.922 & 130.192 & 228.673 & 0.958 & 0.885 \\ 
   \hline
MOBS-GLEAM\_COVID & 222.136 & 327.327 & 165.285 & 241.244 & 0.940 & 0.878 \\ 
   \hline
COVIDhub-baseline & 154.440 & 270.920 & 129.731 & 220.288 & 0.966 & 0.903 \\ 
   \hline
GT-DeepCOVID & 265.648 & 382.921 & 204.227 & 301.182 & 0.909 & 0.817 \\ 
   \hline
JHUAPL-Bucky & 357.833 & 443.747 & 265.208 & 328.514 & 0.888 & 0.855 \\ 
   \hline
\end{tabular}
\caption{Comparison of different methods for state-level COVID-19 1 to 2 weeks ahead hospitalizations predictions in Wisconsin (WI). The MSE, MAE, and correlation are reported and best performed method is highlighted in boldface.} 
\end{table}
\begin{figure}[!h] 
  \centering 
\includegraphics[width=.8\linewidth, page=49]{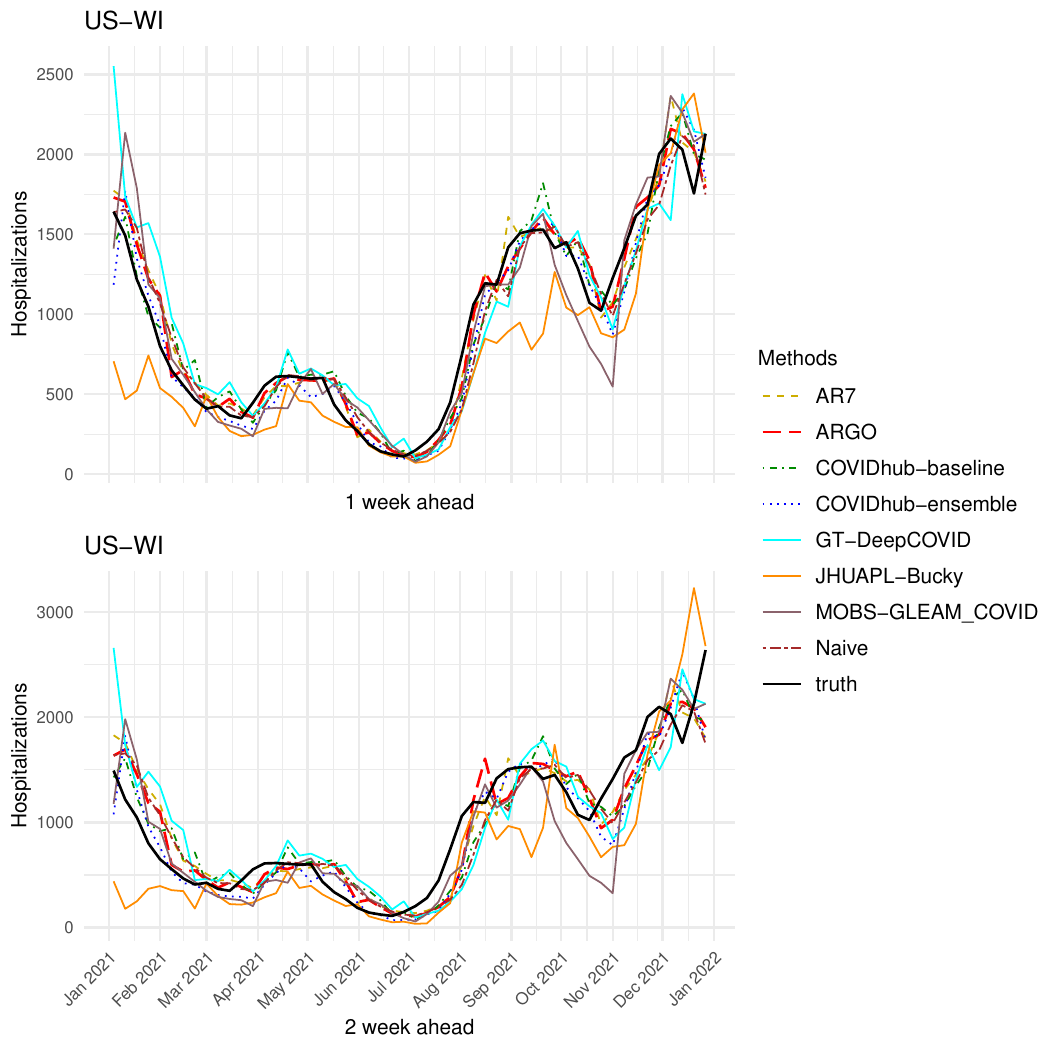} 
\caption{Plots of the COVID-19 hospitalizations 1 week (top), 2 weeks (bottom) ahead estimates of all compared models for New Wisconsin (WI).       }
\end{figure}
\newpage
\begin{table}[ht]
\footnotesize
\centering
\begin{tabular}{|l|r|r|r|r|r|r|}
   \hline\multicolumn{1}{|c}{Methods} & \multicolumn{2}{|c}{RMSE} & \multicolumn{2}{|c}{MAE}& \multicolumn{2}{|c|}{Cor} \\  \cline{2-7} & 1 Week Ahead & 2 Week Ahead & 1 Week Ahead & 2 Week Ahead & 1 Week Ahead & 2 Week Ahead \\ 
  \hline
ARGO & \textbf{201.898} & \textbf{341.561} & \textbf{146.537} & 272.739 & \textbf{0.967} & 0.921 \\ 
   \hline
AR7 & 248.998 & 414.342 & 173.771 & 311.518 & 0.956 & 0.862 \\ 
   \hline
COVIDhub-ensemble & 216.647 & 364.337 & 156.692 & \textbf{260.231} & 0.963 & 0.910 \\ 
   \hline
Naive & 252.979 & 428.509 & 192.692 & 339.538 & 0.940 & 0.837 \\ 
   \hline
MOBS-GLEAM\_COVID & 223.461 & 346.341 & 169.630 & 268.696 & 0.963 & \textbf{0.924} \\ 
   \hline
COVIDhub-baseline & 246.917 & 412.754 & 186.269 & 321.423 & 0.942 & 0.848 \\ 
   \hline
GT-DeepCOVID & 323.479 & 492.643 & 252.133 & 393.445 & 0.908 & 0.812 \\ 
   \hline
JHUAPL-Bucky & 293.929 & 572.103 & 214.833 & 382.662 & 0.941 & 0.891 \\ 
   \hline
\end{tabular}
\caption{Comparison of different methods for state-level COVID-19 1 to 2 weeks ahead hospitalizations predictions in Indiana (IN). The MSE, MAE, and correlation are reported and best performed method is highlighted in boldface.} 
\end{table}
\begin{figure}[!h] 
  \centering 
\includegraphics[width=.8\linewidth, page=50]{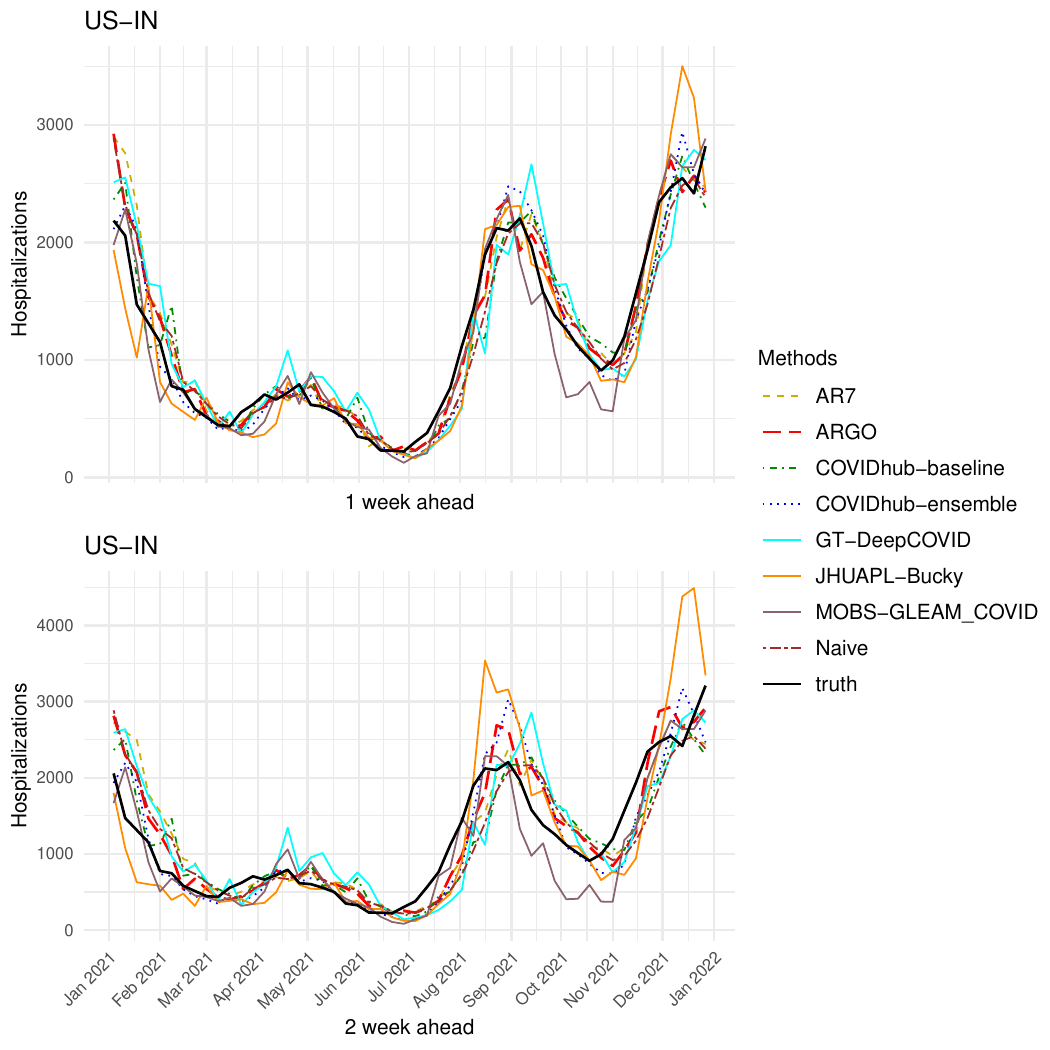} 
\caption{Plots of the COVID-19 hospitalizations 1 week (top), 2 weeks (bottom) ahead estimates of all compared models for New Indiana (IN).       }
\end{figure}
\newpage
\begin{table}[ht]
\footnotesize
\centering
\begin{tabular}{|l|r|r|r|r|r|r|}
   \hline\multicolumn{1}{|c}{Methods} & \multicolumn{2}{|c}{RMSE} & \multicolumn{2}{|c}{MAE}& \multicolumn{2}{|c|}{Cor} \\  \cline{2-7} & 1 Week Ahead & 2 Week Ahead & 1 Week Ahead & 2 Week Ahead & 1 Week Ahead & 2 Week Ahead \\ 
  \hline
ARGO & \textbf{539.491} & \textbf{1870.584} & \textbf{371.383} & \textbf{1142.924} & \textbf{0.991} & \textbf{0.905} \\ 
   \hline
AR7 & 799.802 & 2213.070 & 580.603 & 1510.918 & 0.979 & 0.842 \\ 
   \hline
COVIDhub-ensemble & 1611.048 & 2737.881 & 948.596 & 1625.808 & 0.929 & 0.813 \\ 
   \hline
Naive & 1602.244 & 2993.683 & 999.288 & 1888.058 & 0.913 & 0.712 \\ 
   \hline
MOBS-GLEAM\_COVID & 1695.726 & 3050.612 & 1123.527 & 1934.591 & 0.926 & 0.790 \\ 
   \hline
COVIDhub-baseline & 1563.981 & 2946.363 & 945.750 & 1814.865 & 0.917 & 0.722 \\ 
   \hline
GT-DeepCOVID & 1418.859 & 2834.502 & 988.366 & 1936.426 & 0.941 & 0.770 \\ 
   \hline
JHUAPL-Bucky & 2678.431 & 3979.651 & 1681.721 & 2416.568 & 0.887 & 0.757 \\ 
   \hline
\end{tabular}
\caption{Comparison of different methods for state-level COVID-19 1 to 2 weeks ahead hospitalizations predictions in Florida (FL). The MSE, MAE, and correlation are reported and best performed method is highlighted in boldface.} 
\end{table}
\begin{figure}[!h] 
  \centering 
\includegraphics[width=.8\linewidth, page=51]{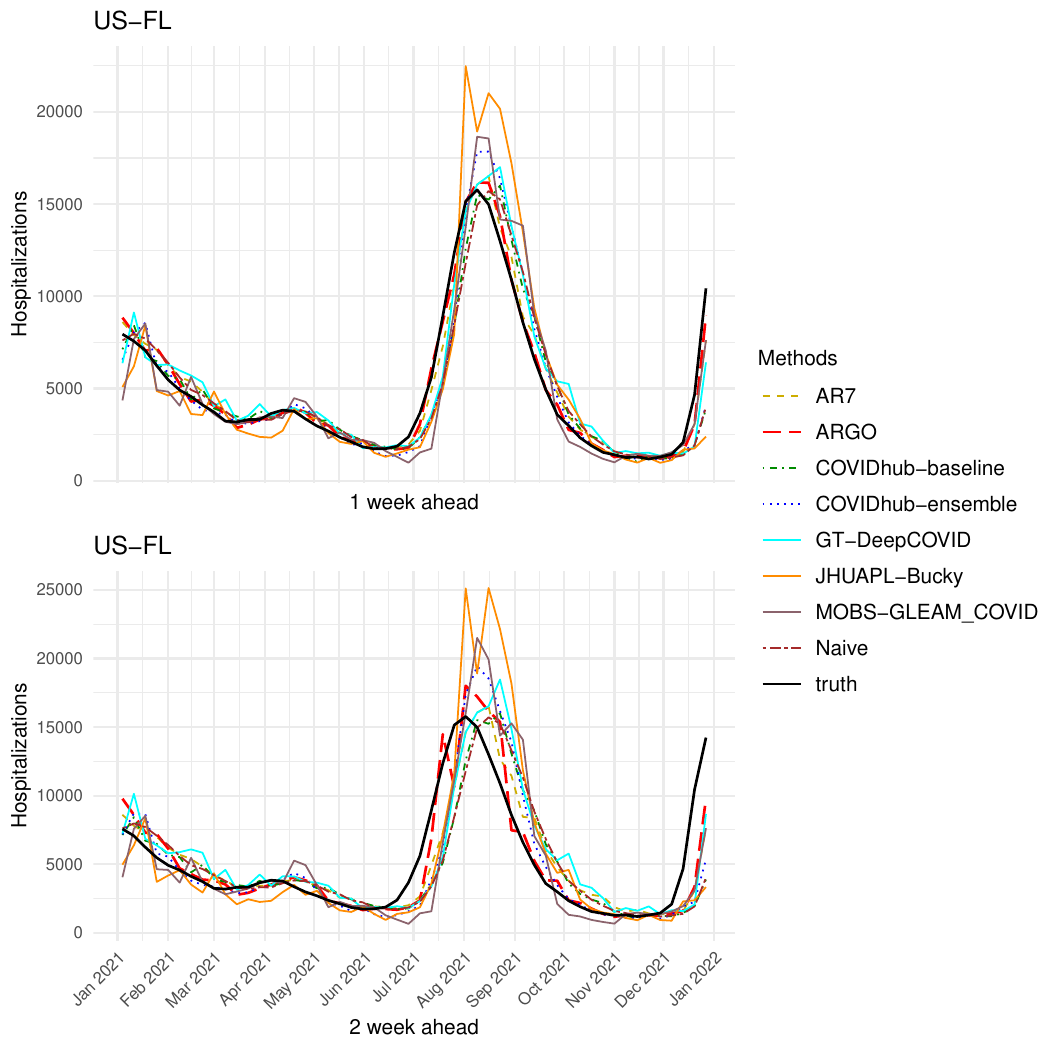} 
\caption{Plots of the COVID-19 hospitalizations 1 week (top), 2 weeks (bottom) ahead estimates of all compared models for New Florida (FL).        }
\end{figure}
\newpage
\begin{table}[ht]
\footnotesize
\centering
\begin{tabular}{|l|r|r|r|r|r|r|}
   \hline\multicolumn{1}{|c}{Methods} & \multicolumn{2}{|c}{RMSE} & \multicolumn{2}{|c}{MAE}& \multicolumn{2}{|c|}{Cor} \\  \cline{2-7} & 1 Week Ahead & 2 Week Ahead & 1 Week Ahead & 2 Week Ahead & 1 Week Ahead & 2 Week Ahead \\ 
  \hline
ARGO & \textbf{133.820} & \textbf{330.992} & \textbf{100.462} & \textbf{209.788} & \textbf{0.971} & 0.857 \\ 
   \hline
AR7 & 152.635 & 362.102 & 115.727 & 235.757 & 0.963 & 0.825 \\ 
   \hline
COVIDhub-ensemble & 221.614 & 395.270 & 127.942 & 230.538 & 0.924 & 0.804 \\ 
   \hline
Naive & 234.116 & 443.354 & 147.269 & 279.788 & 0.909 & 0.730 \\ 
   \hline
MOBS-GLEAM\_COVID & 175.217 & 363.410 & 136.032 & 249.937 & 0.959 & 0.851 \\ 
   \hline
COVIDhub-baseline & 380.940 & 536.810 & 201.308 & 305.519 & 0.762 & 0.600 \\ 
   \hline
GT-DeepCOVID & 250.643 & 466.511 & 159.137 & 299.153 & 0.898 & 0.723 \\ 
   \hline
JHUAPL-Bucky & 317.474 & 484.717 & 205.560 & 338.362 & 0.900 & \textbf{0.873} \\ 
   \hline
\end{tabular}
\caption{Comparison of different methods for state-level COVID-19 1 to 2 weeks ahead hospitalizations predictions in Virginia (VA). The MSE, MAE, and correlation are reported and best performed method is highlighted in boldface.} 
\end{table}
\begin{figure}[!h] 
  \centering 
\includegraphics[width=.8\linewidth, page=52]{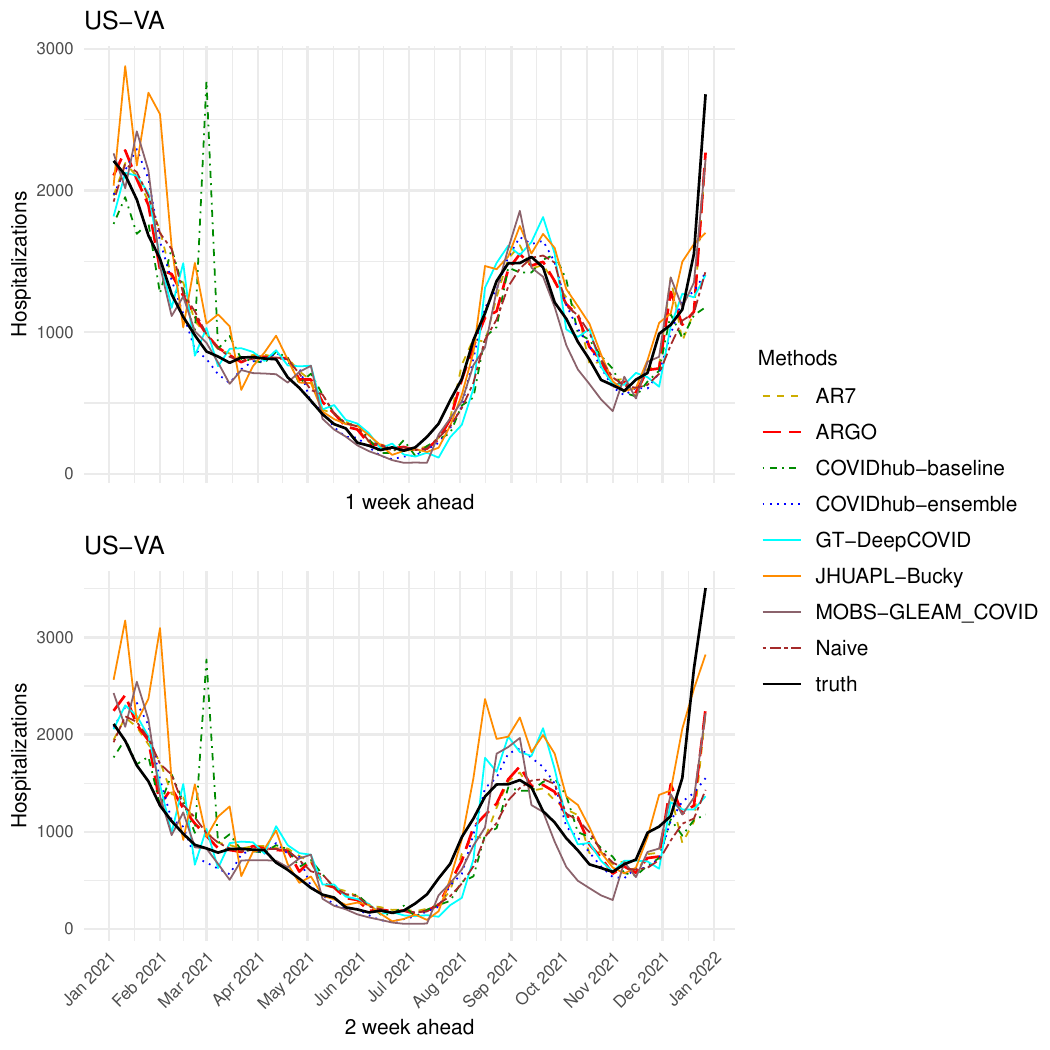} 
\caption{Plots of the COVID-19 hospitalizations 1 week (top), 2 weeks (bottom) ahead estimates of all compared models for New Virginia (VA).        }
\end{figure}
\newpage